\documentclass[11pt]{article}

\usepackage{acl}

\usepackage{latexsym}
\usepackage{iftex}
\ifpdftex
  \usepackage{times}
  \usepackage[T1]{fontenc}
  \usepackage[utf8]{inputenc}
  \usepackage{textcomp}
  \DeclareUnicodeCharacter{2248}{\ensuremath{\approx}}
  \DeclareUnicodeCharacter{2264}{\ensuremath{\leq}}
  \DeclareUnicodeCharacter{2265}{\ensuremath{\geq}}
\else
  \usepackage{fontspec}
\fi
\usepackage{microtype}
\usepackage{inconsolata}
\usepackage{graphicx}
\usepackage{hyperref}
\usepackage{booktabs}
\usepackage{amsmath}
\usepackage{longtable}
\usepackage{fvextra}
\usepackage{multicol}
\usepackage{tcolorbox}
\tcbuselibrary{skins,breakable}

\DefineVerbatimEnvironment{Prompt}{Verbatim}{%
  breaklines=true,breakanywhere=true,fontsize=\footnotesize,%
  breaksymbolleft={},breaksymbolright={},%
  formatcom=\rmfamily}

\newtcolorbox{PromptBox}{%
  enhanced jigsaw, breakable,
  colback=white, colframe=black!50,
  boxrule=0.4pt, arc=1.5pt,
  left=3pt, right=3pt, top=2pt, bottom=2pt,
  before skip=4pt, after skip=4pt}

\providecommand{\tightlist}{%
  \setlength{\itemsep}{0pt}\setlength{\parskip}{0pt}}

\providecommand{\pandocbounded}[1]{#1}

\usepackage{afterpage}
\usepackage{colortbl}  
\usepackage{framed}    
\definecolor{promptrule}{gray}{0.75}
\newenvironment{promptbox}
  {\par\addvspace{4pt}%
   \MakeFramed{\advance\hsize-\width\FrameRestore}\small}
  {\endMakeFramed\addvspace{4pt}}
\usepackage{fontspec}
\usepackage{unicode-math}
\makeatletter
\AtBeginDocument{%
  \let\acl@orig@maketitle\maketitle
  \def\maketitle{\begin{NoHyper}\acl@orig@maketitle\end{NoHyper}}%
}
\makeatother
\makeatletter
\@ifpackageloaded{caption}{}{\usepackage{caption}}
\AtBeginDocument{%
\ifdefined\contentsname
  \renewcommand*\contentsname{Table of contents}
\else
  \newcommand\contentsname{Table of contents}
\fi
\ifdefined\listfigurename
  \renewcommand*\listfigurename{List of Figures}
\else
  \newcommand\listfigurename{List of Figures}
\fi
\ifdefined\listtablename
  \renewcommand*\listtablename{List of Tables}
\else
  \newcommand\listtablename{List of Tables}
\fi
\ifdefined\figurename
  \renewcommand*\figurename{Figure}
\else
  \newcommand\figurename{Figure}
\fi
\ifdefined\tablename
  \renewcommand*\tablename{Table}
\else
  \newcommand\tablename{Table}
\fi
}
\@ifpackageloaded{float}{}{\usepackage{float}}
\floatstyle{ruled}
\@ifundefined{c@chapter}{\newfloat{codelisting}{h}{lop}}{\newfloat{codelisting}{h}{lop}[chapter]}
\floatname{codelisting}{Listing}

\makeatother
\makeatletter
\@ifpackageloaded{caption}{}{\usepackage{caption}}
\@ifpackageloaded{subcaption}{}{\usepackage{subcaption}}
\makeatother

\title{BenGER: Benchmarking LLM Systems on Subsumption-Based Legal
Reasoning in German Law}

\author{
  Sebastian Nagl\textsuperscript{1}\thanks{\,Core Contributor, corresponding author. Correspondence to \texttt{sebastian.nagl@tum.de}.} \quad
  Ann-Kristin Mayrhofer\textsuperscript{2}\thanks{\,Dataset contributor.} \quad
  Martin Heidebach\textsuperscript{2\textdagger} \\
  \bfseries Aleyna Koçak\textsuperscript{3\textdagger} \quad
  Anne Zettelmeier\textsuperscript{4\textdagger} \quad
  Elly Breu\textsuperscript{1\textdagger} \\
  \bfseries Angelina Greiner\textsuperscript{1\textdagger} \quad
  Sofija Milijas\textsuperscript{1\textdagger} \quad
  Matthias Grabmair\textsuperscript{1}\\[6pt]
  \mdseries \textsuperscript{1}Technical University of Munich (TUM) \quad
  \textsuperscript{2}Ludwig Maximilian University of Munich (LMU) \\
  \textsuperscript{3}University of Konstanz \quad
  \textsuperscript{4}University of Saarbrücken
}

\begin{document}
\maketitle

\begin{abstract}

We introduce BenGER (Benchmark for German Law), a benchmark and dataset
for subsumption-based legal reasoning in German law. It combines 596
exam-style free-text case tasks across several levels of legal education
with 531 short doctrinal reasoning tasks, and a controlled subset adds
timed human solutions written both unaided and in human-AI co-creation.
We evaluate 12 contemporary LLM systems (closed flagship,
efficiency-oriented, and open-weight) with a rubric-aligned
LLM-as-a-Judge, which we cross-validate against a multi-rater human
layer (three blind reviews per solution) and across six LLM judges. Our
results show that closed-flagship systems lead across all three corpora,
that human-AI co-creation measurably improves on unaided human work, and
that the judge tracks human grading at Pearson r=0.76 and Cohen's
κ=0.60, with rankings stable across judges and two judges from
independent providers clearing the Calderon single-reviewer replacement
bar on human-written solutions.

\end{abstract}

\afterpage{%
\begin{figure*}[t!]
\centering
\includegraphics[width=\textwidth]{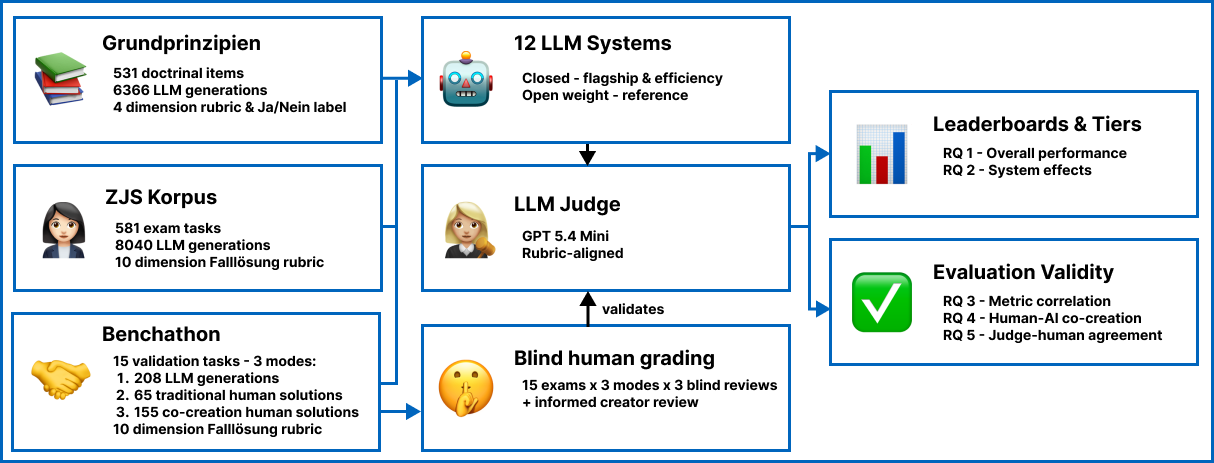}
\caption{Overview of the benchmark and evaluation pipeline. Three German-law corpora (the published ZJS exam corpus, the short Doctrinal Principles question set, and the controlled Benchathon validation subset with human solutions in two working conditions) are answered by a panel of LLM systems and scored by a rubric-aligned LLM judge. The Benchathon subset is additionally graded by a blind human reviewer pool (with one creator review per solution); the resulting human distribution is used to validate the LLM judge.}
\label{fig-overview}
\end{figure*}%
}

\section{Introduction}\label{introduction}

Evaluating legal reasoning is hard. Even human experts grade the same
answer very differently. Hufeld \citep{hufeld2024} shows that identical
answers in German law exams receive markedly different scores depending
on the grader. Doctrinal analysis is interpretive, so this is no
surprise. But it is a problem for benchmarking large language models
(LLMs). If human evaluation is itself noisy, what does it mean to
benchmark legal reasoning?

Legal NLP benchmarks have evolved fast. They moved from classification
\citep{chalkidis2022lexglue, guha2023legalbench} to reasoning
\citep{fei2024lawbench, fan2026lexam, jia2025readyjuristone, shi2026plawbench}
and retrieval-augmented evaluation
\citep{pipitone2024legalbenchrag, butler2026legalragbench}. But most
still emphasise short-form QA, common-law jurisdictions, or stable
ground-truth labels. Open-ended subsumption-based German legal reasoning
under noisy expert judgement remains under-served.

We address these gaps with a benchmark centered on subsumption-based
legal reasoning, the core reasoning paradigm in German law
(Section~\ref{sec-legal-methodology}) and, more broadly, in many
civil-law jurisdictions worldwide.

We present the BenGER (Benchmark for German Law) dataset\footnote{Full
  code and dataset are available at
  \url{https://github.com/SebastianNagl/benger-platform}.} for
evaluating LLMs on subsumption-based legal reasoning in German law. It
has three components. Together they cover 596 exam-style free-text case
tasks across different stages of legal education and 531 short doctrinal
reasoning tasks. It also adds human baselines, both traditional legal
work and human-AI co-creation, on a controlled Benchathon validation
subset. Figure~\ref{fig-overview} summarises the three components and
the generation-plus-evaluation pipeline. Generation, annotation, and
evaluation ran end-to-end on the BenGER platform \citep{nagl2026benger},
the open-source web application we built earlier for benchmarking
legal-domain LLM systems.

How do we evaluate free-text reasoning when the human judgement is
itself noisy? We adopt a hybrid framework. An LLM-as-a-Judge, aligned
with German grading rubrics, does the scoring. We cross-validate it
against three blind human reviews per solution, plus an un-blind review
by the task's creator as a reference signal. We treat evaluation as a
distribution over plausible expert assessments, not a single
deterministic label.

We make three contributions. (i) A benchmark for subsumption-based legal
reasoning in German law. (ii) An empirical evaluation of 12 contemporary
LLM systems (closed flagship, efficiency-oriented, and open-weight)
alongside human and co-creative baselines. (iii) An evaluation
methodology that accounts for the inherent variability of legal grading.

We organise the empirical analysis of the full BenGER dataset around
five questions. \textbf{RQ1 (Performance):} how well do contemporary LLM
systems solve open-ended subsumption-based German legal reasoning tasks,
and how do they compare to human baselines? \textbf{RQ2 (System-tier
effects):} how does performance vary across system tiers and
open-/closed-weight access modes? \textbf{RQ3 (Evaluation validity):}
how well do generic automatic metrics correlate with rubric-based legal
evaluation? \textbf{RQ4 (Human-AI co-creation):} does AI-assisted
drafting improve human performance, and how does it compare to
standalone LLM systems? \textbf{RQ5 (Grading reliability):} how closely
do LLM-judge evaluations align with the empirical distribution of human
grades, and are their deviations comparable to inter-rater variability?

\section{Subsumption-based legal reasoning}\label{sec-legal-methodology}

Subsumption is the core reasoning paradigm of German civil-law doctrine
\citep{larenz1995methodenlehre, moellers2023methodenlehre}. Unlike
common-law systems, where reasoning proceeds primarily through analogy
to precedent, German legal analysis derives outcomes from codified
statutory norms through a structured four-step scaffold: \emph{Obersatz}
(the norm hypothesis to be examined), \emph{Definition} (interpretation
of the norm's elements), \emph{Subsumtion} (anchoring the concrete facts
under those elements)\footnote{\emph{Subsumtion} carries a broad and a
  narrow sense. Broadly, it names the entire norm-application method,
  the sense in the paper title. Narrowly, it names the single
  fact-anchoring step of the scaffold, the sense used for the scaffold
  step and the rubric dimension of the same name.}, and \emph{Ergebnis}
(the resulting legal consequence). The scaffold is the German
articulation of the legal syllogism that underlies statutory reasoning
across civil-law jurisdictions; what is Germany-specific is its
rubric-graded written form, not the reasoning operation itself. This is
the \emph{Gutachtenstil}, the opinion style in which the conclusion
follows the reasoning, distinct from the result-first \emph{Urteilsstil}
used in judicial decisions; German legal education, state examinations,
and entry-level practice are all trained and assessed in this register.
Grading targets the written \emph{Falllösung} (case solution) end-to-end
rather than the final answer alone, because the rubric rewards how each
norm element is anchored in the specific facts.

This makes subsumption a structurally demanding generation task: a
system must identify the governing norm, articulate its elements, anchor
every element in the facts, and conclude, a chain in which surface
fluency can mask a missing \emph{Subsumtion} step. The same structure,
however, makes the task unusually amenable to rubric-based evaluation,
because each step in the scaffold maps to an identifiable part of the
answer rather than to a global quality impression. English-language
legal NLP benchmarks built on classification, multiple-choice, or
precedent-style reasoning
\citep{chalkidis2022lexglue, guha2023legalbench} do not exercise this
scaffold. German terms are glossed at first use throughout; a
consolidated glossary is in Appendix \ref{glossary-of-german-terms}.

\section{Related work}\label{related-work}

Legal NLP benchmarking has moved from classification baselines, LexGLUE
\citep{chalkidis2022lexglue} (seven English legal datasets) and
LegalBench \citep{guha2023legalbench} (162 mostly short-form
US-federal-law tasks), to reasoning-centric, jurisdiction-specific
evaluation. Recent work targets jurisdiction-specific reasoning: Chinese
law and legal practice
\citep{fei2024lawbench, li2024lexeval, shi2026plawbench}, Korean
bar-exam and provision-grounded reasoning
\citep{kim2024kbl, lee2025koblex}, free-text Greek bar exams
\citep{chlapanis2025greekbarbench}, ECHR argument chains
\citep{chlapanis2024larechr}, Thai financial and tax law
\citep{akarajaradwong2025nitibench}, and US labor-law statutory
simplification \citep{hariri2025laborbench}, as well as
retrieval-augmented evaluation
\citep{pipitone2024legalbenchrag, butler2026legalragbench} and
discrepancy-based auditing of LLM legal reasoning
\citep{choudhury2026clause}. The closest precedent is LEXam
\citep{fan2026lexam}: 7,537 questions across 340 Swiss law exams in
English and German, with an ensemble LLM-as-a-Judge validated against
human experts, but its rubric is not the
\emph{Obersatz-Definition-Subsumtion-Ergebnis} scaffold central to
German \emph{Falllösung} grading (Section~\ref{sec-legal-methodology})
and it does not couple its free-text items to the dimension-level
grading criteria used in German law school and state-exam evaluation.

Free-text legal evaluation increasingly relies on LLM-as-a-Judge
approaches, which raises concerns over bias, calibration, and alignment
with expert judgement. German legal education shows how variable human
grading already is: when Hufeld \citep{hufeld2024} had 23 graders score
15 beginner-level law exams, the average range between the lowest and
highest grade per answer was 6.47 points on the 0-18 scale, and only
42\% of grades fell within ±1 point of the answer-specific mean. Legal
grading is therefore better treated as a distribution over plausible
expert judgements than as a single ground truth, which motivates hybrid
methods that combine human expertise with calibrated LLM-based scoring.

Resources for German legal NLP remain limited: GerLayQA
\citep{buttner2024gerlayqa} provides \textasciitilde21,000 lay questions
paired with lawyer answers but focuses on QA rather than subsumption,
GerLeRB \citep{weber2025gerlerb} targets legislative retrieval, and our
prior BenGER platform \citep{nagl2026benger} provides an end-to-end
framework for German legal case analysis and subsumption-based reasoning
workflows, on which the present dataset was built. Overall,
German-law-style reasoning grounded in doctrinal subsumption under
statutory law remains under-served, especially for open-ended,
structured analysis under noisy human judgement. The BenGER dataset
addresses this gap by combining a 596-task exam-style corpus and 531
short doctrinal reasoning items with a controlled Benchathon validation
subset, a German doctrinal rubric, and a multi-rater cross-validation
protocol.

\section{Dataset}\label{dataset}

The BenGER dataset draws on three components covering the main forms of
legal reasoning in German legal education: two exam-style case corpora,
in which a fact pattern is given and a structured \emph{Falllösung} is
expected, and one set of short doctrinal reasoning items probing core
principles without the full case-analysis scaffolding.

Only the first component builds on pre-existing material. The ZJS corpus
contributes 581 publicly available exam-style cases from the
\emph{Zeitschrift für das Juristische Studium} (Journal for the Study of
Law)\footnote{\url{https://www.zjs-online.com/}}, covering civil,
criminal, and public law and spanning early undergraduate exercises
through first and selected second state-examination materials, each
paired with an expert-written reference solution. The cases and
reference solutions pre-exist this work as journal publications; our
contribution is the benchmark artifact: selection, machine-readable
task-reference pairs, per-item IP clearance
(\hyperref[ethical-considerations]{§Ethical considerations}). The
Benchathon corpus adds 15 newly collected exam-style tasks at an
intermediate difficulty level (more demanding than introductory
exercises, less complex than full first-state-exam problems). The tasks
and their reference solutions were authored for this benchmark by the
domain experts on the author team (the task creators of Appendix
\ref{full-benchathon-human-evaluation-assignment-procedure}). It
comprises 220 human solutions in total (65 traditional, 155 human-AI
co-creation; about 15 per task).\footnote{The corpus is named after the
  in-person Benchathon event at which participants produced these
  solutions and the validation human grading. Participants were
  predominantly law students at various stages of legal training, plus a
  smaller group of \emph{Referendare} (post-state-exam legal trainees),
  recent graduates, and a small layperson cohort; full composition in
  Appendix \ref{benchathon-participant-composition}.} The Doctrinal
Principles corpus (\emph{Grundprinzipien}) contributes 531 short
doctrinal reasoning items (question-answer pairs with explanatory
reasoning) on core principles, with recent case law where appropriate.
The items were newly written for this benchmark by a subset of the
authors with the relevant domain expertise and, like Benchathon, were
unpublished prior to this work. The three play distinct roles: ZJS
supports at-scale automated evaluation; Benchathon supports the human
baseline and judge-validation layer; Doctrinal Principles probes the
same reasoning at shorter form.

The benchmark is for evaluation only; we do not define a train/test
split, since the objective is comparative evaluation of deployed LLM
systems and human baselines. Pretraining-contamination considerations
across the three corpora are discussed in
\hyperref[limitations]{§Limitations}.

\subsection{Benchathon task design and human solution
collection}\label{benchathon-task-design-and-human-solution-collection}

Benchathon participants solved tasks in a controlled annotation
environment, with each task randomly assigned to one of two conditions.
In the \emph{traditional} condition, participants could use statutory
texts, legal databases, literature, and internet research. In the
\emph{co-creation} condition, they could additionally use any LLM(s) of
their choosing (commercial or open-weight, in any combination) through
their own accounts. We did not log which tool a participant used. The
co-creation condition therefore measures an upper bound on what
human-LLM teams currently achieve with off-the-shelf assistance, not a
controlled comparison against any specific system. Only the final
written solution is evaluated, mirroring legal examination practice.
Each task had a two-hour time limit; participant expertise was measured
objectively (reported legal grades and qualifications) and subjectively
(self-assessed competence on a Likert scale).

\section{Experimental setup}\label{experimental-setup}

\subsection{Generation}\label{generation}

We evaluate 12 contemporary LLM systems: 6 closed-weight through
provider APIs (OpenAI, Anthropic, Google) and 6 open-weight through a
managed inference provider (DeepInfra). Per-system providers, snapshots,
observed temperatures, token-budget settings, truncation counts, and the
short aliases used in the body are in Table~\ref{tbl-system-overview}
(Appendix~\ref{evaluated-systems-catalogue}). Each task received one
generation per system in the final run. All systems generated the
Benchathon subset. On ZJS, all 12 systems covered up to 581 published
exams.\footnote{Individual questions were occasionally flagged as
  prohibited content by the provider's safety filter, leaving a small
  handful of per-system coverage gaps below the corpus total.}

All systems received the same German-language system and instruction
prompts (Appendix \ref{system-prompts-for-generation} +
\ref{instruction-prompts-for-generation}). Each returned a single JSON
object containing the full case solution (\emph{Falllösung}). The
instruction prompt mirrors the \emph{Gutachtenstil} scaffold
(Section~\ref{sec-legal-methodology}). We used provider-recommended
temperatures and token budgets (temperature 0.6-1.0, max output
8,000-16,000 tokens; ZJS ran at 8,000). Generations that were malformed
or shorter than 200 output tokens after a bounded retry budget were
excluded. Of the ZJS generations truncated at that budget, 1,072 were
re-generated at 20,000 tokens; only the replacement enters the analyses.
An earlier partial Benchathon round (April 2026; different prompt
scaffold and settings, partially different model set) was likewise
superseded by the final run; its 28 generations from retained systems
are excluded from leaderboard aggregates, and the one blind-graded
validation solution drawn from it stays in the judge-validation layer
(Section~\ref{sec-human-eval}). Truncation in kept generations falls
mostly on the smaller open-weight systems, with a few on Gemini-3.1-Pro
(Table~\ref{tbl-system-overview}).

\subsection{Black-box system
evaluation}\label{black-box-system-evaluation}

We evaluate LLM systems through provider APIs in a black-box setting.
Our results therefore measure end-to-end system behaviour, not isolated
base-model capability. We do not enable retrieval or tool-use endpoints.
Components we cannot disable operate by default, notably content filters
and, on some products, internal routing between sub-models. This
reflects realistic deployment conditions. But it limits causal claims
about underlying architectures; for open weights we test the reasoning
mechanism directly with a same-base-model Thinking-vs.-Instruct toggle
(Appendix \ref{reasoning-mode-ablation}).

\subsection{Automatic metrics}\label{automatic-metrics}

Per generation we compute BLEU \citep{papineni2002bleu}, ROUGE
\citep{lin2004rouge}, METEOR \citep{banerjee2005meteor}, chrF
\citep{popovic2015chrf}, BERTScore \citep{zhang2020bertscore},
MoverScore \citep{zhao2019moverscore}, and a Sentence-BERT cosine
similarity \citep{reimers2019sentencebert} against the expert reference
solution. Per-system means appear in the main leaderboard tables.
Per-generation values are correlated against the LLM-judge rubric score
in RQ3.

\subsection{LLM-as-a-Judge evaluation}\label{llm-as-a-judge-evaluation}

For rubric-based evaluation on ZJS and Benchathon, we use a specialised
LLM-as-a-Judge prompt aligned with German legal grading practice. The
judge receives the task, the reference solution, and the solution to be
evaluated. It scores ten dimensions: result correctness
(\emph{Ergebnisrichtigkeit}), issue identification
(\emph{Vollständigkeit}), legal grounding (\emph{Rechtsgrundlagen}),
doctrinal knowledge (\emph{Rechtskenntnis}), subsumption quality
(\emph{Subsumtion}), problem depth (\emph{Schwerpunktsetzung}),
methodological structure (\emph{Methodischer Stil}), organization
(\emph{Gliederung}), terminology (\emph{Sprache}), and formal
correctness (\emph{Formalia}). Dimension scores aggregate to a 100-point
raw score, mapped to the German 0-18 grade-point scale. A generation
\emph{passes} when its grade-point score reaches \emph{ausreichend} (≥ 4
on the 0-18 scale, a raw score of ≥ 50/100), following the German
state-examination (\emph{Staatsexamen}) convention.\footnote{The ≥ 4
  pass mark is statutory: the federal grading ordinance
  \citep{jurprnotskv1981} fixes \emph{ausreichend} (4-6 of 18) as the
  lowest passing grade on a deliberately severe scale (≈30\% fail the
  first state examination). The raw-to-grade conversion is non-linear:
  grade 4 requires ≥ 50/100 raw points (table in Appendix
  \ref{evaluation-prompts-for-the-llm-judge}).} This is the boolean in
the \emph{Pass} columns of Table~\ref{tbl-main-leaderboard} and in the
pass-rate aggregates throughout. The rubric operationalises the
\emph{Gutachtenstil} scaffold (Section~\ref{sec-legal-methodology}):
each step in the \emph{Obersatz-Definition-Subsumtion-Ergebnis} schema
is reflected in a dedicated dimension (mapping and percentage weights in
Table~\ref{tbl-rubric-mapping}, Appendix \ref{full-results-table});
scaffold-step dimensions carry 90 of the 100 raw points, and a
leaderboard on those seven dimensions alone reproduces the full-rubric
system ranking exactly on both \emph{Falllösung} corpora (Spearman ρ =
1.00/1.00; Table~\ref{tbl-reasoning-core}). The per-dimension scores in
Figure~\ref{fig-rubric-dim} therefore track identifiable components of
the reasoning process. Grundprinzipien, the short-form Doctrinal
Principles corpus, uses a smaller 4-dimension rubric (result
correctness, legal knowledge, subsumption, clarity). It is not mapped to
the grade-point scale, since it does not follow the full
\emph{Gutachtenstil} scaffold.

The judge is blind to solution origin and receives inputs in a fixed
(task, reference, candidate) order. The primary judge is
GPT-5.4-mini\footnote{Configuration in Appendix
  \ref{llm-judge-configuration}, prompt in Appendix
  \ref{evaluation-prompts-for-the-llm-judge}. An earlier version of this
  work used GPT-5-mini as primary judge; the final version re-scored
  every generation with GPT-5.4-mini and extended cross-validation to
  six judges, with tier-level conclusions unchanged (details in Appendix
  \ref{operational-metadata-and-reproducibility}).}. Judge-based
evaluation may itself be model-dependent. We therefore cross-validate
the judge against a multi-rater human-grading layer on the Benchathon
subset. We also run two stability checks on that subset:
\emph{within-judge stochasticity} (k=3 re-runs of GPT-5.4-mini, reported
in Appendix \ref{sec-judge-calibration-supplementary-detail}) and
\emph{between-judge bias} (five additional judges, listed in
Section~\ref{sec-human-eval}, reported in RQ5). Outputs conform to a
predefined JSON schema. Non-conforming outputs are retried, and reported
as evaluation failures if they cannot be parsed.

\begin{table*}[!t]
\centering\footnotesize
\caption{Main leaderboard across the three corpora. Per-system mean LLM-judge raw score (0-100, mean\,$\pm$\,half-width of a 95\% CI: percentile bootstrap $B=2000$ on Benchathon for small per-system $n$, analytic 1.96\,$\cdot$\,SE on ZJS and Doctrinal Principles) and pass rate. Doctrinal Principles also reports yes/no (\emph{Ja/Nein}) decision accuracy. Systems are ranked by the mean of their available per-corpus raw scores. Per-system $n$ sits below the corpus total where the provider's safety filter refused individual items (\S\ref{generation}). The two human-baseline rows use the LLM judge's grades of human-written Benchathon solutions.}
\label{tbl-main-leaderboard}
\setlength{\tabcolsep}{4pt}
\begin{tabular*}{\textwidth}{@{\extracolsep{\fill}} l rrr rrr rrrr @{}}
\toprule
 & \multicolumn{3}{c}{Benchathon} & \multicolumn{3}{c}{ZJS} & \multicolumn{4}{c}{Doctrinal Principles} \\
\cmidrule(lr){2-4} \cmidrule(lr){5-7} \cmidrule(lr){8-11}
System / Group & $n$ & Raw & Pass & $n$ & Raw & Pass & $n$ & Raw & Pass & Acc. \\
\midrule
Gemini-3.1-Pro & 15 & 68.3\,$\pm$\,8.3 & 93\% & 579 & 61.8\,$\pm$\,1.0 & 83\% & 531 & 83.2\,$\pm$\,1.8 & 87\% & 83\% \\
Opus-4.7 & 15 & 69.3\,$\pm$\,5.5 & 93\% & 581 & 58.7\,$\pm$\,1.1 & 76\% & 531 & 83.1\,$\pm$\,2.0 & 86\% & 81\% \\
GPT-5.4 & 15 & 66.6\,$\pm$\,4.9 & 100\% & 581 & 60.4\,$\pm$\,1.1 & 78\% & 531 & 80.3\,$\pm$\,2.1 & 82\% & 79\% \\
Sonnet-4.6 & 15 & 58.9\,$\pm$\,5.9 & 73\% & 581 & 51.5\,$\pm$\,1.0 & 50\% & 531 & 81.6\,$\pm$\,2.1 & 83\% & 81\% \\
Gemini-3.1-Flash-Lite & 15 & 63.5\,$\pm$\,5.7 & 87\% & 579 & 44.6\,$\pm$\,0.8 & 26\% & 531 & 73.3\,$\pm$\,2.3 & 76\% & 76\% \\
GPT-5.4-mini & 15 & 58.6\,$\pm$\,5.6 & 73\% & 581 & 49.7\,$\pm$\,0.9 & 46\% & 531 & 71.8\,$\pm$\,2.4 & 72\% & 71\% \\
DeepSeek-V4-Pro & 15 & 56.1\,$\pm$\,6.1 & 67\% & 581 & 50.1\,$\pm$\,1.0 & 45\% & 531 & 73.5\,$\pm$\,2.3 & 73\% & 73\% \\
DeepSeek-V4-Flash & 15 & 49.3\,$\pm$\,3.7 & 47\% & 581 & 45.9\,$\pm$\,0.8 & 30\% & 530 & 70.7\,$\pm$\,2.4 & 73\% & 69\% \\
Qwen3.5-122B & 15 & 49.4\,$\pm$\,4.9 & 40\% & 581 & 44.2\,$\pm$\,0.8 & 25\% & 526 & 69.8\,$\pm$\,2.4 & 75\% & 78\% \\
Qwen3.6-35B & 15 & 42.4\,$\pm$\,7.6 & 27\% & 581 & 39.8\,$\pm$\,0.8 & 13\% & 531 & 69.5\,$\pm$\,2.5 & 69\% & 71\% \\
Qwen3-235B & 15 & 41.9\,$\pm$\,4.3 & 20\% & 581 & 35.9\,$\pm$\,0.7 & 6\% & 531 & 67.3\,$\pm$\,2.5 & 69\% & 73\% \\
Llama-4 & 15 & 37.8\,$\pm$\,4.3 & 20\% & 581 & 36.1\,$\pm$\,0.6 & 4\% & 531 & 70.1\,$\pm$\,2.3 & 75\% & 77\% \\
\midrule
Human (all) & 220 & 61.1\,$\pm$\,2.1 & 77\% & --- & --- & --- & --- & --- & --- & --- \\
\quad traditional & 65 & 50.0\,$\pm$\,3.6 & 52\% & --- & --- & --- & --- & --- & --- & --- \\
\quad co-creation & 155 & 65.7\,$\pm$\,2.1 & 88\% & --- & --- & --- & --- & --- & --- & --- \\
\bottomrule
\end{tabular*}
\end{table*}

\subsection{Human evaluation}\label{sec-human-eval}

Human evaluation runs on a controlled subset of 45 Benchathon solutions:
one traditional human, one human-AI co-created, and one LLM-generated
solution per Benchathon task. Human submissions are the median-expertise
participant per legal area (\emph{Bereich}). LLM submissions are drawn
through a tier-balanced rotation across closed-flagship,
efficiency-oriented, and open-weight reference systems. Each selected
solution receives three independent blind reviews and one un-blind
\emph{creator review} by the task's creator (180 reviews total).
Reviewers grade against the same rubric as the LLM judge and form the
IRR pool used to validate it. Creator reviews are excluded from IRR,
since creators have privileged knowledge of the intended solution. The
selection procedure, the seven-grader roster, and the \(m=5\) blind
subset (the graders who performed blind reviews; each solution is
reviewed by three of them) are described in
Appendix~\ref{full-benchathon-human-evaluation-assignment-procedure}.

The central question is not whether the LLM judge reproduces a single
human grade, but whether its deviation is comparable to that among human
graders. Alongside the Calderon alt-test \citep{calderon2025}, RQ5
reports descriptive agreement statistics: Spearman, Pearson, MAE,
Cohen's \(\kappa\) \citep{cohen1960coefficient}, ICC(2,1) and ICC(2,k)
\citep{shrout1979intraclass, koo2016guideline}. Each answers a distinct
question: rank vs.~absolute agreement, error magnitude, the pass/fail
threshold decision, and the reliability ceiling of the human pool itself
(per-statistic rationale in Appendix
\ref{judge-human-agreement-statistics}). The Calderon alt-test is the
primary criterion: it is purpose-built to decide whether the judge can
stand in for a blind reviewer. For the alt-test we follow the blind-pool
procedure at \(\varepsilon=0.15\) (Calderon's skilled-annotator tier),
with Benjamini-Yekutieli FDR at \(q=0.05\) across the \(m\) blind
reviewers. We also report the single-expert variant against the un-blind
creator grade, at \(\varepsilon=0.20\) because Calderon recommends the
stricter expert-tier penalty when the reference grader is an expert,
which the creator is. Following Calderon's single-expert case, this
variant treats the un-blind creator grade as the sole reference: each
blind reviewer and the judge are scored by absolute distance to that
single creator grade on every shared solution, and \(\omega\) is the
fraction of the \(m\) blind reviewers the judge matches or beats. So it
needs one creator grade per solution, not a pool of expert votes. The
judge passes when the winning rate \(\omega \geq 0.5\). The primary
judge is GPT-5.4-mini; we additionally re-score the same solutions with
Opus-4.7, Gemini-3.1-Pro, DeepSeek-V4-Pro, Qwen3.5-397B-A17B, and
Sonnet-4.6 (per-judge alt-test in
Tables~\ref{tbl-calderon-per-annotator-human} and
\ref{tbl-calderon-per-annotator-llm}).

\section{Results}\label{results}

Unless otherwise noted, scores in this section are the LLM judge's
rubric scores, as 0-100 raw points or 0-18 German grade points. The
human-grading layer covers the 45-solution validation design: one
creator review and three blind reviews per solution (180 reviews total).
The released analysis pipeline regenerates every number reported below
from the data files published with the paper.\footnote{\url{https://doi.org/10.5281/zenodo.20489788}}

\subsection{RQ1: Overall performance of LLM systems and
humans}\label{rq1-overall-performance-of-llm-systems-and-humans}

Table~\ref{tbl-main-leaderboard} gives the unified per-system view
across the three corpora; per-corpus full leaderboards with
surface-metric columns are in Appendix~\ref{full-results-table}. On
Benchathon, the top of the table is a closed-flagship cluster (Opus-4.7,
Gemini-3.1-Pro, and GPT-5.4 in the upper 60s on the 0-100 judge scale,
with pass rates above 90\%), followed by the efficiency-oriented systems
in the upper 50s to lower 60s. The strongest open-weight system,
DeepSeek-V4-Pro, trails the top of the flagship cluster by roughly 13
raw points. The open-weight tier spans a wide range; the nominal
flagship Qwen3-235B sits 27 raw points behind the lead and underperforms
its size here, despite no truncation. The same ordering broadly
reproduces on ZJS, where Gemini-3.1-Pro edges ahead of GPT-5.4 and the
weakest open-weight systems trail the field; per-system scores are
reported in Table~\ref{tbl-main-leaderboard}.

Within-task variance is substantial. Under the GPT-5.4-mini judge,
per-system standard deviations across the Benchathon tasks span roughly
8-18 raw points across all tiers. So single-task performance is a noisy
signal, even for top-tier systems. On Benchathon, the human cohort
scores in the mid-range of the LLM distribution under both conditions,
shifted upward under co-creation (see human rows in
Table~\ref{tbl-main-leaderboard} and Figure~\ref{fig-human-vs-ai} in
RQ4).

The same flagship cluster leads on the short-form Doctrinal Principles
corpus, with Sonnet-4.6 joining it. Doctrinal Principles items carry a
yes/no (\emph{Ja/Nein}) decision label; across systems, decision
accuracy ranges from 69\% to 83\%.\footnote{The corpus's gold labels are
  52\% yes, 48\% no.}

\subsection{RQ2: Scaling and system
effects}\label{rq2-scaling-and-system-effects}

Closed flagship systems from OpenAI, Anthropic, and Google occupy the
top of every per-corpus leaderboard. Efficiency-tier closed systems and
the strongest open-weight system (DeepSeek-V4-Pro) form an overlapping
middle group. Closed-API dominance narrows below the flagship tier. By
tier, the flagship-vs-open-weight gap is 20 raw points on Benchathon and
narrows to 16 on ZJS. The closed-vs-open-weights gap shrinks similarly,
from 18 to 12 raw points. The same direction holds on Doctrinal
Principles. The flagship-vs-open-weight ordering therefore survives all
three task formats and the 4-dimension rubric variant. We frame these as
system-level effects, not necessarily model-level ones. Black-box API
access entangles the open-closed split with provider-side prompts,
decoding, post-processing, and possible hidden retrieval or tool use
(see \hyperref[limitations]{Limitations}).

\subsection{RQ3: Evaluation validity - automatic metric
correlation}\label{rq3-evaluation-validity---automatic-metric-correlation}

Across all three corpora, lexical surface metrics correlate with the
judge's rubric raw score more strongly than embedding-based metrics do.
This runs partly against the common expectation that semantic-embedding
metrics dominate lexical ones on long-form legal text. On Benchathon the
lexical metrics lead: METEOR (r=+0.61), ROUGE (r=+0.69), and BLEU
(r=+0.58). MoverScore (r=+0.62) is the one embedding metric on par with
them. BERTScore (r=+0.48) and sentence-embedding similarity (r=+0.28)
trail. The same ordering broadly holds on ZJS and Doctrinal Principles
(Appendix~\ref{automatic-metric-vs.-llm-judge-correlation}:
Table~\ref{tbl-metric-correlation},
Figure~\ref{fig-metric-correlation}). This fits the highly templated
structure of doctrinal German exam writing, where reference and
high-quality system answers share substantial n-gram material around
statutory citations and the
\emph{Obersatz-Definition-Subsumtion-Ergebnis} scaffolding.

At the per-system level, the ordering shifts. Embedding-based metrics
rank systems at least as well as lexical ones. On Benchathon, ROUGE
leads (Spearman +0.88), but BERTScore (+0.83) and MoverScore (+0.85)
outrank BLEU (+0.69) and METEOR (+0.63). On ZJS, BERTScore is the single
strongest ranker (+0.97). A leaderboard built on a strong surface or
embedding metric would broadly preserve the judge's coarse-tier
ordering, but disagree on the mid-tier. This motivates a calibrated LLM
judge as the primary scoring instrument.

A natural concern is that the lexical advantage could reflect a
judge-specific preference for reference-like text rather than a property
of the task. Two analyses on the 45-solution validation subset say
otherwise. First, the lexical metrics correlate strongly with the mean
blind-human grade as well (METEOR: r≈+0.72 pooled). Second, we correlate
the judge's deviation from that human grade (judge − mean blind grade)
with each lexical metric; the result is \emph{negative} for all three:
relative to blind human graders, the judge \emph{under}-rewards
high-overlap solutions, the opposite of a reference-style bias
(Table~\ref{tbl-metric-vs-human} and Appendix
\ref{automatic-metric-vs.-llm-judge-correlation}, with per-metric
values, length-controlled partials, and small-\(n\) caveats).

\subsection{RQ4: Human-AI co-creation under LLM-judge
scoring}\label{rq4-human-ai-co-creation-under-llm-judge-scoring}

\begin{figure}[t]

\centering{

\pandocbounded{\includegraphics[keepaspectratio]{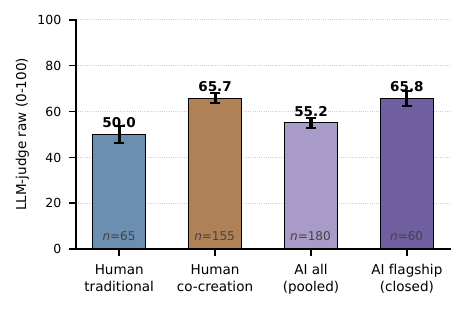}}

}

\caption{\label{fig-human-vs-ai}Mean LLM-judge raw score (0-100) per
group on the Benchathon validation subset, with 95\% bootstrap CIs on
each group's mean. The two AI bars pool per-generation scores across all
evaluated LLM systems (\emph{pooled}) and across the closed-flagship
tier only (\emph{flagship}).}

\end{figure}%

Co-creation solutions score higher than traditional ones on Benchathon,
by +15.7 raw points on average, about +2.8 grade points on the 0-18
scale (Figure~\ref{fig-human-vs-ai}; participant-clustered 95\% CI
{[}+9.6, +22.4{]}). The gap holds within the 23 participants who
submitted solutions in both conditions, traditional and co-creation. It
repeats across all three legal areas. It is in fact a lower bound on the
per-participant gain: less-experienced participants skipped traditional
items more often than co-creation ones, leaving the traditional pool
expertise-biased upward (full robustness arc in
Appendix~\ref{rq4-supplementary}). With that lift, the AI-assisted
humans sit alongside the best standalone LLMs we evaluated. A TOST
equivalence test rules out any difference of more than \(\pm 5\) raw
points against the closed-flagship tier
(Table~\ref{tbl-main-leaderboard}, Appendix~\ref{rq4-supplementary}).
Re-graded by blind human reviewers on the RQ5 validation subset, the gap
shrinks to about 14 raw points. That is +2 points smaller than the
LLM-judge estimate, a small absolute discrepancy. It is consistent with
the GPT-5.4-mini judge's near-zero content-dependent calibration offset,
and within the blind-pool's inter-rater spread
(\hyperref[limitations]{Limitations}).

\subsection{RQ5: Grading reliability}\label{rq5-grading-reliability}

The reliability question is simple: can the LLM judge stand in for a
blind human reviewer? We answer it in three stages: how closely the
judge agrees with humans per solution, whether it clears the formal
substitution bar, and what its disagreements look like when they happen.

Across the validation subset, the primary GPT-5.4-mini judge agrees with
the per-solution mean of the blind reviewers at Pearson r=0.76,
pass/fail Cohen's \(\kappa\)=0.60, and MAE 10.8 raw / 2.7 grade points.
Averaged over solutions, the judge sits just -0.01 raw points from the
blind-reviewer mean. The reviewers themselves disagree by an average
within-annotation max-min spread of 23.9 raw points, comparable to the
several-grade-point range Hufeld \citep{hufeld2024} documents for German
legal grading. The pool reaches ICC(2,\(k\))=0.84. So the judge sits
inside the cloud of human disagreement, not outside it
(Table~\ref{tbl-agreement}).

Per-solution agreement does not by itself show that the judge can
\emph{replace} a randomly drawn reviewer. The Calderon alt-test sets a
stricter bar: \(\varepsilon=0.15\) for the blind grader pool
(\emph{skilled}) and \(\varepsilon=0.2\) for the single un-blind creator
reference (\emph{expert}), with Benjamini-Yekutieli FDR
(§\ref{sec-human-eval}). We apply it across six judges: the primary
GPT-5.4-mini and five additional judges, Opus-4.7, Gemini-3.1-Pro,
DeepSeek-V4-Pro, Qwen3.5-397B-A17B, and Sonnet-4.6.

For the blind pool, two judges clear the \(\omega \geq 0.5\) replacement
bar on the human-written solutions: GPT-5.4-mini at \(\omega=\) 0.60 and
Opus-4.7 at \(\omega=\) 0.60. Sonnet-4.6 (\(\omega=\) 0.40) and
Gemini-3.1-Pro (\(\omega=\) 0.40) fall one annotator-rejection short.
DeepSeek-V4-Pro and Qwen3.5-397B-A17B fail at \(\omega \leq 0.20\). At
\(m=5\) these are point estimates; the two-judge pass shows a direction,
but the reviewer sample is too small to establish it statistically
(\hyperref[limitations]{Limitations}).

Against the single-expert creator reference, GPT-5.4-mini passes on
human-written solutions (\(\omega=\) 0.60) but fails on LLM-generated
ones (\(\omega=\) 0.00) (Table~\ref{tbl-agreement}; per-judge
\(\omega\), per-annotator \(p\)-values, and a bootstrap robustness check
in Appendix~\ref{sec-judge-calibration-supplementary-detail}).

Agreement is highest on scoring dimensions with structured criteria and
lowest on presentation and form (Figure~\ref{fig-rubric-dim}). Several
outcome and reasoning dimensions drop on LLM-generated solutions. Result
correctness goes from r=0.76 (human-written) to r=0.62 (LLM), with
comparable or larger drops on Issue identification, Doctrinal knowledge,
and Problem depth.

\subsection{Error analysis}\label{error-analysis}

On Benchathon and ZJS, the two free-text corpora, we find two dominant
failure modes, one typical of the flagship tier and one of the
open-weight tier. (i) \emph{Long-but-shallow generations}, concentrated
in the flagship tier, run to over 2,000 words yet score at most half
marks on \emph{Subsumtion} and problem depth. The pattern is
over-writing, which the rubric does not reward. (ii) \emph{Wrong-norm
framing generations}, concentrated in open-weight systems, are
methodologically well-formed but score low on legal grounding and
doctrinal knowledge. These solutions argue from the wrong legal norms
rather than making mistakes while applying the right ones. The two modes
are largely distinct (119 of the \textasciitilde1,800 flagged ZJS
generations fall in both). Counts of each failure mode, and how often it
occurs per tier, are in Appendix \ref{error-analysis-mode-counts} for
all three corpora.

On the short-form Doctrinal Principles corpus, yes/no accuracy follows
the same tier ordering, and the wrong decision itself is the dominant
error. Fact-bound \emph{Subsumtion} and outcome correctness are tightly
coupled on every corpus, so reasoning-outcome divergence is rare but
does occur. Where it occurs, the dimension scores show a consistency
failure rather than a reasoning failure: the model reaches a wrong
decision despite sound \emph{Subsumtion}, or the right decision on
flawed \emph{Subsumtion}. Its stated answer is decoupled from its own
reasoning, an inconsistency that yes/no accuracy cannot capture but the
rubric does.

\section{Discussion}\label{discussion}

Across the three corpora, Opus-4.7, Gemini-3.1-Pro, and GPT-5.4
consistently form the top of the leaderboard, with Sonnet-4.6 joining
them on Doctrinal Principles. The flagship-vs-open-weight gap is
\textasciitilde20 raw points on Benchathon (RQ2), narrower on the longer
ZJS corpus, and most compressed on Doctrinal Principles. The strongest
open-weight system, DeepSeek-V4-Pro, scores within the range of the
closed efficiency-oriented systems on the longer corpora and comes close
to the flagships on the short one, so the practical question for German
legal NLP is not whether open-weight systems can be used, but whether
the residual flagship advantage justifies the open- versus closed-weight
access constraints (Appendix~\ref{evaluated-systems-catalogue}).

Surface metrics (BLEU, ROUGE, METEOR) correlate with the rubric judge
more strongly than embedding metrics do (RQ3), consistent with the
templated \emph{Obersatz-Definition-Subsumtion-Ergebnis} structure of
German exam writing. A lexical-only leaderboard would keep the coarse
tiers but disagree on the mid-tier, so the rubric judge remains the
primary instrument and surface metrics read as cheap correlates.

Human-AI co-creation lifts unaided humans into the closed-flagship score
range, with a magnitude caveat: AI-assisted human work scores +15.7 raw
points (≈ +2.8 German grade points) above unaided human work under the
judge (RQ4), the gap holds within-participant and across the three legal
areas, and a TOST equivalence test renders it statistically
indistinguishable from the closed-flagship LLM tier
(Appendix~\ref{rq4-supplementary}). Re-graded by blind human reviewers,
the gap comes out about 2 raw points smaller, with the same sign and the
same conclusion. That difference is far smaller than the disagreement
among the human graders themselves.

The primary GPT-5.4-mini judge tracks the per-solution blind-reviewer
mean at Pearson r=0.76 and sits inside the human disagreement cloud
(RQ5). On the Calderon alt-test at \(\varepsilon=0.15\) it also clears
the \(\omega \geq 0.5\) replacement bar on human-written solutions
(\(\omega=\) 0.60), as does Opus-4.7 (\(\omega=\) 0.60), while the other
four judges fall short; at \(m=5\) reviewers this is a point estimate,
not yet statistically established (see
\hyperref[limitations]{Limitations}). Rank agreement across the six
judges is high: re-ranking the 12 systems under each cross-validation
judge yields system-level Spearman ρ of 0.92-0.97 against the primary
judge (Table~\ref{tbl-judge-swap}), so the RQ1-RQ2 ordering survives any
single judge's score bias. That the two passing judges come from
different providers, and agree on which blind reviewers they match,
suggests same-provider bias is not a large confound. Re-scoring the
leaderboard with Opus-4.7 as primary judge is a natural future
robustness check.

LLM judges are thus scalable raters alongside humans rather than
standalone ground truth. Where the judge still diverges, the
disagreement concentrates on presentation and on outcomes reached via
reasoning paths other than the reference solution's: agreement is
highest on the outcome and issue-identification dimensions, lowest on
formal correctness and terminology, and slips further on LLM-generated
solutions.

\section{Conclusion}\label{conclusion}

We introduce the BenGER dataset for subsumption-based legal reasoning in
German law. It combines two exam-style corpora (ZJS, Benchathon) with a
short doctrinal-reasoning corpus (Doctrinal Principles), a Benchathon
validation subset with human and human-AI baselines, a rubric-aligned
LLM-as-a-Judge, and a multi-rater cross-validation protocol. It
comprises 12 systems and 13514 generations across the three corpora, all
run end-to-end on our BenGER platform \citep{nagl2026benger}. BenGER is
released as an open evaluation framework with corpora, judge prompts,
rubric, and analysis pipeline. The methodology transfers to any
expert-graded free-text domain: treat grading as a distribution over
expert assessments, validate a judge by reviewer substitution, and swap
in the domain's own rubric.

\phantomsection

\section*{Limitations}\label{limitations}
\addcontentsline{toc}{section}{Limitations}

All LLM systems are evaluated through hosted provider APIs in a
black-box setting, so reported scores reflect end-to-end deployed
behaviour, not isolated base-model capability. We do not vary
provider-side prompts, decoding configurations, retrieval, or tool use,
and make no claim about the contribution of individual architectural
choices.

The dimension rubric and the LLM-judge prompt were co-developed for this
benchmark, not adopted from independent prior work. We validate them as
a pair against the multi-rater human-grading layer (RQ5). Its
within-annotation inter-rater spread (5.2 grade points) sits within the
range Hufeld \citep{hufeld2024} documents for German legal grading.
Neither component has been validated in isolation.

The Benchathon subset comprises 15 tasks and a modest participant pool.
This suffices for the per-task, per-system comparisons and the judge
validation reported here. But it limits statistical power for
fine-grained interaction effects, such as expertise-stratified
co-creation analyses.

Judge bias is a known confounder in LLM evaluation
\citep{zheng2023mtbench, panickssery2024self}. We address it via the
multi-rater agreement protocol on the Benchathon validation subset and
the inter-judge agreement analysis (RQ5): the six judges differ in how
strictly they score, but their pairwise rank agreement is comparable to
the agreement among the human reviewers.

Same-provider preference might occur, but its plausible size is limited,
for four reasons. First, re-ranking the systems under every
cross-validation judge preserves the ordering across providers
(system-level ρ ≥ 0.92, Table~\ref{tbl-judge-swap}). Second, judge-human
agreement is \emph{lower} on the LLM-generated half of the validation
subset (Pearson r=0.64 vs.~r=0.76), the opposite of what same-family
inflation would produce. Third, the Calderon winning rate drops from
\(\omega=\) 0.60 on human-written solutions to 0.00 on LLM-generated
ones, the same direction. Fourth, that drop appears for all six judges
tested, closed and open-weight, including judges whose model families
lead the system rankings, so the pattern is broad across judge models
rather than aligned with any one family
(Tables~\ref{tbl-judge-calibration},~\ref{tbl-calderon-per-annotator-human},~\ref{tbl-calderon-per-annotator-llm}).

Our per-annotator sample sizes (\(n_j \in [7, 21]\) scored solutions per
blind reviewer) are well below the 50-100 Calderon recommends. Every
test therefore fell back to Wilcoxon, with reduced power. At this sample
size the two-judge alt-test pass is a point estimate, not a robust
result. The winning rate moves in steps of \(0.20\) at \(m=5\)
reviewers. The 95\% CI on the primary judge's \(\omega\) spans {[}0.15,
0.95{]}. No bootstrap lower bound clears \(\omega=0.5\) for any judge
(Appendix~\ref{sec-judge-calibration-supplementary-detail}). The
two-judge pass should therefore be read as a promising direction, not an
established result.

Absolute RQ1-RQ2 leaderboard scores carry the GPT-5.4-mini calibration
offset (Table~\ref{tbl-judge-calibration}, measured against the
blind-reviewer pool mean). They should be read as a calibrated ordering,
not human-pool estimates. Calderon's FAQ recommends computing \(\omega\)
per domain, but splitting by legal area would leave each blind reviewer
only 3-7 scored solutions per area (civil, public, and criminal law), so
we pool across the three areas.

The rubric is grounded in German doctrinal grading conventions and is
not designed to generalise to common-law or non-German civil-law
systems. Metrics and aggregate scores in the BenGER dataset should not
be transferred to other jurisdictions.

ZJS material is publicly indexed, so provider models may have
encountered some of it during pretraining; we have not run an explicit
memorisation probe. The Benchathon and Doctrinal Principles corpora were
unpublished at the time of all generation runs. They are not a clean
contamination control: their orderings agreeing with the ZJS leaderboard
is consistent both with low contamination and with contamination that
scales proportionally across systems.

Per-area CIs (RQ4), pairwise inter-judge correlations (RQ5), and
error-analysis patterns are reported descriptively. We draw no
inferential claims from individual cell-level comparisons.

The Benchathon and ZJS exam-style corpora are graded with the
10-dimension case-solution (\emph{Falllösung}) rubric. The short-form
Doctrinal Principles corpus uses a 4-dimension rubric (result
correctness, legal knowledge, subsumption, clarity). Per-dimension
comparisons across corpora are therefore not meaningful. Only aggregate
raw scores, pass rates, and yes/no accuracy (Doctrinal Principles only)
should be compared across the three. The Google efficiency-tier slot
additionally mixes two Google models across corpora; Appendix
\ref{evaluated-systems-catalogue} details the change and how to read
cross-corpus comparisons for it.

\phantomsection

\section*{Ethical considerations}\label{ethical-considerations}
\addcontentsline{toc}{section}{Ethical considerations}

Benchathon participants were recruited from a pool of law students and
recent graduates who consented to the use of their written solutions and
self-reported expertise data for research purposes. Solutions and
self-reports are anonymized in the released BenGER dataset. Human
grading is conducted by domain experts under the same consent regime. We
release dimension-level rubric scores and aggregate statistics, not
free-text feedback that could re-identify a participant. The Benchathon
study design for the human and co-creation baseline was approved by an
ethics committee. The LLM-as-a-Judge methodology in BenGER is intended
for research-scale evaluation of legal-reasoning systems. It is not
validated for use as a graded assessment instrument in legal education,
or for deployment in legal practice.

The ZJS exams and reference solutions are copyright-protected. We have
obtained re-publishing consent for the vast majority of items, bundled
directly with the released dataset. For the remainder we provide URL
pointers to publicly accessible free locations on the web. All model
generations and LLM-judge evaluation scores are released under a
Creative Commons Attribution 4.0 (CC BY 4.0) licence. The benchmark is
therefore end-to-end reproducible: recomputing any reported number
requires only a short fetch step for the ZJS items without re-publishing
consent.

A public legal-reasoning benchmark can be optimised against. Future
systems may improve on the benchmark without commensurate gains in real
legal-reasoning quality. To mitigate this, we release the rubric and
judge prompts alongside the data, encourage reporting on the full
per-dimension breakdown rather than a single headline score, and treat
the benchmark as one signal among several in the evaluation of
legal-domain LLMs.

\section*{AI usage}\label{ai-usage}
\addcontentsline{toc}{section}{AI usage}

The authors used Anthropic Claude Opus 4.7, OpenAI GPT-5.5 Pro, and
Anthropic Claude Fable 5 (camera-ready revision) for coding, drafting
text from code and outlines, and writing assistance throughout. AI
agents implemented parts of the analysis code; the authors checked the
implementations and verified all reported results. Generated text was
checked, edited, and approved by the authors. All citations were
verified against their sources. The authors take full responsibility for
the paper's contents.

\section*{Acknowledgments}\label{acknowledgments}
\addcontentsline{toc}{section}{Acknowledgments}

We thank our colleagues in the TITAN project group - Philipp Reuß
(University of Göttingen), Sarah Rachut (TU Braunschweig), Liane Wörner
(University of Konstanz), Dominik Brodowski (Saarland University), and
Markus Langer and Asya Caroei (both University of Freiburg) - for the
discussions that shaped this work. We are grateful to the Legal Tech
Verband (LTV) and its members Clemens Hufeld (Beck/Noxtua), Daniella
Domokos (Capgemini), Ben Kühnel (Forvis Mazars), and Katharina Hertel
(Recode Law) for their support in organising the Benchathon, and to all
Benchathon participants for their dedicated work. We thank Markus Wagner
and the \emph{Zeitschrift für das Juristische Studium} (ZJS) author
community for generously permitting use and redistribution of their
cases. Finally, we thank Peter Moser (LMU) for connecting us with the
right experts and materials.

BenGER is generously funded by the Daimler Benz Foundation as part of
project TITAN (\emph{Technologische Intelligenz zur Transformation,
Automatisierung und Nutzerorientierung des Justizsystems}) and by the
German Federal Ministry of Justice and Consumer Protection as part of
the \emph{Digitalisierungsinitiative des Bundes für die Justiz} under
the GSJ project (\emph{Generatives Sprachmodell der Justiz}), a
collaboration between the justice ministries of North Rhine-Westphalia
and Bavaria, TU Munich, and the University of Cologne.

\bibliography{references.bib}

\begin{thebibliography}{36}
\providecommand{\natexlab}[1]{#1}

\bibitem[{Akarajaradwong et~al.(2025)Akarajaradwong, Pothavorn, Chaksangchaichot, Tasawong, Nopparatbundit, Pratai, and Nutanong}]{akarajaradwong2025nitibench}
Pawitsapak Akarajaradwong, Pirat Pothavorn, Chompakorn Chaksangchaichot, Panuthep Tasawong, Thitiwat Nopparatbundit, Keerakiat Pratai, and Sarana Nutanong. 2025.
\newblock \href {https://doi.org/10.18653/v1/2025.emnlp-main.1739} {{NitiBench}: Benchmarking {LLM} frameworks on {Thai} legal question answering capabilities}.
\newblock In \emph{Proceedings of the 2025 Conference on Empirical Methods in Natural Language Processing}, pages 34304--34327, Suzhou, China. Association for Computational Linguistics.

\bibitem[{Banerjee and Lavie(2005)}]{banerjee2005meteor}
Satanjeev Banerjee and Alon Lavie. 2005.
\newblock \href {https://aclanthology.org/W05-0909/} {{METEOR}: An automatic metric for {MT} evaluation with improved correlation with human judgments}.
\newblock In \emph{Proceedings of the ACL Workshop on Intrinsic and Extrinsic Evaluation Measures for Machine Translation and/or Summarization}, pages 65--72, Ann Arbor, Michigan. Association for Computational Linguistics.

\bibitem[{{Bundesministerium der Justiz}(1981)}]{jurprnotskv1981}
{Bundesministerium der Justiz}. 1981.
\newblock \href {https://www.gesetze-im-internet.de/jurprnotskv/} {Verordnung {\"u}ber eine {Noten-} und {Punkteskala} f{\"u}r die erste und zweite juristische {Pr{\"u}fung} ({JurPrNotSkV})}.
\newblock BGBl.~I S.~1243.
\newblock Federal grading ordinance for the two German state law examinations.

\bibitem[{Butler and Butler(2026)}]{butler2026legalragbench}
Abdur-Rahman Butler and Umar Butler. 2026.
\newblock \href {https://arxiv.org/abs/2603.01710} {{Legal RAG Bench}: An end-to-end benchmark for legal {RAG}}.
\newblock \emph{Preprint}, arXiv:2603.01710.

\bibitem[{B{\"u}ttner and Habernal(2024)}]{buttner2024gerlayqa}
Marius B{\"u}ttner and Ivan Habernal. 2024.
\newblock \href {https://doi.org/10.18653/v1/2024.eacl-long.122} {Answering legal questions from laymen in {German} civil law system}.
\newblock In \emph{Proceedings of the 18th Conference of the European Chapter of the Association for Computational Linguistics (Volume 1: Long Papers)}, pages 2015--2027, St. Julian's, Malta. Association for Computational Linguistics.

\bibitem[{Calderon et~al.(2025)Calderon, Reichart, and Dror}]{calderon2025}
Nitay Calderon, Roi Reichart, and Rotem Dror. 2025.
\newblock \href {https://doi.org/10.18653/v1/2025.acl-long.782} {The alternative annotator test for {LLM}-as-a-judge: How to statistically justify replacing human annotators with {LLMs}}.
\newblock In \emph{Proceedings of the 63rd Annual Meeting of the Association for Computational Linguistics (Volume 1: Long Papers)}, pages 16051--16081, Vienna, Austria. Association for Computational Linguistics.

\bibitem[{Chalkidis et~al.(2022)Chalkidis, Jana, Hartung, Bommarito, Androutsopoulos, Katz, and Aletras}]{chalkidis2022lexglue}
Ilias Chalkidis, Abhik Jana, Dirk Hartung, Michael Bommarito, Ion Androutsopoulos, Daniel Katz, and Nikolaos Aletras. 2022.
\newblock \href {https://doi.org/10.18653/v1/2022.acl-long.297} {{LexGLUE}: A benchmark dataset for legal language understanding in {English}}.
\newblock In \emph{Proceedings of the 60th Annual Meeting of the Association for Computational Linguistics (Volume 1: Long Papers)}, pages 4310--4330, Dublin, Ireland. Association for Computational Linguistics.

\bibitem[{Chlapanis et~al.(2025)Chlapanis, Galanis, Aletras, and Androutsopoulos}]{chlapanis2025greekbarbench}
Odysseas~S. Chlapanis, Dimitrios Galanis, Nikolaos Aletras, and Ion Androutsopoulos. 2025.
\newblock \href {https://doi.org/10.18653/v1/2025.findings-emnlp.1368} {{GreekBarBench}: A challenging benchmark for free-text legal reasoning and citations}.
\newblock In \emph{Findings of the Association for Computational Linguistics: EMNLP 2025}, pages 25099--25119, Suzhou, China. Association for Computational Linguistics.

\bibitem[{Chlapanis et~al.(2024)Chlapanis, Galanis, and Androutsopoulos}]{chlapanis2024larechr}
Odysseas~S. Chlapanis, Dimitrios Galanis, and Ion Androutsopoulos. 2024.
\newblock \href {https://doi.org/10.18653/v1/2024.nllp-1.22} {{LAR-ECHR}: A new legal argument reasoning task and dataset for cases of the {European Court of Human Rights}}.
\newblock In \emph{Proceedings of the Natural Legal Language Processing Workshop 2024}, pages 267--279, Miami, FL, USA. Association for Computational Linguistics.

\bibitem[{Choudhury et~al.(2026)Choudhury, Chandramouli, Anand, and Gupta}]{choudhury2026clause}
Manan~Roy Choudhury, Adithya Chandramouli, Mannan Anand, and Vivek Gupta. 2026.
\newblock \href {https://doi.org/10.18653/v1/2026.findings-eacl.305} {Better call {CLAUSE}: A discrepancy benchmark for auditing {LLMs} legal reasoning capabilities}.
\newblock In \emph{Findings of the Association for Computational Linguistics: EACL 2026}, pages 5776--5818, Rabat, Morocco. Association for Computational Linguistics.

\bibitem[{Cohen(1960)}]{cohen1960coefficient}
Jacob Cohen. 1960.
\newblock \href {https://doi.org/10.1177/001316446002000104} {A coefficient of agreement for nominal scales}.
\newblock \emph{Educational and Psychological Measurement}, 20(1):37--46.

\bibitem[{Fan et~al.(2026)Fan, Ni, Merane, Tian, Hermstr{\"u}wer, Huang, Akhtar, Salimbeni, Geering, Dreyer, Brunner, Leippold, Sachan, Stremitzer, Engel, Ash, and Niklaus}]{fan2026lexam}
Yu~Fan, Jingwei Ni, Jakob Merane, Yang Tian, Yoan Hermstr{\"u}wer, Yinya Huang, Mubashara Akhtar, Etienne Salimbeni, Florian Geering, Oliver Dreyer, Daniel Brunner, Markus Leippold, Mrinmaya Sachan, Alexander Stremitzer, Christoph Engel, Elliott Ash, and Joel Niklaus. 2026.
\newblock \href {https://openreview.net/forum?id=xNhbMyXsJn} {{LEXam}: Benchmarking legal reasoning on 340 law exams}.
\newblock In \emph{Proceedings of the Fourteenth International Conference on Learning Representations}.
\newblock ArXiv:2505.12864.

\bibitem[{Fei et~al.(2024)Fei, Shen, Zhu, Zhou, Han, Huang, Zhang, Chen, Yin, Shen, Ge, and Ng}]{fei2024lawbench}
Zhiwei Fei, Xiaoyu Shen, Dawei Zhu, Fengzhe Zhou, Zhuo Han, Alan Huang, Songyang Zhang, Kai Chen, Zhixin Yin, Zongwen Shen, Jidong Ge, and Vincent Ng. 2024.
\newblock \href {https://doi.org/10.18653/v1/2024.emnlp-main.452} {{LawBench}: Benchmarking legal knowledge of large language models}.
\newblock In \emph{Proceedings of the 2024 Conference on Empirical Methods in Natural Language Processing}, pages 7933--7962, Miami, Florida, USA. Association for Computational Linguistics.

\bibitem[{Guha et~al.(2023)Guha, Nyarko, Ho, R{\'e}, Chilton, Narayana, Chohlas-Wood, Peters, Waldon, Rockmore, Zambrano, Talisman, Hoque, Surani, Fagan, Sarfaty, Dickinson, Porat, Hegland, Wu, Nudell, Niklaus, Nay, Choi, Tobia, Hagan, Ma, Livermore, Rasumov-Rahe, Holzenberger, Kolt, Henderson, Rehaag, Goel, Gao, Williams, Gandhi, Zur, Iyer, and Li}]{guha2023legalbench}
Neel Guha, Julian Nyarko, Daniel~E. Ho, Christopher R{\'e}, Adam Chilton, Aditya Narayana, Alex Chohlas-Wood, Austin Peters, Brandon Waldon, Daniel~N. Rockmore, Diego Zambrano, Dmitry Talisman, Enam Hoque, Faiz Surani, Frank Fagan, Galit Sarfaty, Gregory~M. Dickinson, Haggai Porat, Jason Hegland, and 21 others. 2023.
\newblock \href {https://doi.org/10.52202/075280-1915} {{LegalBench}: A collaboratively built benchmark for measuring legal reasoning in large language models}.
\newblock In \emph{Advances in Neural Information Processing Systems}, volume~36, pages 44123--44279. Curran Associates, Inc.

\bibitem[{Hariri and Ho(2025)}]{hariri2025laborbench}
Emaan Hariri and Daniel~E. Ho. 2025.
\newblock \href {https://doi.org/10.1145/3769126.3769256} {{AI} for statutory simplification: A comprehensive state legal corpus and labor benchmark}.
\newblock In \emph{Proceedings of the Twentieth International Conference on Artificial Intelligence and Law}, pages 177--187, Chicago, IL, USA. ACM.
\newblock ArXiv:2508.19365.

\bibitem[{Hufeld(2024)}]{hufeld2024}
Clemens Hufeld. 2024.
\newblock \href {https://doi.org/10.5771/2196-7261-2024-1-59} {{Jede Korrektur eine andere Note: Quantitative Untersuchung der Objektivit\"at juristischer Klausurbewertungen}}.
\newblock \emph{Zeitschrift f\"ur Didaktik der Rechtswissenschaft}, 11(1):59--83.

\bibitem[{Jia et~al.(2026)Jia, Yue, Chen, Wang, Liu, Li, Song, and Wei}]{jia2025readyjuristone}
Zheng Jia, Shengbin Yue, Wei Chen, Siyuan Wang, Yidong Liu, Zejun Li, Yun Song, and Zhongyu Wei. 2026.
\newblock \href {https://doi.org/10.18653/v1/2026.acl-long.471} {Ready jurist one: Benchmarking language agents for legal intelligence in dynamic environments}.
\newblock In \emph{Proceedings of the 64th Annual Meeting of the Association for Computational Linguistics (Volume 1: Long Papers)}, pages 10351--10376. Association for Computational Linguistics.

\bibitem[{Kim et~al.(2024)Kim, Choi, Choi, Choi, Park, and Hwang}]{kim2024kbl}
Yeeun Kim, Youngrok Choi, Eunkyung Choi, JinHwan Choi, Hai~Jin Park, and Wonseok Hwang. 2024.
\newblock \href {https://doi.org/10.18653/v1/2024.findings-emnlp.319} {Developing a pragmatic benchmark for assessing {Korean} legal language understanding in large language models}.
\newblock In \emph{Findings of the Association for Computational Linguistics: EMNLP 2024}, pages 5573--5595, Miami, Florida, USA. Association for Computational Linguistics.

\bibitem[{Koo and Li(2016)}]{koo2016guideline}
Terry~K. Koo and Mae~Y. Li. 2016.
\newblock \href {https://doi.org/10.1016/j.jcm.2016.02.012} {A guideline of selecting and reporting intraclass correlation coefficients for reliability research}.
\newblock \emph{Journal of Chiropractic Medicine}, 15(2):155--163.

\bibitem[{Larenz and Canaris(1995)}]{larenz1995methodenlehre}
Karl Larenz and Claus-Wilhelm Canaris. 1995.
\newblock \href {https://doi.org/10.1007/978-3-662-08709-1} {\emph{Methodenlehre der Rechtswissenschaft}}, 3 edition.
\newblock Springer, Berlin, Heidelberg.

\bibitem[{Lee et~al.(2025)Lee, Kim, Hwang, Kim, and Lee}]{lee2025koblex}
Jihyung Lee, Daehui Kim, Seonjeong Hwang, Hyounghun Kim, and Gary Lee. 2025.
\newblock \href {https://doi.org/10.18653/v1/2025.emnlp-main.200} {{KoBLEX}: Open legal question answering with multi-hop reasoning}.
\newblock In \emph{Proceedings of the 2025 Conference on Empirical Methods in Natural Language Processing}, pages 4019--4053, Suzhou, China. Association for Computational Linguistics.

\bibitem[{Li et~al.(2024)Li, Chen, Ai, Wu, Zhang, and Liu}]{li2024lexeval}
Haitao Li, You Chen, Qingyao Ai, Yueyue Wu, Ruizhe Zhang, and Yiqun Liu. 2024.
\newblock \href {https://doi.org/10.52202/079017-0790} {{LexEval}: A comprehensive {Chinese} legal benchmark for evaluating large language models}.
\newblock In \emph{Advances in Neural Information Processing Systems}, volume~37, pages 25061--25094. Curran Associates, Inc.

\bibitem[{Lin(2004)}]{lin2004rouge}
Chin-Yew Lin. 2004.
\newblock \href {https://aclanthology.org/W04-1013/} {{ROUGE}: A package for automatic evaluation of summaries}.
\newblock In \emph{Text Summarization Branches Out}, pages 74--81, Barcelona, Spain. Association for Computational Linguistics.

\bibitem[{M{\"o}llers(2023)}]{moellers2023methodenlehre}
Thomas M.~J. M{\"o}llers. 2023.
\newblock \emph{Juristische Methodenlehre}, 5 edition.
\newblock C.H. Beck, M{\"u}nchen.

\bibitem[{Nagl and Grabmair(2026)}]{nagl2026benger}
Sebastian Nagl and Matthias Grabmair. 2026.
\newblock {BenGER Platform}: A collaborative web platform for end-to-end benchmarking of {German} legal tasks.
\newblock In \emph{Proceedings of the Twenty-First International Conference on Artificial Intelligence and Law}, Singapore. ACM.
\newblock To appear. arXiv:2604.13583.

\bibitem[{Panickssery et~al.(2024)Panickssery, Bowman, and Feng}]{panickssery2024self}
Arjun Panickssery, Samuel~R. Bowman, and Shi Feng. 2024.
\newblock \href {https://doi.org/10.52202/079017-2197} {{LLM} evaluators recognize and favor their own generations}.
\newblock In \emph{Advances in Neural Information Processing Systems}, volume~37, pages 68772--68802. Curran Associates, Inc.

\bibitem[{Papineni et~al.(2002)Papineni, Roukos, Ward, and Zhu}]{papineni2002bleu}
Kishore Papineni, Salim Roukos, Todd Ward, and Wei-Jing Zhu. 2002.
\newblock \href {https://doi.org/10.3115/1073083.1073135} {{B}leu: A method for automatic evaluation of machine translation}.
\newblock In \emph{Proceedings of the 40th Annual Meeting of the Association for Computational Linguistics}, pages 311--318. Association for Computational Linguistics.

\bibitem[{Pipitone and Houir~Alami(2024)}]{pipitone2024legalbenchrag}
Nicholas Pipitone and Ghita Houir~Alami. 2024.
\newblock \href {https://arxiv.org/abs/2408.10343} {{LegalBench-RAG}: A benchmark for retrieval-augmented generation in the legal domain}.
\newblock \emph{Preprint}, arXiv:2408.10343.

\bibitem[{Popovi{\'c}(2015)}]{popovic2015chrf}
Maja Popovi{\'c}. 2015.
\newblock \href {https://doi.org/10.18653/v1/W15-3049} {{chrF}: Character n-gram {F}-score for automatic {MT} evaluation}.
\newblock In \emph{Proceedings of the Tenth Workshop on Statistical Machine Translation}, pages 392--395, Lisbon, Portugal. Association for Computational Linguistics.

\bibitem[{Reimers and Gurevych(2019)}]{reimers2019sentencebert}
Nils Reimers and Iryna Gurevych. 2019.
\newblock \href {https://doi.org/10.18653/v1/D19-1410} {{Sentence-BERT}: Sentence embeddings using {Siamese BERT-Networks}}.
\newblock In \emph{Proceedings of the 2019 Conference on Empirical Methods in Natural Language Processing and the 9th International Joint Conference on Natural Language Processing (EMNLP-IJCNLP)}, pages 3982--3992. Association for Computational Linguistics.

\bibitem[{Shi et~al.(2026)Shi, Liu, Hu, Song, Xu, Ma, Tang, Zhang, Chen, Feng, Lv, Wu, Yang, Yang, Wang, Shi, Qiu, Qi, Zhang, Sui, Chen, Zhang, Yang, Yu, Liu, Lin, Shen, Zhao, Clarke, and Wei}]{shi2026plawbench}
Yuzhen Shi, Huanghai Liu, Yiran Hu, Gaojie Song, Xinran Xu, Yubo Ma, Tianyi Tang, Li~Zhang, Qingjing Chen, Di~Feng, Wenbo Lv, Weiheng Wu, Kexin Yang, Sen Yang, Wei Wang, Rongyao Shi, Yuanyang Qiu, Yuemeng Qi, Jingwen Zhang, and 11 others. 2026.
\newblock \href {https://doi.org/10.18653/v1/2026.acl-long.458} {{PLawBench}: A rubric-based benchmark for evaluating {LLMs} in real-world legal practice}.
\newblock In \emph{Proceedings of the 64th Annual Meeting of the Association for Computational Linguistics (Volume 1: Long Papers)}, pages 10067--10116. Association for Computational Linguistics.

\bibitem[{Shrout and Fleiss(1979)}]{shrout1979intraclass}
Patrick~E. Shrout and Joseph~L. Fleiss. 1979.
\newblock \href {https://doi.org/10.1037/0033-2909.86.2.420} {Intraclass correlations: Uses in assessing rater reliability}.
\newblock \emph{Psychological Bulletin}, 86(2):420--428.

\bibitem[{Weber et~al.(2025)Weber, Paritala, Rechu, Feddoul, Bonagiri, Klewer, Karg, Unger, Mauch, and K{\"o}nig-Ries}]{weber2025gerlerb}
Malte Weber, Balaramakrishna Paritala, Abhilash~Reddy Rechu, Leila Feddoul, Suresh~Kumar Bonagiri, Norman Klewer, Pirmin~Mathias Karg, Christoph Unger, Marianne Mauch, and Birgitta K{\"o}nig-Ries. 2025.
\newblock \href {https://doi.org/10.18420/rvi2025-112} {{GerLeRB} -- {German} legislative retrieval benchmark}.
\newblock In \emph{Fachtagung Rechts- und Verwaltungsinformatik (RVI 2025)}, Recht und Informatik, pages 157--168, Bonn. Gesellschaft f{\"u}r Informatik.

\bibitem[{Zhang et~al.(2020)Zhang, Kishore, Wu, Weinberger, and Artzi}]{zhang2020bertscore}
Tianyi Zhang, Varsha Kishore, Felix Wu, Kilian~Q. Weinberger, and Yoav Artzi. 2020.
\newblock \href {https://openreview.net/forum?id=SkeHuCVFDr} {{BERTScore}: Evaluating text generation with {BERT}}.
\newblock In \emph{International Conference on Learning Representations}.
\newblock ArXiv:1904.09675.

\bibitem[{Zhao et~al.(2019)Zhao, Peyrard, Liu, Gao, Meyer, and Eger}]{zhao2019moverscore}
Wei Zhao, Maxime Peyrard, Fei Liu, Yang Gao, Christian~M. Meyer, and Steffen Eger. 2019.
\newblock \href {https://doi.org/10.18653/v1/D19-1053} {{MoverScore}: Text generation evaluating with contextualized embeddings and earth mover distance}.
\newblock In \emph{Proceedings of the 2019 Conference on Empirical Methods in Natural Language Processing and the 9th International Joint Conference on Natural Language Processing (EMNLP-IJCNLP)}, pages 563--578. Association for Computational Linguistics.

\bibitem[{Zheng et~al.(2023)Zheng, Chiang, Sheng, Zhuang, Wu, Zhuang, Lin, Li, Li, Xing, Zhang, Gonzalez, and Stoica}]{zheng2023mtbench}
Lianmin Zheng, Wei-Lin Chiang, Ying Sheng, Siyuan Zhuang, Zhanghao Wu, Yonghao Zhuang, Zi~Lin, Zhuohan Li, Dacheng Li, Eric Xing, Hao Zhang, Joseph~E. Gonzalez, and Ion Stoica. 2023.
\newblock \href {https://doi.org/10.52202/075280-2020} {Judging {LLM-as-a-Judge} with {MT-Bench} and {Chatbot Arena}}.
\newblock In \emph{Advances in Neural Information Processing Systems}, volume~36, pages 46595--46623. Curran Associates, Inc.

\end{thebibliography}
\appendix
\nolinenumbers

\section*{Contents of the Appendix}
\noindent
A\quad Full results table \dotfill \pageref{full-results-table}\par
B\quad Evaluated systems catalogue \dotfill \pageref{evaluated-systems-catalogue}\par
C\quad Operational metadata and reproducibility \dotfill \pageref{operational-metadata-and-reproducibility}\par
D\quad Per-task per-system heatmap \dotfill \pageref{per-task-per-system-heatmap}\par
E\quad Benchathon participant composition \dotfill \pageref{benchathon-participant-composition}\par
F\quad Full Benchathon human-evaluation assignment procedure \dotfill \pageref{full-benchathon-human-evaluation-assignment-procedure}\par
G\quad RQ4 supplementary detail \dotfill \pageref{rq4-supplementary}\par
H\quad Automatic-metric vs.\ LLM-judge correlation \dotfill \pageref{automatic-metric-vs.-llm-judge-correlation}\par
I\quad Judge-human agreement statistics \dotfill \pageref{judge-human-agreement-statistics}\par
J\quad System prompts for generation \dotfill \pageref{system-prompts-for-generation}\par
K\quad Instruction prompts for generation \dotfill \pageref{instruction-prompts-for-generation}\par
L\quad Evaluation prompts for the LLM judge \dotfill \pageref{evaluation-prompts-for-the-llm-judge}\par
M\quad LLM-judge configuration \dotfill \pageref{llm-judge-configuration}\par
N\quad Judge calibration - supplementary detail \dotfill \pageref{sec-judge-calibration-supplementary-detail}\par
O\quad Error-analysis mode counts \dotfill \pageref{error-analysis-mode-counts}\par
P\quad Reasoning-mode ablation \dotfill \pageref{reasoning-mode-ablation}\par
Q\quad Glossary of German terms \dotfill \pageref{glossary-of-german-terms}\par

\section{Full results table}\label{full-results-table}

The full per-system leaderboards are split into six tables, three
corpora \(\times\) two metric groups: rubric-based metrics (LLM-judge
raw / grade-points / pass rate; plus human-grader rubric columns on
Benchathon) in
Tables~\ref{tbl-full-results-rubric-benchathon},~\ref{tbl-full-results-rubric-zjs},~\ref{tbl-full-results-rubric-grundpr},
and surface / embedding-based automatic metrics (plus Doctrinal
Principles yes/no decision accuracy, a task-correctness measure rather
than a rubric one) in
Tables~\ref{tbl-full-results-auto-benchathon},~\ref{tbl-full-results-auto-zjs},~\ref{tbl-full-results-auto-grundpr}.
The two Benchathon human-baseline rows (traditional, co-creation) appear
at the top of the Benchathon tables. The CI is the percentile interval
over \(B=2000\) resamples on the per-generation values (seed \(42\));
each cell shows the mean stacked above \(\pm\) half the width of the
95\% CI (in a smaller font). Empty cells indicate the metric is not
computed on that dataset.

\begin{table*}[t]
\centering\footnotesize
\caption{Rubric-based results on the \textbf{Benchathon} corpus: per-system bootstrap 95\% CIs ($B=2000$, seed $42$) on the LLM-judge raw / grade-points / pass rate columns plus the parallel human-grader rubric columns (\texttt{H.}-prefixed). Each cell shows the mean stacked above $\pm$ half the width of the 95\% CI. $n$ is the per-system generation count; $n_H$ is the per-system human-grader row count. The two human-baseline rows (traditional, co-creation) sit at the top.}
\label{tbl-full-results-rubric-benchathon}
\setlength{\tabcolsep}{4pt}
\begin{tabular*}{\textwidth}{@{\extracolsep{\fill}}lrrrrrrrr@{}}
\toprule
System & $n$ & $n_H$ & Judge raw & Gr.\,pts & Pass & H.\,raw & H.\,Gr.pts & H.\,pass \\
\midrule
\textit{Human (traditional)} & 65 & 60 & \shortstack[r]{50.0\\{\scriptsize $\pm$3.7}} & \shortstack[r]{4.72\\{\scriptsize $\pm$0.76}} & \shortstack[r]{52.3\%\\{\scriptsize $\pm$11.5}} & \shortstack[r]{50.7\\{\scriptsize $\pm$5.6}} & \shortstack[r]{5.67\\{\scriptsize $\pm$1.04}} & \shortstack[r]{63.3\%\\{\scriptsize $\pm$11.6}} \\
\textit{Human (co-creation)} & 155 & 60 & \shortstack[r]{65.7\\{\scriptsize $\pm$2.2}} & \shortstack[r]{8.50\\{\scriptsize $\pm$0.58}} & \shortstack[r]{87.7\%\\{\scriptsize $\pm$5.2}} & \shortstack[r]{65.4\\{\scriptsize $\pm$5.7}} & \shortstack[r]{9.08\\{\scriptsize $\pm$1.25}} & \shortstack[r]{86.7\%\\{\scriptsize $\pm$8.4}} \\
Opus-4.7 & 15 & 8 & \shortstack[r]{69.3\\{\scriptsize $\pm$5.7}} & \shortstack[r]{9.47\\{\scriptsize $\pm$1.80}} & \shortstack[r]{93.3\%\\{\scriptsize $\pm$10.0}} & \shortstack[r]{62.7\\{\scriptsize $\pm$7.2}} & \shortstack[r]{7.62\\{\scriptsize $\pm$1.75}} & \shortstack[r]{87.5\%\\{\scriptsize $\pm$18.8}} \\
Gemini-3.1-Pro & 15 & 8 & \shortstack[r]{68.3\\{\scriptsize $\pm$8.4}} & \shortstack[r]{9.60\\{\scriptsize $\pm$1.83}} & \shortstack[r]{93.3\%\\{\scriptsize $\pm$10.0}} & \shortstack[r]{76.2\\{\scriptsize $\pm$6.4}} & \shortstack[r]{11.62\\{\scriptsize $\pm$1.88}} & \shortstack[r]{100.0\%\\{\scriptsize $\pm$0.0}} \\
GPT-5.4 & 15 & 4 & \shortstack[r]{66.6\\{\scriptsize $\pm$4.8}} & \shortstack[r]{8.47\\{\scriptsize $\pm$1.43}} & \shortstack[r]{100.0\%\\{\scriptsize $\pm$0.0}} & \shortstack[r]{61.1\\{\scriptsize $\pm$16.3}} & \shortstack[r]{7.50\\{\scriptsize $\pm$4.25}} & \shortstack[r]{75.0\%\\{\scriptsize $\pm$37.5}} \\
Gemini-3.1-Flash-Lite & 15 & 4 & \shortstack[r]{63.5\\{\scriptsize $\pm$6.1}} & \shortstack[r]{7.73\\{\scriptsize $\pm$1.80}} & \shortstack[r]{86.7\%\\{\scriptsize $\pm$16.6}} & \shortstack[r]{67.0\\{\scriptsize $\pm$14.1}} & \shortstack[r]{9.25\\{\scriptsize $\pm$3.38}} & \shortstack[r]{75.0\%\\{\scriptsize $\pm$37.5}} \\
Sonnet-4.6 & 15 & 4 & \shortstack[r]{58.9\\{\scriptsize $\pm$6.0}} & \shortstack[r]{6.47\\{\scriptsize $\pm$1.66}} & \shortstack[r]{73.3\%\\{\scriptsize $\pm$23.3}} & \shortstack[r]{53.8\\{\scriptsize $\pm$6.5}} & \shortstack[r]{5.25\\{\scriptsize $\pm$1.75}} & \shortstack[r]{75.0\%\\{\scriptsize $\pm$37.5}} \\
GPT-5.4-mini & 15 & 4 & \shortstack[r]{58.6\\{\scriptsize $\pm$5.7}} & \shortstack[r]{6.27\\{\scriptsize $\pm$1.53}} & \shortstack[r]{73.3\%\\{\scriptsize $\pm$23.3}} & \shortstack[r]{54.6\\{\scriptsize $\pm$10.6}} & \shortstack[r]{5.50\\{\scriptsize $\pm$2.25}} & \shortstack[r]{75.0\%\\{\scriptsize $\pm$37.5}} \\
DeepSeek-V4-Pro & 15 & 4 & \shortstack[r]{56.1\\{\scriptsize $\pm$5.8}} & \shortstack[r]{5.73\\{\scriptsize $\pm$1.47}} & \shortstack[r]{66.7\%\\{\scriptsize $\pm$23.4}} & \shortstack[r]{77.6\\{\scriptsize $\pm$6.1}} & \shortstack[r]{11.75\\{\scriptsize $\pm$1.75}} & \shortstack[r]{100.0\%\\{\scriptsize $\pm$0.0}} \\
Qwen3.5-122B & 15 & 4 & \shortstack[r]{49.4\\{\scriptsize $\pm$4.9}} & \shortstack[r]{4.33\\{\scriptsize $\pm$1.07}} & \shortstack[r]{40.0\%\\{\scriptsize $\pm$23.4}} & \shortstack[r]{47.1\\{\scriptsize $\pm$6.9}} & \shortstack[r]{3.75\\{\scriptsize $\pm$1.12}} & \shortstack[r]{25.0\%\\{\scriptsize $\pm$37.5}} \\
DeepSeek-V4-Flash & 15 & 4 & \shortstack[r]{49.3\\{\scriptsize $\pm$3.6}} & \shortstack[r]{4.00\\{\scriptsize $\pm$0.73}} & \shortstack[r]{46.7\%\\{\scriptsize $\pm$26.6}} & \shortstack[r]{52.4\\{\scriptsize $\pm$16.8}} & \shortstack[r]{5.75\\{\scriptsize $\pm$3.38}} & \shortstack[r]{75.0\%\\{\scriptsize $\pm$37.5}} \\
Qwen3.6-35B & 15 & 0 & \shortstack[r]{42.4\\{\scriptsize $\pm$7.2}} & \shortstack[r]{3.60\\{\scriptsize $\pm$1.20}} & \shortstack[r]{26.7\%\\{\scriptsize $\pm$20.0}} & --- & --- & --- \\
Qwen3-235B & 15 & 8 & \shortstack[r]{41.9\\{\scriptsize $\pm$4.6}} & \shortstack[r]{3.00\\{\scriptsize $\pm$0.51}} & \shortstack[r]{20.0\%\\{\scriptsize $\pm$20.0}} & \shortstack[r]{25.5\\{\scriptsize $\pm$9.5}} & \shortstack[r]{1.50\\{\scriptsize $\pm$0.69}} & \shortstack[r]{0.0\%\\{\scriptsize $\pm$0.0}} \\
Llama-4 & 15 & 8 & \shortstack[r]{37.8\\{\scriptsize $\pm$4.4}} & \shortstack[r]{2.60\\{\scriptsize $\pm$0.40}} & \shortstack[r]{20.0\%\\{\scriptsize $\pm$20.0}} & \shortstack[r]{19.9\\{\scriptsize $\pm$5.1}} & \shortstack[r]{0.88\\{\scriptsize $\pm$0.50}} & \shortstack[r]{0.0\%\\{\scriptsize $\pm$0.0}} \\
\bottomrule
\end{tabular*}
\end{table*}

\begin{table*}[t]
\centering\footnotesize
\caption{Rubric-based results on the \textbf{ZJS} corpus: per-system bootstrap 95\% CIs ($B=2000$, seed $42$) on the LLM-judge raw, grade-points, and pass-rate columns. Each cell shows the mean stacked above $\pm$ half the width of the 95\% CI. $n$ is the per-system generation count.}
\label{tbl-full-results-rubric-zjs}
\setlength{\tabcolsep}{4pt}
\begin{tabular*}{\textwidth}{@{\extracolsep{\fill}}lrrrr@{}}
\toprule
System & $n$ & Judge raw & Gr.\,pts & Pass \\
\midrule
Gemini-3.1-Pro & 579 & \shortstack[r]{61.7\\{\scriptsize $\pm$1.0}} & \shortstack[r]{7.33\\{\scriptsize $\pm$0.27}} & \shortstack[r]{83.2\%\\{\scriptsize $\pm$3.1}} \\
GPT-5.4 & 581 & \shortstack[r]{60.4\\{\scriptsize $\pm$1.1}} & \shortstack[r]{6.99\\{\scriptsize $\pm$0.28}} & \shortstack[r]{78.5\%\\{\scriptsize $\pm$3.5}} \\
Opus-4.7 & 581 & \shortstack[r]{58.7\\{\scriptsize $\pm$1.0}} & \shortstack[r]{6.54\\{\scriptsize $\pm$0.27}} & \shortstack[r]{75.7\%\\{\scriptsize $\pm$3.5}} \\
Sonnet-4.6 & 581 & \shortstack[r]{51.6\\{\scriptsize $\pm$1.0}} & \shortstack[r]{4.87\\{\scriptsize $\pm$0.23}} & \shortstack[r]{50.4\%\\{\scriptsize $\pm$4.2}} \\
DeepSeek-V4-Pro & 581 & \shortstack[r]{50.0\\{\scriptsize $\pm$1.0}} & \shortstack[r]{4.53\\{\scriptsize $\pm$0.23}} & \shortstack[r]{44.6\%\\{\scriptsize $\pm$4.0}} \\
GPT-5.4-mini & 581 & \shortstack[r]{49.7\\{\scriptsize $\pm$1.0}} & \shortstack[r]{4.43\\{\scriptsize $\pm$0.21}} & \shortstack[r]{46.1\%\\{\scriptsize $\pm$4.1}} \\
DeepSeek-V4-Flash & 581 & \shortstack[r]{46.0\\{\scriptsize $\pm$0.8}} & \shortstack[r]{3.68\\{\scriptsize $\pm$0.17}} & \shortstack[r]{30.1\%\\{\scriptsize $\pm$3.6}} \\
Gemini-3.1-Flash-Lite & 579 & \shortstack[r]{44.6\\{\scriptsize $\pm$0.8}} & \shortstack[r]{3.48\\{\scriptsize $\pm$0.16}} & \shortstack[r]{26.3\%\\{\scriptsize $\pm$3.5}} \\
Qwen3.5-122B & 581 & \shortstack[r]{44.2\\{\scriptsize $\pm$0.8}} & \shortstack[r]{3.43\\{\scriptsize $\pm$0.15}} & \shortstack[r]{25.6\%\\{\scriptsize $\pm$3.7}} \\
Qwen3.6-35B & 581 & \shortstack[r]{39.8\\{\scriptsize $\pm$0.9}} & \shortstack[r]{2.84\\{\scriptsize $\pm$0.12}} & \shortstack[r]{13.4\%\\{\scriptsize $\pm$2.8}} \\
Llama-4 & 581 & \shortstack[r]{36.2\\{\scriptsize $\pm$0.6}} & \shortstack[r]{2.36\\{\scriptsize $\pm$0.07}} & \shortstack[r]{4.6\%\\{\scriptsize $\pm$1.7}} \\
Qwen3-235B & 581 & \shortstack[r]{35.9\\{\scriptsize $\pm$0.7}} & \shortstack[r]{2.41\\{\scriptsize $\pm$0.09}} & \shortstack[r]{6.2\%\\{\scriptsize $\pm$1.9}} \\
\bottomrule
\end{tabular*}
\end{table*}

\begin{table*}[t]
\centering\footnotesize
\caption{Rubric-based results on the \textbf{Doctrinal Principles} corpus: per-system bootstrap 95\% CIs ($B=2000$, seed $42$) on the LLM-judge raw, grade-points, and pass-rate columns. Each cell shows the mean stacked above $\pm$ half the width of the 95\% CI. $n$ is the per-system generation count.}
\label{tbl-full-results-rubric-grundpr}
\setlength{\tabcolsep}{4pt}
\begin{tabular*}{\textwidth}{@{\extracolsep{\fill}}lrrrr@{}}
\toprule
System & $n$ & Judge raw & Gr.\,pts & Pass \\
\midrule
Gemini-3.1-Pro & 531 & \shortstack[r]{83.2\\{\scriptsize $\pm$1.8}} & --- & \shortstack[r]{87.0\%\\{\scriptsize $\pm$2.9}} \\
Opus-4.7 & 531 & \shortstack[r]{83.1\\{\scriptsize $\pm$1.9}} & --- & \shortstack[r]{85.9\%\\{\scriptsize $\pm$2.9}} \\
Sonnet-4.6 & 531 & \shortstack[r]{81.6\\{\scriptsize $\pm$2.0}} & --- & \shortstack[r]{82.9\%\\{\scriptsize $\pm$3.2}} \\
GPT-5.4 & 531 & \shortstack[r]{80.3\\{\scriptsize $\pm$2.1}} & --- & \shortstack[r]{81.5\%\\{\scriptsize $\pm$3.2}} \\
DeepSeek-V4-Pro & 531 & \shortstack[r]{73.5\\{\scriptsize $\pm$2.3}} & --- & \shortstack[r]{72.7\%\\{\scriptsize $\pm$3.8}} \\
Gemini-3.1-Flash-Lite & 531 & \shortstack[r]{73.3\\{\scriptsize $\pm$2.2}} & --- & \shortstack[r]{75.5\%\\{\scriptsize $\pm$3.6}} \\
GPT-5.4-mini & 531 & \shortstack[r]{71.8\\{\scriptsize $\pm$2.4}} & --- & \shortstack[r]{71.6\%\\{\scriptsize $\pm$3.8}} \\
DeepSeek-V4-Flash & 530 & \shortstack[r]{70.7\\{\scriptsize $\pm$2.3}} & --- & \shortstack[r]{73.4\%\\{\scriptsize $\pm$3.7}} \\
Llama-4 & 531 & \shortstack[r]{70.1\\{\scriptsize $\pm$2.4}} & --- & \shortstack[r]{75.0\%\\{\scriptsize $\pm$3.6}} \\
Qwen3.5-122B & 526 & \shortstack[r]{69.8\\{\scriptsize $\pm$2.3}} & --- & \shortstack[r]{75.3\%\\{\scriptsize $\pm$3.6}} \\
Qwen3.6-35B & 531 & \shortstack[r]{69.5\\{\scriptsize $\pm$2.5}} & --- & \shortstack[r]{68.9\%\\{\scriptsize $\pm$3.8}} \\
Qwen3-235B & 531 & \shortstack[r]{67.3\\{\scriptsize $\pm$2.4}} & --- & \shortstack[r]{69.5\%\\{\scriptsize $\pm$4.0}} \\
\bottomrule
\end{tabular*}
\end{table*}

\begin{table*}[t]
\centering\footnotesize
\caption{Automatic-metric results on the \textbf{Benchathon} corpus: per-system bootstrap 95\% CIs ($B=2000$, seed $42$) on the surface and embedding-based metrics. Each cell shows the mean stacked above $\pm$ half the width of the 95\% CI; dashes mark metrics not computed on this dataset.}
\label{tbl-full-results-auto-benchathon}
\setlength{\tabcolsep}{4pt}
\begin{tabular*}{\textwidth}{@{\extracolsep{\fill}}lrrrrrrr@{}}
\toprule
System & $n$ & BLEU & ROUGE & METEOR & BERTSc & MovSc & SemSim \\
\midrule
\textit{Human (traditional)} & 65 & \shortstack[r]{0.055\\{\scriptsize $\pm$0.014}} & \shortstack[r]{0.171\\{\scriptsize $\pm$0.013}} & \shortstack[r]{0.181\\{\scriptsize $\pm$0.020}} & \shortstack[r]{0.325\\{\scriptsize $\pm$0.029}} & \shortstack[r]{0.908\\{\scriptsize $\pm$0.005}} & \shortstack[r]{0.705\\{\scriptsize $\pm$0.056}} \\
\textit{Human (co-creation)} & 155 & \shortstack[r]{0.083\\{\scriptsize $\pm$0.005}} & \shortstack[r]{0.186\\{\scriptsize $\pm$0.006}} & \shortstack[r]{0.255\\{\scriptsize $\pm$0.012}} & \shortstack[r]{0.338\\{\scriptsize $\pm$0.016}} & \shortstack[r]{0.916\\{\scriptsize $\pm$0.003}} & \shortstack[r]{0.761\\{\scriptsize $\pm$0.025}} \\
Opus-4.7 & 15 & \shortstack[r]{0.065\\{\scriptsize $\pm$0.017}} & \shortstack[r]{0.193\\{\scriptsize $\pm$0.021}} & \shortstack[r]{0.203\\{\scriptsize $\pm$0.027}} & \shortstack[r]{0.384\\{\scriptsize $\pm$0.032}} & \shortstack[r]{0.916\\{\scriptsize $\pm$0.004}} & \shortstack[r]{0.795\\{\scriptsize $\pm$0.049}} \\
Gemini-3.1-Pro & 15 & \shortstack[r]{0.070\\{\scriptsize $\pm$0.020}} & \shortstack[r]{0.207\\{\scriptsize $\pm$0.030}} & \shortstack[r]{0.202\\{\scriptsize $\pm$0.036}} & \shortstack[r]{0.399\\{\scriptsize $\pm$0.040}} & \shortstack[r]{0.921\\{\scriptsize $\pm$0.008}} & \shortstack[r]{0.790\\{\scriptsize $\pm$0.064}} \\
GPT-5.4 & 15 & \shortstack[r]{0.101\\{\scriptsize $\pm$0.019}} & \shortstack[r]{0.194\\{\scriptsize $\pm$0.018}} & \shortstack[r]{0.269\\{\scriptsize $\pm$0.027}} & \shortstack[r]{0.363\\{\scriptsize $\pm$0.046}} & \shortstack[r]{0.920\\{\scriptsize $\pm$0.005}} & \shortstack[r]{0.789\\{\scriptsize $\pm$0.050}} \\
Gemini-3.1-Flash-Lite & 15 & \shortstack[r]{0.055\\{\scriptsize $\pm$0.015}} & \shortstack[r]{0.191\\{\scriptsize $\pm$0.018}} & \shortstack[r]{0.185\\{\scriptsize $\pm$0.022}} & \shortstack[r]{0.384\\{\scriptsize $\pm$0.033}} & \shortstack[r]{0.920\\{\scriptsize $\pm$0.003}} & \shortstack[r]{0.794\\{\scriptsize $\pm$0.063}} \\
Sonnet-4.6 & 15 & \shortstack[r]{0.094\\{\scriptsize $\pm$0.010}} & \shortstack[r]{0.180\\{\scriptsize $\pm$0.011}} & \shortstack[r]{0.270\\{\scriptsize $\pm$0.018}} & \shortstack[r]{0.338\\{\scriptsize $\pm$0.042}} & \shortstack[r]{0.919\\{\scriptsize $\pm$0.004}} & \shortstack[r]{0.768\\{\scriptsize $\pm$0.054}} \\
GPT-5.4-mini & 15 & \shortstack[r]{0.045\\{\scriptsize $\pm$0.010}} & \shortstack[r]{0.167\\{\scriptsize $\pm$0.013}} & \shortstack[r]{0.175\\{\scriptsize $\pm$0.018}} & \shortstack[r]{0.336\\{\scriptsize $\pm$0.033}} & \shortstack[r]{0.909\\{\scriptsize $\pm$0.006}} & \shortstack[r]{0.785\\{\scriptsize $\pm$0.049}} \\
DeepSeek-V4-Pro & 15 & \shortstack[r]{0.083\\{\scriptsize $\pm$0.019}} & \shortstack[r]{0.182\\{\scriptsize $\pm$0.014}} & \shortstack[r]{0.246\\{\scriptsize $\pm$0.036}} & \shortstack[r]{0.353\\{\scriptsize $\pm$0.043}} & \shortstack[r]{0.918\\{\scriptsize $\pm$0.006}} & \shortstack[r]{0.766\\{\scriptsize $\pm$0.054}} \\
Qwen3.5-122B & 15 & \shortstack[r]{0.031\\{\scriptsize $\pm$0.011}} & \shortstack[r]{0.144\\{\scriptsize $\pm$0.017}} & \shortstack[r]{0.149\\{\scriptsize $\pm$0.027}} & \shortstack[r]{0.329\\{\scriptsize $\pm$0.027}} & \shortstack[r]{0.905\\{\scriptsize $\pm$0.008}} & \shortstack[r]{0.767\\{\scriptsize $\pm$0.058}} \\
DeepSeek-V4-Flash & 15 & \shortstack[r]{0.067\\{\scriptsize $\pm$0.013}} & \shortstack[r]{0.182\\{\scriptsize $\pm$0.014}} & \shortstack[r]{0.214\\{\scriptsize $\pm$0.020}} & \shortstack[r]{0.374\\{\scriptsize $\pm$0.033}} & \shortstack[r]{0.912\\{\scriptsize $\pm$0.005}} & \shortstack[r]{0.778\\{\scriptsize $\pm$0.066}} \\
Qwen3.6-35B & 15 & \shortstack[r]{0.015\\{\scriptsize $\pm$0.007}} & \shortstack[r]{0.123\\{\scriptsize $\pm$0.019}} & \shortstack[r]{0.114\\{\scriptsize $\pm$0.023}} & \shortstack[r]{0.257\\{\scriptsize $\pm$0.035}} & \shortstack[r]{0.896\\{\scriptsize $\pm$0.012}} & \shortstack[r]{0.698\\{\scriptsize $\pm$0.052}} \\
Qwen3-235B & 15 & \shortstack[r]{0.010\\{\scriptsize $\pm$0.003}} & \shortstack[r]{0.121\\{\scriptsize $\pm$0.008}} & \shortstack[r]{0.099\\{\scriptsize $\pm$0.010}} & \shortstack[r]{0.277\\{\scriptsize $\pm$0.017}} & \shortstack[r]{0.883\\{\scriptsize $\pm$0.004}} & \shortstack[r]{0.761\\{\scriptsize $\pm$0.044}} \\
Llama-4 & 15 & \shortstack[r]{0.014\\{\scriptsize $\pm$0.007}} & \shortstack[r]{0.143\\{\scriptsize $\pm$0.013}} & \shortstack[r]{0.114\\{\scriptsize $\pm$0.016}} & \shortstack[r]{0.311\\{\scriptsize $\pm$0.022}} & \shortstack[r]{0.892\\{\scriptsize $\pm$0.005}} & \shortstack[r]{0.776\\{\scriptsize $\pm$0.051}} \\
\bottomrule
\end{tabular*}
\end{table*}

\begin{table*}[t]
\centering\footnotesize
\caption{Automatic-metric results on the \textbf{ZJS} corpus: per-system bootstrap 95\% CIs ($B=2000$, seed $42$) on the surface and embedding-based metrics. Each cell shows the mean stacked above $\pm$ half the width of the 95\% CI; dashes mark metrics not computed on this dataset.}
\label{tbl-full-results-auto-zjs}
\setlength{\tabcolsep}{4pt}
\begin{tabular*}{\textwidth}{@{\extracolsep{\fill}}lrrrrrrr@{}}
\toprule
System & $n$ & BLEU & ROUGE & METEOR & BERTSc & MovSc & SemSim \\
\midrule
Gemini-3.1-Pro & 579 & \shortstack[r]{0.039\\{\scriptsize $\pm$0.003}} & \shortstack[r]{0.166\\{\scriptsize $\pm$0.003}} & \shortstack[r]{0.152\\{\scriptsize $\pm$0.004}} & \shortstack[r]{0.369\\{\scriptsize $\pm$0.007}} & \shortstack[r]{0.919\\{\scriptsize $\pm$0.001}} & \shortstack[r]{0.780\\{\scriptsize $\pm$0.009}} \\
GPT-5.4 & 581 & \shortstack[r]{0.078\\{\scriptsize $\pm$0.003}} & \shortstack[r]{0.165\\{\scriptsize $\pm$0.002}} & \shortstack[r]{0.218\\{\scriptsize $\pm$0.004}} & \shortstack[r]{0.350\\{\scriptsize $\pm$0.006}} & \shortstack[r]{0.919\\{\scriptsize $\pm$0.001}} & \shortstack[r]{0.767\\{\scriptsize $\pm$0.009}} \\
Opus-4.7 & 581 & \shortstack[r]{0.042\\{\scriptsize $\pm$0.002}} & \shortstack[r]{0.159\\{\scriptsize $\pm$0.003}} & \shortstack[r]{0.157\\{\scriptsize $\pm$0.004}} & \shortstack[r]{0.354\\{\scriptsize $\pm$0.007}} & \shortstack[r]{0.913\\{\scriptsize $\pm$0.001}} & \shortstack[r]{0.769\\{\scriptsize $\pm$0.009}} \\
Sonnet-4.6 & 581 & \shortstack[r]{0.081\\{\scriptsize $\pm$0.003}} & \shortstack[r]{0.164\\{\scriptsize $\pm$0.002}} & \shortstack[r]{0.224\\{\scriptsize $\pm$0.005}} & \shortstack[r]{0.335\\{\scriptsize $\pm$0.006}} & \shortstack[r]{0.917\\{\scriptsize $\pm$0.001}} & \shortstack[r]{0.747\\{\scriptsize $\pm$0.009}} \\
DeepSeek-V4-Pro & 581 & \shortstack[r]{0.071\\{\scriptsize $\pm$0.003}} & \shortstack[r]{0.165\\{\scriptsize $\pm$0.002}} & \shortstack[r]{0.208\\{\scriptsize $\pm$0.005}} & \shortstack[r]{0.340\\{\scriptsize $\pm$0.006}} & \shortstack[r]{0.920\\{\scriptsize $\pm$0.001}} & \shortstack[r]{0.762\\{\scriptsize $\pm$0.009}} \\
GPT-5.4-mini & 581 & \shortstack[r]{0.031\\{\scriptsize $\pm$0.002}} & \shortstack[r]{0.141\\{\scriptsize $\pm$0.002}} & \shortstack[r]{0.145\\{\scriptsize $\pm$0.004}} & \shortstack[r]{0.324\\{\scriptsize $\pm$0.006}} & \shortstack[r]{0.907\\{\scriptsize $\pm$0.001}} & \shortstack[r]{0.751\\{\scriptsize $\pm$0.009}} \\
DeepSeek-V4-Flash & 581 & \shortstack[r]{0.047\\{\scriptsize $\pm$0.003}} & \shortstack[r]{0.155\\{\scriptsize $\pm$0.003}} & \shortstack[r]{0.169\\{\scriptsize $\pm$0.004}} & \shortstack[r]{0.330\\{\scriptsize $\pm$0.006}} & \shortstack[r]{0.909\\{\scriptsize $\pm$0.001}} & \shortstack[r]{0.745\\{\scriptsize $\pm$0.009}} \\
Gemini-3.1-Flash-Lite & 579 & \shortstack[r]{0.004\\{\scriptsize $\pm$0.001}} & \shortstack[r]{0.108\\{\scriptsize $\pm$0.002}} & \shortstack[r]{0.073\\{\scriptsize $\pm$0.002}} & \shortstack[r]{0.308\\{\scriptsize $\pm$0.005}} & \shortstack[r]{0.896\\{\scriptsize $\pm$0.001}} & \shortstack[r]{0.762\\{\scriptsize $\pm$0.008}} \\
Qwen3.5-122B & 581 & \shortstack[r]{0.018\\{\scriptsize $\pm$0.002}} & \shortstack[r]{0.131\\{\scriptsize $\pm$0.002}} & \shortstack[r]{0.122\\{\scriptsize $\pm$0.003}} & \shortstack[r]{0.317\\{\scriptsize $\pm$0.005}} & \shortstack[r]{0.905\\{\scriptsize $\pm$0.001}} & \shortstack[r]{0.759\\{\scriptsize $\pm$0.009}} \\
Qwen3.6-35B & 581 & \shortstack[r]{0.010\\{\scriptsize $\pm$0.002}} & \shortstack[r]{0.116\\{\scriptsize $\pm$0.002}} & \shortstack[r]{0.100\\{\scriptsize $\pm$0.002}} & \shortstack[r]{0.279\\{\scriptsize $\pm$0.005}} & \shortstack[r]{0.899\\{\scriptsize $\pm$0.002}} & \shortstack[r]{0.715\\{\scriptsize $\pm$0.009}} \\
Llama-4 & 581 & \shortstack[r]{0.004\\{\scriptsize $\pm$0.001}} & \shortstack[r]{0.108\\{\scriptsize $\pm$0.003}} & \shortstack[r]{0.077\\{\scriptsize $\pm$0.003}} & \shortstack[r]{0.297\\{\scriptsize $\pm$0.006}} & \shortstack[r]{0.887\\{\scriptsize $\pm$0.001}} & \shortstack[r]{0.749\\{\scriptsize $\pm$0.009}} \\
Qwen3-235B & 581 & \shortstack[r]{0.006\\{\scriptsize $\pm$0.001}} & \shortstack[r]{0.108\\{\scriptsize $\pm$0.002}} & \shortstack[r]{0.080\\{\scriptsize $\pm$0.003}} & \shortstack[r]{0.267\\{\scriptsize $\pm$0.005}} & \shortstack[r]{0.882\\{\scriptsize $\pm$0.001}} & \shortstack[r]{0.724\\{\scriptsize $\pm$0.008}} \\
\bottomrule
\end{tabular*}
\end{table*}

\begin{table*}[t]
\centering\footnotesize
\caption{Automatic-metric and task-correctness results on the \textbf{Doctrinal Principles} corpus: per-system bootstrap 95\% CIs ($B=2000$, seed $42$) on yes/no (\emph{Ja/Nein}) decision accuracy (Acc.; task-correctness, not a rubric measure) together with the surface and embedding-based metrics. Each cell shows the mean stacked above $\pm$ half the width of the 95\% CI; dashes mark metrics not computed on this dataset.}
\label{tbl-full-results-auto-grundpr}
\setlength{\tabcolsep}{4pt}
\begin{tabular*}{\textwidth}{@{\extracolsep{\fill}}lrrrrrrrrr@{}}
\toprule
System & $n$ & Acc. & chrF & BLEU & ROUGE & METEOR & BERTSc & MovSc & SemSim \\
\midrule
Gemini-3.1-Pro & 531 & \shortstack[r]{83.4\%\\{\scriptsize $\pm$3.1}} & \shortstack[r]{0.414\\{\scriptsize $\pm$0.007}} & \shortstack[r]{0.033\\{\scriptsize $\pm$0.003}} & \shortstack[r]{0.164\\{\scriptsize $\pm$0.005}} & \shortstack[r]{0.212\\{\scriptsize $\pm$0.008}} & \shortstack[r]{0.250\\{\scriptsize $\pm$0.007}} & \shortstack[r]{0.840\\{\scriptsize $\pm$0.002}} & \shortstack[r]{0.745\\{\scriptsize $\pm$0.008}} \\
Opus-4.7 & 531 & \shortstack[r]{81.0\%\\{\scriptsize $\pm$3.3}} & \shortstack[r]{0.415\\{\scriptsize $\pm$0.007}} & \shortstack[r]{0.030\\{\scriptsize $\pm$0.003}} & \shortstack[r]{0.159\\{\scriptsize $\pm$0.005}} & \shortstack[r]{0.210\\{\scriptsize $\pm$0.007}} & \shortstack[r]{0.247\\{\scriptsize $\pm$0.007}} & \shortstack[r]{0.838\\{\scriptsize $\pm$0.002}} & \shortstack[r]{0.740\\{\scriptsize $\pm$0.009}} \\
Sonnet-4.6 & 531 & \shortstack[r]{80.6\%\\{\scriptsize $\pm$3.3}} & \shortstack[r]{0.425\\{\scriptsize $\pm$0.006}} & \shortstack[r]{0.028\\{\scriptsize $\pm$0.003}} & \shortstack[r]{0.147\\{\scriptsize $\pm$0.004}} & \shortstack[r]{0.233\\{\scriptsize $\pm$0.006}} & \shortstack[r]{0.220\\{\scriptsize $\pm$0.007}} & \shortstack[r]{0.843\\{\scriptsize $\pm$0.002}} & \shortstack[r]{0.729\\{\scriptsize $\pm$0.009}} \\
GPT-5.4 & 531 & \shortstack[r]{78.7\%\\{\scriptsize $\pm$3.4}} & \shortstack[r]{0.421\\{\scriptsize $\pm$0.006}} & \shortstack[r]{0.031\\{\scriptsize $\pm$0.003}} & \shortstack[r]{0.160\\{\scriptsize $\pm$0.005}} & \shortstack[r]{0.210\\{\scriptsize $\pm$0.007}} & \shortstack[r]{0.249\\{\scriptsize $\pm$0.007}} & \shortstack[r]{0.845\\{\scriptsize $\pm$0.002}} & \shortstack[r]{0.758\\{\scriptsize $\pm$0.009}} \\
DeepSeek-V4-Pro & 531 & \shortstack[r]{72.9\%\\{\scriptsize $\pm$3.8}} & \shortstack[r]{0.419\\{\scriptsize $\pm$0.006}} & \shortstack[r]{0.030\\{\scriptsize $\pm$0.003}} & \shortstack[r]{0.161\\{\scriptsize $\pm$0.005}} & \shortstack[r]{0.210\\{\scriptsize $\pm$0.007}} & \shortstack[r]{0.245\\{\scriptsize $\pm$0.008}} & \shortstack[r]{0.843\\{\scriptsize $\pm$0.002}} & \shortstack[r]{0.752\\{\scriptsize $\pm$0.009}} \\
Gemini-3.1-Flash-Lite & 531 & \shortstack[r]{75.5\%\\{\scriptsize $\pm$3.5}} & \shortstack[r]{0.394\\{\scriptsize $\pm$0.007}} & \shortstack[r]{0.029\\{\scriptsize $\pm$0.003}} & \shortstack[r]{0.162\\{\scriptsize $\pm$0.005}} & \shortstack[r]{0.187\\{\scriptsize $\pm$0.008}} & \shortstack[r]{0.253\\{\scriptsize $\pm$0.007}} & \shortstack[r]{0.838\\{\scriptsize $\pm$0.002}} & \shortstack[r]{0.753\\{\scriptsize $\pm$0.009}} \\
GPT-5.4-mini & 531 & \shortstack[r]{70.6\%\\{\scriptsize $\pm$3.8}} & \shortstack[r]{0.380\\{\scriptsize $\pm$0.007}} & \shortstack[r]{0.028\\{\scriptsize $\pm$0.002}} & \shortstack[r]{0.162\\{\scriptsize $\pm$0.005}} & \shortstack[r]{0.177\\{\scriptsize $\pm$0.006}} & \shortstack[r]{0.250\\{\scriptsize $\pm$0.007}} & \shortstack[r]{0.839\\{\scriptsize $\pm$0.002}} & \shortstack[r]{0.755\\{\scriptsize $\pm$0.009}} \\
DeepSeek-V4-Flash & 530 & \shortstack[r]{68.9\%\\{\scriptsize $\pm$3.8}} & \shortstack[r]{0.412\\{\scriptsize $\pm$0.006}} & \shortstack[r]{0.031\\{\scriptsize $\pm$0.003}} & \shortstack[r]{0.164\\{\scriptsize $\pm$0.005}} & \shortstack[r]{0.209\\{\scriptsize $\pm$0.006}} & \shortstack[r]{0.249\\{\scriptsize $\pm$0.007}} & \shortstack[r]{0.841\\{\scriptsize $\pm$0.002}} & \shortstack[r]{0.749\\{\scriptsize $\pm$0.008}} \\
Llama-4 & 531 & \shortstack[r]{77.0\%\\{\scriptsize $\pm$3.7}} & \shortstack[r]{0.388\\{\scriptsize $\pm$0.008}} & \shortstack[r]{0.035\\{\scriptsize $\pm$0.003}} & \shortstack[r]{0.175\\{\scriptsize $\pm$0.005}} & \shortstack[r]{0.199\\{\scriptsize $\pm$0.007}} & \shortstack[r]{0.254\\{\scriptsize $\pm$0.008}} & \shortstack[r]{0.845\\{\scriptsize $\pm$0.002}} & \shortstack[r]{0.747\\{\scriptsize $\pm$0.010}} \\
Qwen3.5-122B & 526 & \shortstack[r]{77.6\%\\{\scriptsize $\pm$3.4}} & \shortstack[r]{0.326\\{\scriptsize $\pm$0.009}} & \shortstack[r]{0.026\\{\scriptsize $\pm$0.003}} & \shortstack[r]{0.165\\{\scriptsize $\pm$0.006}} & \shortstack[r]{0.157\\{\scriptsize $\pm$0.008}} & \shortstack[r]{0.252\\{\scriptsize $\pm$0.009}} & \shortstack[r]{0.827\\{\scriptsize $\pm$0.003}} & \shortstack[r]{0.742\\{\scriptsize $\pm$0.009}} \\
Qwen3.6-35B & 531 & \shortstack[r]{70.6\%\\{\scriptsize $\pm$3.6}} & \shortstack[r]{0.387\\{\scriptsize $\pm$0.007}} & \shortstack[r]{0.027\\{\scriptsize $\pm$0.003}} & \shortstack[r]{0.154\\{\scriptsize $\pm$0.005}} & \shortstack[r]{0.181\\{\scriptsize $\pm$0.008}} & \shortstack[r]{0.239\\{\scriptsize $\pm$0.007}} & \shortstack[r]{0.834\\{\scriptsize $\pm$0.002}} & \shortstack[r]{0.747\\{\scriptsize $\pm$0.008}} \\
Qwen3-235B & 531 & \shortstack[r]{72.7\%\\{\scriptsize $\pm$3.8}} & \shortstack[r]{0.374\\{\scriptsize $\pm$0.008}} & \shortstack[r]{0.025\\{\scriptsize $\pm$0.002}} & \shortstack[r]{0.156\\{\scriptsize $\pm$0.005}} & \shortstack[r]{0.167\\{\scriptsize $\pm$0.006}} & \shortstack[r]{0.241\\{\scriptsize $\pm$0.007}} & \shortstack[r]{0.824\\{\scriptsize $\pm$0.002}} & \shortstack[r]{0.740\\{\scriptsize $\pm$0.009}} \\
\bottomrule
\end{tabular*}
\end{table*}

\subsection{Reasoning-core robustness}\label{reasoning-core-robustness}

Table~\ref{tbl-rubric-mapping} maps each rubric dimension to the
\emph{Gutachtenstil} scaffold step it grades, with its percentage
weight. Table~\ref{tbl-reasoning-core} recomputes each system's mean
score using only the seven scaffold-step rubric dimensions (90 of 100
points; dropping \emph{Gliederung}, \emph{Sprache}, \emph{Formalia}),
rescaled to /100. The full-rubric system ranking is reproduced exactly
on both corpora.

\begin{table}[t]
\centering
\footnotesize
\setlength{\tabcolsep}{4pt}
\begin{tabular}{@{}p{0.58\columnwidth}p{0.32\columnwidth}@{}}
\toprule
\textbf{Dimension (weight / 100)} & \textbf{Scaffold step graded} \\
\midrule
Result correctness / \emph{Ergebnisrichtigkeit} (20) & \emph{Ergebnis} \\
Doctrinal knowledge / \emph{Rechtskenntnis} (15) & \emph{Definition} \\
Subsumption quality / \emph{Subsumtion} (15) & \emph{Subsumtion} \\
Issue identification / \emph{Vollständigkeit} (10) & \emph{Obersatz} (issue spotting) \\
Legal grounding / \emph{Rechtsgrundlagen} (10) & \emph{Obersatz} (norm citation) \\
Problem depth / \emph{Schwerpunktsetzung} (10) & \emph{Subsumtion} (focal-issue depth) \\
Methodological structure / \emph{Methodischer Stil} (10) & whole-scaffold execution \\
Organization / \emph{Gliederung} (5) & form \\
Terminology / \emph{Sprache} (3) & form \\
Formal correctness / \emph{Formalia} (2) & form \\
\bottomrule
\end{tabular}
\caption{Rubric dimensions mapped to the \emph{Gutachtenstil} scaffold step each grades, with /100 weights from the judge prompt (Appendix~\ref{evaluation-prompts-for-the-llm-judge}). The judge's global principle additionally gates all credit on the method: ``Keine Punkte für bloße Schlagworte ohne Definition/Subsumtion'' (no points for mere buzzwords without definition/subsumption).}
\label{tbl-rubric-mapping}
\end{table}

\begin{table}[t]
\centering\footnotesize
\caption{Per-system mean judge score under the full 10-dimension rubric (\emph{Full}) vs.\ the seven-dimension reasoning core rescaled to /100 (\emph{Core}), primary judge. System rank agreement full vs.\ core: Spearman $\rho$ = 1.00 (Benchathon), 1.00 (ZJS); Kendall $\tau$ = 1.00 / 1.00.}
\label{tbl-reasoning-core}
\setlength{\tabcolsep}{3pt}
\begin{tabular*}{\columnwidth}{@{\extracolsep{\fill}}lcccc@{}}
\toprule
Corpus & System & n & Full & Core \\
\midrule
Benchathon & Opus-4.7 & 15 & 69.3 & 67.6 \\
 & Gemini-3.1-Pro & 15 & 68.3 & 66.7 \\
 & GPT-5.4 & 15 & 66.6 & 64.6 \\
 & Gemini-3.1-Flash-Lite & 15 & 63.5 & 61.3 \\
 & Sonnet-4.6 & 15 & 58.9 & 56.1 \\
 & GPT-5.4-mini & 15 & 58.6 & 55.9 \\
 & DeepSeek-V4-Pro & 15 & 56.1 & 53.2 \\
 & Qwen3.5-122B & 15 & 49.4 & 46.1 \\
 & DeepSeek-V4-Flash & 15 & 49.3 & 45.7 \\
 & Qwen3.6-35B & 15 & 42.4 & 38.4 \\
 & Qwen3-235B & 15 & 41.9 & 38.2 \\
 & Llama-4 & 15 & 37.8 & 33.4 \\
ZJS & Gemini-3.1-Pro & 578 & 61.8 & 59.3 \\
 & GPT-5.4 & 580 & 60.4 & 57.8 \\
 & Opus-4.7 & 580 & 58.7 & 55.9 \\
 & Sonnet-4.6 & 580 & 51.5 & 48.1 \\
 & DeepSeek-V4-Pro & 580 & 50.1 & 46.7 \\
 & GPT-5.4-mini & 581 & 49.7 & 46.2 \\
 & DeepSeek-V4-Flash & 579 & 45.9 & 42.1 \\
 & Gemini-3.1-Flash-Lite & 578 & 44.6 & 40.8 \\
 & Qwen3.5-122B & 581 & 44.2 & 40.3 \\
 & Qwen3.6-35B & 580 & 39.8 & 35.6 \\
 & Llama-4 & 581 & 36.1 & 31.7 \\
 & Qwen3-235B & 581 & 35.9 & 31.5 \\
\bottomrule
\end{tabular*}
\end{table}

\section{Evaluated systems catalogue}\label{evaluated-systems-catalogue}

Table~\ref{tbl-system-overview} catalogues the LLM systems evaluated in
this work alongside their provider family, the short alias used
throughout the body of the paper, the full model id, tier classification
(closed flagship, efficiency-oriented, or open-weight reference),
weights status (open vs.~closed), the decoding temperature and
max-output-tokens setting used, the total generation count on the
Benchathon subset, and the count of truncated generations over that
total.

The Google efficiency-tier slot was generated with two distinct Google
models: Gemini-3-Flash on the Benchathon subset, and
Gemini-3.1-Flash-Lite (a separate efficiency product line, not a
successor release of Flash) on the ZJS and Doctrinal Principles corpora.
The model set was finalised between the two generation runs. We treat
them as the same conceptual tier slot for tier-level comparisons and
disclose the raw model ids in Table~\ref{tbl-system-overview}. Row-level
cross-corpus comparisons for this slot should be read with this model
change in mind, and tier-level claims should not be read as claims about
either Flash or Flash-Lite alone.

\begin{table*}[t]
\centering\footnotesize
\caption{Evaluated LLM systems with provider family, the short alias used throughout the paper, the full model id, tier (flagship / efficiency-oriented / open-weight reference), weights (open vs.\ closed), reasoning-mode disclosure (explicit thinking variant / no explicit mechanism / undisclosed for closed products with possible internal routing), observed generation temperature, max-output-tokens setting, total generation count on the Benchathon subset, and the count of truncated generations (truncated flag or finish\_reason in length / max\_tokens) over total. The Google efficiency-tier row aggregates two distinct Google models across corpora: Gemini-3-Flash on Benchathon and Gemini-3.1-Flash-Lite (a separate product line) on ZJS and Doctrinal Principles; see the appendix text.}
\label{tbl-system-overview}
\setlength{\tabcolsep}{4pt}
\resizebox{\textwidth}{!}{%
\begin{tabular}{lccccccccc}
\toprule
Provider & Alias & System & Tier & Weights & Reasoning & $T$ & Max out & Gens & Trunc. \\
\midrule
Anthropic & Opus-4.7 & \texttt{claude-opus-4-7} & Flagship & Closed & undisclosed & 1 & 8000 & 15 & 0/15 \\
Anthropic & Sonnet-4.6 & \texttt{claude-sonnet-4-6} & Flagship & Closed & undisclosed & 1 & 8000 & 15 & 0/15 \\
Google & Gemini-3.1-Flash-Lite & \texttt{gemini-3.1-flash-lite-preview} & Efficiency & Closed & undisclosed & 1 & 8000 & 15 & 0/15 \\
Google & Gemini-3.1-Pro & \texttt{gemini-3.1-pro-preview} & Flagship & Closed & undisclosed & 1 & 8000 & 15 & 2/15 \\
OpenAI & GPT-5.4 & \texttt{gpt-5.4} & Flagship & Closed & undisclosed & 1 & 16000 & 15 & 0/15 \\
OpenAI & GPT-5.4-mini & \texttt{gpt-5.4-mini} & Efficiency & Closed & undisclosed & 1 & 8000 & 15 & 0/15 \\
Alibaba & Qwen3-235B & \texttt{Qwen3-235B-A22B-Thinking-2507} & Open ref. & Open & explicit (thinking) & 0.6 & 16000 & 15 & 0/15 \\
Alibaba & Qwen3.5-122B & \texttt{Qwen3.5-122B-A10B} & Open ref. & Open & none & 0.7 & 8000 & 15 & 6/15 \\
Alibaba & Qwen3.6-35B & \texttt{Qwen3.6-35B-A3B} & Open ref. & Open & none & 0.7 & 8000 & 15 & 4/15 \\
DeepSeek & DeepSeek-V4-Flash & \texttt{DeepSeek-V4-Flash} & Open ref. & Open & none & 1 & 8000 & 15 & 0/15 \\
DeepSeek & DeepSeek-V4-Pro & \texttt{DeepSeek-V4-Pro} & Open ref. & Open & none & 1 & 8000 & 15 & 0/15 \\
Meta & Llama-4 & \texttt{Llama-4-Maverick-17B-128E-Instruct-FP8} & Open ref. & Open & none & 0.6 & 8000 & 15 & 0/15 \\
\bottomrule
\end{tabular}
}
\end{table*}

\section{Operational metadata and
reproducibility}\label{operational-metadata-and-reproducibility}

Table~\ref{tbl-operational} reports per-system operational metadata
(mean per-generation cost, latency, and output length) from the per-call
\texttt{response\_metadata} the platform logs for every request. Cost
cells are empty where the provider route did not report billing at
generation time.

\begin{table*}[t]
\centering\footnotesize
\caption{Per-system operational metadata from logged per-call metadata: mean per-generation cost (USD), mean latency (s), and mean output length (words), per corpus. Doctrinal Principles cost was not billed through the logging route and is omitted.}
\label{tbl-operational}
\setlength{\tabcolsep}{4pt}
\begin{tabular*}{\textwidth}{@{\extracolsep{\fill}}lcccccccc@{}}
\toprule
System & B.~\$/gen & B.~s & B.~words & Z.~\$/gen & Z.~s & Z.~words & G.~s & G.~words \\
\midrule
Opus-4.7 & \$0.150 & 73.3 & 1422 & \$0.180 & 75.2 & 1710 & 9.4 & 104 \\
Sonnet-4.6 & \$0.101 & 104.0 & 2355 & \$0.120 & 126.1 & 2796 & 10.0 & 167 \\
Gemini-3.1-Flash-Lite & \$0.008 & 14.9 & 1281 & \$0.003 & 6.5 & 699 & 1.4 & 79 \\
Gemini-3.1-Pro & \$0.035 & 102.7 & 1349 & \$0.041 & 72.6 & 1558 & 13.0 & 95 \\
GPT-5.4 & — & 70.0 & 2190 & — & 79.9 & 2649 & 4.6 & 96 \\
GPT-5.4-mini & — & 17.2 & 1252 & — & 21.4 & 1570 & 2.0 & 67 \\
Qwen3-235B & \$0.014 & 125.1 & 825 & \$0.015 & 148.4 & 971 & 46.1 & 72 \\
Qwen3.5-122B & \$0.018 & 77.0 & 1022 & \$0.020 & 90.1 & 1297 & 94.2 & 49 \\
Qwen3.6-35B & \$0.008 & 74.8 & 802 & \$0.008 & 64.8 & 1122 & 18.5 & 77 \\
DeepSeek-V4-Flash & \$0.001 & 89.1 & 1650 & \$0.001 & 129.1 & 1906 & 53.9 & 101 \\
DeepSeek-V4-Pro & \$0.019 & 98.6 & 2012 & \$0.023 & 130.2 & 2498 & 16.1 & 109 \\
Llama-4 & \$0.001 & 37.8 & 691 & \$0.001 & 37.2 & 709 & 9.4 & 86 \\
\bottomrule
\end{tabular*}
\end{table*}

\paragraph{Reproducibility addendum.} The same per-call metadata records
the run settings beyond the per-system temperatures and token budgets of
Table~\ref{tbl-system-overview}. \emph{Run dates}: Benchathon:
2026-05-08 to 2026-05-08. ZJS: 2026-05-13 to 2026-05-14. Doctrinal
Principles: 2026-05-19 to 2026-05-19. The ZJS range spans the initial
8,000-token run and the next-day re-generation of truncated attempts;
superseded attempts (the ZJS first attempts and the April Benchathon
round; \hyperref[experimental-setup]{§Experimental setup}) are excluded
from all per-generation aggregates here. \emph{Retry budget}: generation
calls run under the worker-side retry budget of 3 (the same default as
the judge calls in Appendix \ref{llm-judge-configuration}); observed
retries across kept generation calls: 0. \emph{Sampling parameters}:
temperature and max output tokens were set explicitly per system
(Table~\ref{tbl-system-overview}); no other sampling parameter (top-p,
top-k, seed, thinking budget, reasoning effort) appears in any logged
call; those ran at provider defaults, including the reasoning behaviour
of the one explicitly reasoning-tagged system
(Qwen3-235B-A22B-Thinking-2507). \emph{Earlier version}: an earlier
version of this work used GPT-5-mini as the primary judge; the final
version re-scored every generation with GPT-5.4-mini and extended
cross-validation to six judges, leaving tier-level conclusions unchanged
while absolute scores shift as a calibrated ordering
(\hyperref[limitations]{Limitations}). That version's ZJS and Benchathon
aggregates also double-counted superseded generation attempts;
deduplicating raises truncation-prone ZJS means by up to
\textasciitilde3 raw points and makes Benchathon aggregates
final-run-only (n=15).

\section{Per-task per-system heatmap}\label{per-task-per-system-heatmap}

\begin{figure*}[!t]

\centering{

\pandocbounded{\includegraphics[keepaspectratio]{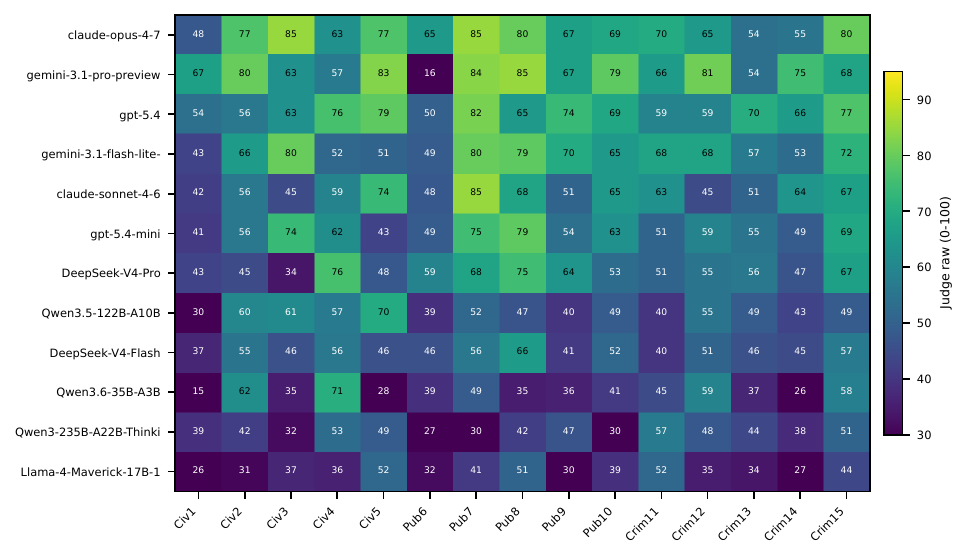}}

}

\caption{\label{fig-task-system-heatmap}Per-task per-system mean
LLM-judge raw score on the Benchathon subset. Rows are systems ordered
by overall mean (top = highest). Columns are the 15 Benchathon tasks
ordered by inner\_id, grouped by legal area (Civ = civil law, Pub =
public law, Crim = criminal law).}

\end{figure*}

Figure~\ref{fig-task-system-heatmap} visualises the per-task per-system
mean LLM-judge raw score on the Benchathon subset. Rows are systems
ordered by overall mean (top = highest); columns are the 15 Benchathon
tasks ordered by inner identifier and grouped by legal area (Civ = civil
law, Pub = public law, Crim = criminal law). The heatmap exposes the
within-task variance discussed in RQ1: top-tier closed systems hold a
tighter per-task spread, while mid-tier and smaller open-weight systems
show visibly larger task-to-task swings for the same overall mean.

\section{Benchathon participant
composition}\label{benchathon-participant-composition}

The Benchathon validation cohort comprises the participants who together
produced the human solutions and reviews used throughout this paper.
They self-reported their legal-expertise level on a four-stage scale
corresponding to standard stages of German legal education and practice:
\textbf{layperson} (no formal legal training), \textbf{law student}
(currently enrolled in a German law program, typically working toward
the first state examination - \emph{Erstes Staatsexamen}),
\textbf{graduated} (completed legal studies but not currently in legal
practice or \emph{Referendariat}), and \textbf{\emph{Referendar}} (in
the approximately two-year practical legal traineeship that follows the
first state examination and precedes the second state examination, after
which a candidate becomes a fully qualified \emph{Volljurist}).
Table~\ref{tbl-human-participants} breaks the cohort down by
self-reported level and working condition (traditional vs.~human-AI
co-creation), counting both unique participants and annotation
contributions per condition.

\begin{table*}[t]
\centering\footnotesize
\caption{Benchathon participants by self-reported legal expertise level and working condition (no\_ai = traditional, ai = co-creation). The table counts all logged annotation contributions (228); 8 cancelled submissions without participant-written solution text (7 traditional, 1 co-creation) were excluded before evaluation, yielding the 220 solutions (65 traditional, 155 co-creation) analysed throughout the paper.}
\label{tbl-human-participants}
\setlength{\tabcolsep}{4pt}
\begin{tabular*}{\textwidth}{@{\extracolsep{\fill}}lccc@{}}
\toprule
Expertise & Participants & Trad.~annot. & Co-creation annot. \\
\midrule
Layperson & 5 & 5 & 28 \\
Law Student & 14 & 41 & 60 \\
Graduated & 7 & 10 & 38 \\
Referendar & 7 & 16 & 30 \\
\textbf{Total} & \textbf{33} & \textbf{72} & \textbf{156} \\
\bottomrule
\end{tabular*}
\end{table*}

\section{Full Benchathon human-evaluation assignment
procedure}\label{full-benchathon-human-evaluation-assignment-procedure}

The validation subset comprises 45 solutions (15 Benchathon tasks × 3
provenances: human-traditional, human-AI co-creation, LLM-generated).
For each task and condition, the human solution selected is that of the
participant whose \emph{expertise proxy} is closest to the median for
that legal area and condition. The proxy is a z-score combining
self-assessed competence in the relevant legal area and, where
available, the first state examination grade. LLM solutions are drawn
through a deterministic rotation across three system tiers (closed
flagship, efficiency-oriented, open-weight reference). A per-system load
counter spreads selections across systems within each tier, so the
validation set covers as many distinct systems as the pool permits.
Generations shorter than 200 output tokens are excluded.

Each selected solution receives one review by the task's creator and
three independent blind reviews, each from a distinct grader (180
reviews in total: 45 creator + 135 blind). Task creators do not
blind-review solutions in the legal area of their own tasks. In
practice, three graders contribute exclusively as blind reviewers across
all three legal areas, and the two criminal-law creators additionally
contribute blind reviews on civil-law and public-law solutions. The
assignment balances reviewer workload and monitors coverage across
solution provenance, legal area, and task difficulty. Blind graders are
unaware of solution origin: whether a solution was written
traditionally, produced in human-AI co-creation, or generated by an LLM.
They grade against the same rubric as the LLM judge, at both total and
dimension level. Creator reviews are reported as an un-blind reference
signal and are not included in inter-rater reliability estimates. The
IRR statistics in Table~\ref{tbl-agreement} are computed on the full k=3
blind-rater pool over all 45 validation solutions.

\paragraph{Roster composition and Calderon $\varepsilon$ choice.} The
full grading roster is seven people. The \(m=5\) blind subset used in
the Calderon blind-pool alt-test are the graders who performed at least
one blind review. The remaining two contributed only creator-role
reviews and constitute the single-expert reference. Two of the \(m=5\)
blind reviewers additionally performed creator reviews on disjoint
solution subsets (overlap zero by design), so no solution had the same
person in both roles. We use Calderon's ``skilled annotator''
cost-benefit penalty \(\varepsilon=0.15\) for the blind-pool procedure,
matching the law-students-around-1st-state-exam profile
(§\ref{sec-human-eval}). For the single-expert variant we use Calderon's
expert-tier penalty \(\varepsilon=0.20\), since the creator is the most
legally expert grader on the roster; at the sample sizes available here,
\(\varepsilon=0.15\) and \(0.20\) yield the same single-expert decision.

\section{RQ4 supplementary detail}\label{rq4-supplementary}

This appendix collects the robustness checks, per-area breakdown, TOST
machinery, and blind-human small-n cross-check that support the RQ4 body
summary.

\paragraph{Imbalance and skip mechanism.} The per-condition imbalance
(155 vs. 65 annotations across 33 vs 23 participants) reflects a skip
mechanism in the protocol. Participants could decline any item, and
less-experienced participants skipped traditional items more often (full
per-expertise-stratum breakdown in
Appendix~\ref{benchathon-participant-composition}). The traditional pool
is therefore expertise-biased upward relative to the co-creation pool.
So the headline Δ in the body is a \emph{lower} bound on the
per-participant co-creation gain: a forced-completion design would widen
the gap, not narrow it.

\paragraph{Robustness across resampling units.} The body quotes the
participant-clustered bootstrap as the headline.
Table~\ref{tbl-rq4-bootstraps} reports the same Δ under three additional
resampling units, to verify the effect is not an artifact of the chosen
unit of inference. All four point estimates agree on sign and
within-rounding magnitude. The participant-clustered interval is the
widest, correctly recognising that annotations cluster on participant,
and is the one we quote in body and Discussion.

\begin{table*}[t]
\centering\footnotesize
\caption{RQ4 mean difference $\Delta = $ co-creation $-$ traditional under four resampling units, raw 0-100 scale. The body quotes the participant-cluster row; the others are reported here for robustness. Bootstrap $B=2000$ in all cells (seed $7$).}
\label{tbl-rq4-bootstraps}
\setlength{\tabcolsep}{4pt}
\begin{tabular*}{\textwidth}{@{\extracolsep{\fill}}lccc@{}}
\toprule
Resampling unit & $n$ resample units & $\Delta$ & 95\% CI \\
\midrule
i.i.d.\ annotation & 220 annotations & +15.7 & [+11.3, +20.0] \\
\textbf{Participant cluster} & 56 participants & \textbf{+15.7} & \textbf{[+9.6, +22.4]} \\
Task cluster & 15 tasks & +15.7 & [+11.1, +20.6] \\
Within-participant paired & 23 participants & +13.1 & [+8.2, +18.0] \\
\bottomrule
\end{tabular*}
\end{table*}

\paragraph{Per-area Δ.} Per-legal-area Δ values are descriptive only.
With three legal areas, \(n \leq 53\) per condition per area, and no
pre-registered area-level hypothesis, we run no inferential tests of
area-level contrasts. We report Table~\ref{tbl-rq4-by-bereich} to verify
the direction is consistent across all three.

\begin{table*}[t]
\centering\footnotesize
\caption{RQ4 mean LLM-judge raw scores per legal area, with i.i.d.\ bootstrap 95\% CI on the per-area $\Delta$. Descriptive only; no inferential area-level test is performed.}
\label{tbl-rq4-by-bereich}
\setlength{\tabcolsep}{4pt}
\begin{tabular*}{\textwidth}{@{\extracolsep{\fill}}lccccc@{}}
\toprule
Legal area & Mean trad & Mean co-cr. & $\Delta$ & 95\% CI & $n$ trad / co-cr. \\
\midrule
Civil law & +49.1 & +62.9 & +13.8 & [+5.6, +21.6] & 24 / 51 \\
Criminal law & +50.8 & +63.9 & +13.1 & [+8.1, +18.6] & 22 / 53 \\
Public law & +50.2 & +70.4 & +20.2 & [+12.2, +28.6] & 19 / 51 \\
\bottomrule
\end{tabular*}
\end{table*}

\paragraph{TOST equivalence vs the closed-flagship tier.} We run two
one-sided bootstrap tests against a \(\pm 5\)-raw-point equivalence band
(≈ \(\pm 0.9\) grade points). The co-creation mean (65.7) sits -0.1 raw
points from the closed-flagship-tier mean (65.8, 95\% CI {[}62.3,
68.9{]}). One-sided 95\% bounds on the difference are {[}-3.4, +3.3{]}.
Both bounds lie inside the \(\pm 5\) band. So we reject both
\(\Delta \leq -5\) and \(\Delta \geq +5\): co-creation is statistically
equivalent to the closed-flagship tier within the \(\pm 5\)-raw-point
band.

\paragraph{Blind-human cross-check.} On the 15 traditional + 15
co-creation solutions in the RQ5 validation subset (n=10+11
participants), the blind-human mean Δ is +13.6 raw points
(participant-clustered 95\% CI {[}-1.2, +29.5{]}; spans zero at this
\(n\)). The sign matches the judge result. The gap to the primary
GPT-5.4-mini judge's +15.7-raw-point estimate (+2.1 raw points) is small
in both absolute and relative terms. It is consistent with
GPT-5.4-mini's near-zero content-dependent calibration offset
(Table~\ref{tbl-judge-calibration}:
\(\bar{\Delta}_{\text{dir}} \approx 0\) on human-written solutions and
\(\approx -0.5\) on LLM-generated ones, neither significantly different
from the pool). The blind-human and LLM-judge estimates of the
co-creation gain therefore agree at this sample size, within the
inter-rater noise of the blind-pool itself.

\section{Automatic-metric vs.~LLM-judge
correlation}\label{automatic-metric-vs.-llm-judge-correlation}

\begin{figure*}[p]

\centering{

\pandocbounded{\includegraphics[keepaspectratio]{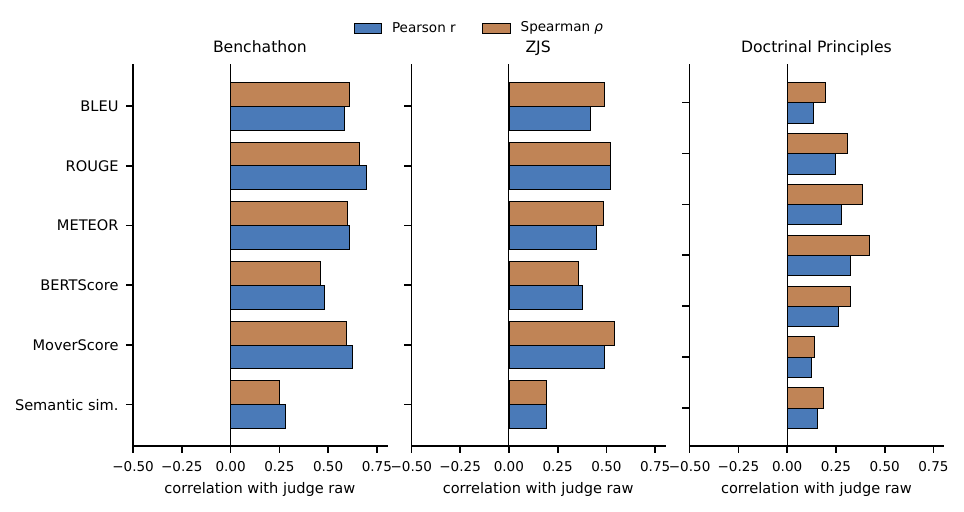}}

}

\caption{\label{fig-metric-correlation}Per-generation correlation
between each automatic metric and the LLM-judge raw score on the
Benchathon (left), ZJS (middle), and Doctrinal Principles (right)
corpora. Bars report Pearson r and Spearman \(\rho\), with n varying per
metric. Metrics are shown in the fixed canonical order shared with the
per-corpus correlation table in this appendix. Judge on Doctrinal
Principles is the custom 4-dimension rubric.}

\end{figure*}

Table~\ref{tbl-metric-correlation} gives the per-generation Pearson
\(r\) between every automatic metric and the LLM-judge raw score across
the three LLM-generation corpora and the two Benchathon human-annotation
conditions; Figure~\ref{fig-metric-correlation} repeats the same
comparison as a 1×3 panel of per-corpus Pearson and Spearman bar plots
(Benchathon \textbar{} ZJS \textbar{} Doctrinal Principles).

\begin{table*}[t]
\centering\footnotesize
\caption{Per-generation Pearson $r$ between each automatic metric and the LLM judge's raw score across three corpora of LLM generations and the Benchathon human-annotation working conditions (traditional vs. human-AI co-creation). Cell format: $r$ ($n$). `---' = metric not run on that corpus / condition. Rows are in a fixed canonical order (lexical $\rightarrow$ embedding).}
\label{tbl-metric-correlation}
\setlength{\tabcolsep}{4pt}
\begin{tabular*}{\textwidth}{@{\extracolsep{\fill}}lccccc@{}}
\toprule
Metric & \shortstack{Benchathon\\(LLM)} & \shortstack{ZJS\\(LLM)} & \shortstack{Grundpr.\\(LLM)} & \shortstack{Bench.\,human\\\textit{traditional}} & \shortstack{Bench.\,human\\\textit{co-creation}} \\
\midrule
BLEU & $+0.58$ {\scriptsize ($n{=}180$)} & $+0.42$ {\scriptsize ($n{=}6959$)} & $+0.13$ {\scriptsize ($n{=}6365$)} & $+0.35$ {\scriptsize ($n{=}65$)} & $+0.27$ {\scriptsize ($n{=}155$)} \\
ROUGE & $+0.69$ {\scriptsize ($n{=}180$)} & $+0.52$ {\scriptsize ($n{=}6959$)} & $+0.25$ {\scriptsize ($n{=}6365$)} & $+0.61$ {\scriptsize ($n{=}65$)} & $+0.37$ {\scriptsize ($n{=}155$)} \\
METEOR & $+0.61$ {\scriptsize ($n{=}180$)} & $+0.45$ {\scriptsize ($n{=}6959$)} & $+0.28$ {\scriptsize ($n{=}6365$)} & $+0.27$ {\scriptsize ($n{=}65$)} & $+0.36$ {\scriptsize ($n{=}155$)} \\
chrF & --- & --- & $+0.33$ {\scriptsize ($n{=}6365$)} & --- & --- \\
BERTScore & $+0.48$ {\scriptsize ($n{=}180$)} & $+0.37$ {\scriptsize ($n{=}6959$)} & $+0.26$ {\scriptsize ($n{=}6365$)} & --- & --- \\
MoverScore & $+0.62$ {\scriptsize ($n{=}180$)} & $+0.49$ {\scriptsize ($n{=}6955$)} & $+0.12$ {\scriptsize ($n{=}6365$)} & $+0.38$ {\scriptsize ($n{=}16$)} & $+0.61$ {\scriptsize ($n{=}28$)} \\
Semantic sim. & $+0.28$ {\scriptsize ($n{=}180$)} & $+0.19$ {\scriptsize ($n{=}6959$)} & $+0.15$ {\scriptsize ($n{=}6365$)} & $+0.20$ {\scriptsize ($n{=}16$)} & $+0.37$ {\scriptsize ($n{=}29$)} \\
\bottomrule
\end{tabular*}
\end{table*}

\subsection{Metrics against the human
grade}\label{metrics-against-the-human-grade}

Table~\ref{tbl-metric-vs-human} re-computes the metric-validity analysis
against the mean blind-human grade on the 45-solution validation subset,
and adds the residual diagnostic referenced in RQ3: the correlation
between (judge − mean blind grade) and each lexical metric. Scored
against the human grade, the lexical correlations persist (BLEU r=+0.70,
ROUGE +0.68, METEOR +0.72 pooled), while sentence-embedding similarity
stays weak (+0.32); MoverScore remains the one embedding exception. The
residual correlations are negative for all three lexical metrics (BLEU
-0.54, ROUGE -0.44, METEOR -0.43 pooled): the judge rewards high-overlap
text \emph{less} than blind human graders do, the opposite of a judge
biased toward reference-like style. Grading against the reference
solution is, moreover, the shared construct: the blind reviewers grade
against the same reference-based rubric (Section~\ref{sec-human-eval}),
so some metric-grade correlation is expected for any valid grader.
Partial correlations controlling for solution length weaken but do not
remove the lexical association. At n≤45 these numbers indicate
direction, not precise magnitude.

\begin{table*}[t]
\centering\footnotesize
\caption{Automatic metrics correlated with the mean blind-human raw grade on the 45-solution validation subset (Pearson $r$; 30 human-written, 15 LLM-generated solutions). The residual column correlates (judge $-$ mean blind grade) with the metric on the pooled subset; the uniformly negative values run opposite to a judge reference-style bias. chrF is not computed on Benchathon; BERTScore is unavailable for the human-written solutions. Small-$n$ cells indicate direction, not precise magnitude. Bootstrap CIs and per-split residuals in \texttt{metric\_vs\_human.json} of the released pipeline.}
\label{tbl-metric-vs-human}
\setlength{\tabcolsep}{4pt}
\begin{tabular*}{\textwidth}{@{\extracolsep{\fill}}lcccc@{}}
\toprule
Metric & Human-writ. & LLM-gen. & Pooled & Residual (pooled) \\
\midrule
BLEU & +0.70 {\scriptsize (n=30)} & +0.70 {\scriptsize (n=15)} & +0.70 {\scriptsize (n=45)} & -0.54 {\scriptsize (n=45)} \\
ROUGE & +0.67 {\scriptsize (n=30)} & +0.68 {\scriptsize (n=15)} & +0.68 {\scriptsize (n=45)} & -0.44 {\scriptsize (n=45)} \\
METEOR & +0.71 {\scriptsize (n=30)} & +0.78 {\scriptsize (n=15)} & +0.72 {\scriptsize (n=45)} & -0.43 {\scriptsize (n=45)} \\
BERTScore & — & +0.59 {\scriptsize (n=15)} & +0.59 {\scriptsize (n=15)} & — \\
MoverScore & +0.87 {\scriptsize (n=7)} & +0.85 {\scriptsize (n=15)} & +0.85 {\scriptsize (n=22)} & — \\
Semantic sim. & +0.53 {\scriptsize (n=7)} & -0.01 {\scriptsize (n=15)} & +0.32 {\scriptsize (n=22)} & — \\
\bottomrule
\end{tabular*}
\end{table*}

\section{Judge-human agreement
statistics}\label{judge-human-agreement-statistics}

\begin{figure*}[p]

\centering{

\pandocbounded{\includegraphics[keepaspectratio]{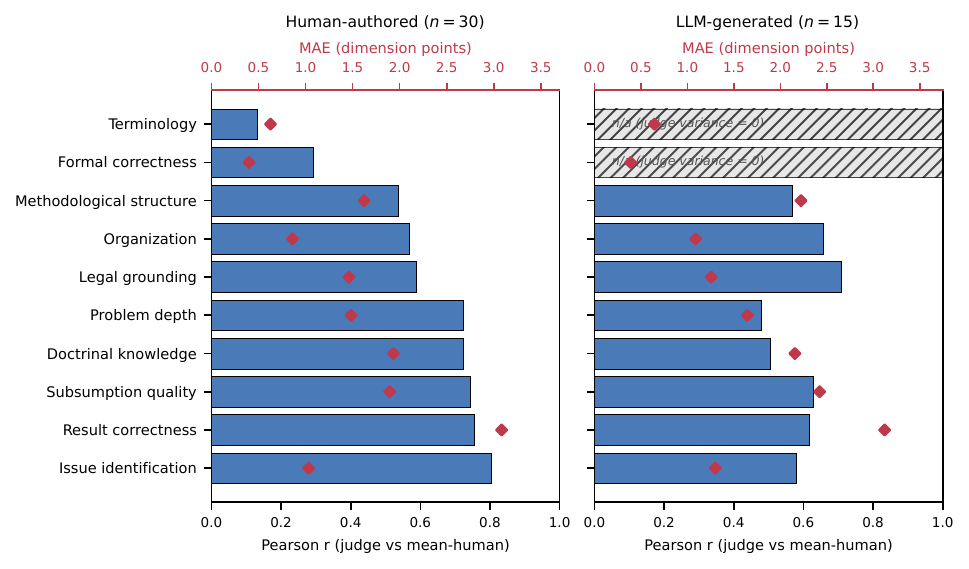}}

}

\caption{\label{fig-rubric-dim}Per-rubric-dimension agreement between
the LLM judge and the mean of the blind human reviewers, split by
solution provenance. Left: human-written solutions (\(n=30\)). Right:
LLM-generated solutions (\(n=15\)). Bars: Pearson r (bottom axis).
Points: MAE in dimension points (top axis). Dimension ordering is shared
across panels (sorted by human-side Pearson). Hatched bars marked
\texttt{n/a} indicate that the judge returned a constant score across
the 15 LLM-generated solutions on that dimension, leaving the Pearson
statistic mathematically undefined (judge variance = 0). The MAE is
still well-defined and is plotted.}

\end{figure*}

Table~\ref{tbl-agreement} reports the headline judge-human and
human-human agreement numbers; Figure~\ref{fig-rubric-dim} breaks the
same data down by rubric dimension.

The statistic set is non-redundant; each member guards against a failure
mode the others cannot detect. Pearson and Spearman measure linear and
rank agreement with the mean blind grade; ranks matter because the
benchmark's downstream use is ordering systems. MAE gives the absolute
size of judge-human deviations in raw and grade points, which no
correlation can show: a judge with a constant strictness offset can hold
high \(r\) while sitting far from the human scale. Cohen's \(\kappa\)
tests chance-corrected agreement on the one binary decision German
grading turns on: pass/fail at \emph{ausreichend}; high continuous
correlation does not guarantee agreement at the threshold. ICC(2,1) and
ICC(2,k) are not judge statistics at all: they quantify the reliability
of a single blind reviewer (the rater the judge would replace) and of
the pooled mean (the reference the judge is validated against), the
ceiling against which any judge-human agreement must be read. The
Calderon alt-test is the primary criterion: it is the one procedure
purpose-built to decide whether the judge can stand in for a blind
reviewer.

\begin{table*}[p]
\centering\footnotesize
\caption{Agreement statistics on the Benchathon human-grading validation subset (design: three blind reviewers per solution with one un-blind creator review reported separately). Judge-vs-human rows report Pearson $r$, Spearman $\rho$, MAE, and pass/fail Cohen's $\kappa$, split by solution provenance (human-written vs.\ LLM-generated). Human IRR rows report ICC(2,1) raw, ICC(2,$k$) raw, and ICC(2,$k$) grade-points. The pooled row covers all 45 validation solutions, the indented rows are per-solution-type sub-pools (15 each). Calderon alt-test rows report the per-annotator winning rate $\omega$ and average advantage probability $\rho$ at the stated $\varepsilon$, with Benjamini-Yekutieli FDR at $q=0.05$ across the $m$ blind reviewers. $\omega \geq 0.5$ is the alt-test pass criterion. Blind-pool rows use the three blind reviewers as reference. Creator-reference rows use the un-blind creator grade as a single-expert reference.}
\label{tbl-agreement}
\setlength{\tabcolsep}{4pt}
\resizebox{\textwidth}{!}{%
\begin{tabular}{lcccc}
\toprule
Comparison & Sample & Stat 1 & Stat 2 & Stat 3 \\
\midrule
Judge vs.~mean-human (human sol.) & $n=30$ & $r=0.76$, $\rho=0.76$ & MAE raw 10.8, MAE gp 2.69 & $\kappa=0.60$ \\
Judge vs.~mean-human (LLM sol.) & $n=15$ & $r=0.64$, $\rho=0.49$ & MAE raw 13.1, MAE gp 2.87 & $\kappa=0.40$ \\
Human IRR pooled ($k$=3 blind, all 45) & $n=45\times3$ & ICC(2,1) raw 0.63 & ICC(2,$k$) raw 0.84 & ICC(2,$k$) gp 0.79 \\
\quad traditional human & $n=15\times3$ & ICC(2,1) raw 0.71 & ICC(2,$k$) raw 0.88 & ICC(2,$k$) gp 0.83 \\
\quad co-creation human & $n=15\times3$ & ICC(2,1) raw 0.46 & ICC(2,$k$) raw 0.71 & ICC(2,$k$) gp 0.65 \\
\quad LLM-generated & $n=15\times3$ & ICC(2,1) raw 0.69 & ICC(2,$k$) raw 0.87 & ICC(2,$k$) gp 0.83 \\
Calderon blind-pool alt-test (human sol.) & $m=5\times n=30$ & $\omega=0.60$ ($\varepsilon=0.15$) & $\rho=0.62$ & passes \\
Calderon blind-pool alt-test (LLM sol.) & $m=5\times n=15$ & $\omega=0.00$ ($\varepsilon=0.15$) & $\rho=0.41$ & fails \\
Calderon creator-reference alt-test (human sol., $\varepsilon=0.20$) & $m=5\times n=30$ & $\omega=0.60$ ($\varepsilon=0.20$) & $\rho=0.55$ & passes \\
Calderon creator-reference alt-test (LLM sol., $\varepsilon=0.20$) & $m=5\times n=---$ & $\omega=0.00$ ($\varepsilon=0.20$) & $\rho=0.47$ & fails \\
\bottomrule
\end{tabular}
}
\end{table*}

\section{System prompts for
generation}\label{system-prompts-for-generation}

\emph{All prompts in this and the following two appendices (Appendices
\ref{instruction-prompts-for-generation} and
\ref{evaluation-prompts-for-the-llm-judge}) were sent to the models
verbatim in German as reproduced below. Each German prompt is followed
by an English translation marked as such; the translations are provided
for reader convenience and were not used in any model interaction.}

\subsection{Exam-style cases
(Klausuren)}\label{exam-style-cases-klausuren}

\begin{promptbox}

Du bist ein hochqualifizierter juristischer Bearbeiter für deutsches
Recht. Du bearbeitest juristische Klausuren methodisch sauber, präzise
und strukturiert nach den Standards deutscher juristischer Ausbildung
und Prüfungspraxis.

Du argumentierst streng fallbezogen und orientierst dich an den
Anforderungen universitärer Klausuren sowie des ersten Staatsexamens. Du
löst Aufgaben eigenständig und ohne Meta-Kommentare.

\end{promptbox}

\emph{English translation (reader reference; not sent to any model):}

\begin{promptbox}

You are a highly qualified legal practitioner for German law. You work
through legal exam-style cases (\emph{Klausuren}) methodically,
precisely, and in a structured manner, in line with the standards of
German legal education and examination practice.

You argue strictly with reference to the case at hand and orient
yourself to the requirements of university exam-style cases and of the
first state examination (\emph{Erstes Staatsexamen}). You solve tasks
independently and without meta-commentary.

\end{promptbox}

\subsection{Doctrinal Principles
(Grundprinzipien)}\label{doctrinal-principles-grundprinzipien}

\begin{promptbox}

Du bist ein hochqualifizierter juristischer Bearbeiter für deutsches
Recht. Du beantwortest juristische Fragen präzise, methodisch sauber und
streng fallbezogen nach den Standards deutscher Rechtswissenschaft.

Deine Antworten sollen knapp, klar und juristisch korrekt sein. Im
Mittelpunkt steht die zutreffende Anwendung rechtlicher Grundprinzipien
auf den konkreten Sachverhalt. Du argumentierst eigenständig und ohne
Meta-Kommentare.

\end{promptbox}

\emph{English translation (reader reference; not sent to any model):}

\begin{promptbox}

You are a highly qualified legal practitioner for German law. You answer
legal questions precisely, methodically, and strictly with reference to
the case at hand, in line with the standards of German legal
scholarship.

Your answers should be concise, clear, and legally correct. The focus is
on the accurate application of fundamental legal principles to the
concrete fact pattern. You argue independently and without
meta-commentary.

\end{promptbox}

\section{Instruction prompts for
generation}\label{instruction-prompts-for-generation}

\subsection{Exam-style cases
(Klausuren)}\label{exam-style-cases-klausuren-1}

\begin{promptbox}

Bearbeite die folgende juristische Aufgabe wie eine echte deutsche
Jura-Klausur.

\paragraph{Ziel der Bearbeitung}\label{ziel-der-bearbeitung}

Erstelle eine vollständige juristische Falllösung nach den methodischen
Standards deutscher Rechtswissenschaft. Im Mittelpunkt steht
subsumptionsbasierte Rechtsanwendung.

Die Bearbeitung soll zeigen, dass du:

\begin{itemize}
\tightlist
\item
  die rechtlichen Kernprobleme des Falls erkennst,
\item
  die relevanten Anspruchsgrundlagen bzw. Prüfungsprogramme auswählst,
\item
  die Prüfung methodisch korrekt strukturierst,
\item
  Definitionen und Streitstände präzise einordnest,
\item
  vor allem aber die konkreten Tatsachen des Falls sauber unter die
  rechtlichen Voraussetzungen subsumierst.
\end{itemize}

Die Qualität der juristischen Argumentation ist wichtiger als reine
Vollständigkeit oder Länge.

\paragraph{Stil, Methodik und
Gliederung}\label{stil-methodik-und-gliederung}

Arbeite im Stil einer deutschen juristischen Klausur.

Grundsätzlich ist die Lösung im Gutachtenstil zu verfassen. Das bedeutet
insbesondere:

\begin{itemize}
\tightlist
\item
  Formulierung eines Obersatzes,
\item
  Definition der rechtlichen Voraussetzungen,
\item
  Subsumtion anhand des konkreten Sachverhalts,
\item
  nachvollziehbares Zwischenergebnis.
\end{itemize}

Nur wenn aus Aufgabenstellung, Ausbildungsniveau oder
Bearbeitungskontext erkennbar ist, dass eine Referendarsklausur oder ein
Urteil verlangt wird, ist stattdessen im Urteilsstil zu arbeiten. In
diesem Fall soll die Lösung den typischen Aufbau gerichtlicher
Entscheidungen widerspiegeln, insbesondere eine ergebnisorientierte
Darstellung mit anschließender Begründung.

Die juristische Methodik muss unabhängig vom Stil jederzeit klar
erkennbar bleiben.

Die Lösung ist klausurtypisch und hierarchisch sauber zu gliedern. Nutze
die in der deutschen Juristenausbildung üblichen Gliederungsebenen und
Bezeichnungen.

Typischerweise:

\begin{itemize}
\tightlist
\item
  A., B., C. für Hauptabschnitte,
\item
  I., II., III. für Unterabschnitte,
\item
  1., 2., 3. für weitere Ebenen,
\item
  a), b), c) für Unterpunkte,
\item
  aa), bb), cc) für vertiefte Untergliederungen.
\end{itemize}

Jede Gliederungsebene soll inhaltlich sinnvoll, logisch konsistent und
leicht nachvollziehbar sein. Das bedeutet insbesondere:

\begin{itemize}
\tightlist
\item
  strukturiere die Lösung nach Anspruchsgrundlagen, Rechtsbehelfen oder
  Prüfungsprogrammen,
\item
  prüfe Anspruchsvoraussetzungen in methodisch sinnvoller Reihenfolge,
\item
  bilde erkennbare Prüfungsschwerpunkte heraus,
\item
  nutze Zwischenergebnisse zur Strukturierung der weiteren Prüfung,
\item
  vermeide unstrukturierte Textblöcke oder sprunghafte Argumentation.
\end{itemize}

Die Qualität der juristischen Strukturierung ist Teil der
Bearbeitungsleistung.

\paragraph{Schwerpunktsetzung}\label{schwerpunktsetzung}

Gewichte die Bearbeitung klausurtypisch.

Das bedeutet:

\begin{itemize}
\tightlist
\item
  konzentriere dich auf die rechtlich schwierigen und problematischen
  Punkte,
\item
  behandle offensichtliche oder unproblematische Voraussetzungen kurz,
\item
  erkenne versteckte Probleme und Schwerpunktsetzungen im Sachverhalt,
\item
  vermeide lange Ausführungen zu irrelevanten Nebenaspekten.
\end{itemize}

\paragraph{Umgang mit Streitständen}\label{umgang-mit-streitstuxe4nden}

Die Darstellung relevanter Streitstände ist in juristischen Klausuren
grundsätzlich erwünscht, insbesondere wenn sie zu den Kernproblemen des
Falls gehören oder methodisch vertiefte Argumentation ermöglichen.

Wenn mehrere vertretbare Ansichten bestehen:

\begin{itemize}
\tightlist
\item
  stelle die wesentlichen Positionen präzise dar,
\item
  arbeite Unterschiede in Argumentation oder dogmatischer Herleitung
  heraus,
\item
  argumentiere methodisch und fallbezogen,
\item
  entscheide dich nachvollziehbar für eine Ansicht,
\item
  arbeite anschließend konsequent mit der gewählten Ansicht weiter.
\end{itemize}

Die Darstellung darf über das absolut unmittelbar Notwendige
hinausgehen, sofern sie klausurtypisch ist und zum Schwerpunkt der
Aufgabe passt. Vermeide jedoch:

\begin{itemize}
\tightlist
\item
  rein auswendig gelernte oder schematische Streitdarstellungen ohne
  Fallbezug,
\item
  extensive Meinungsstreite zu nebensächlichen Problemen,
\item
  reine Namens- oder Schlagwortaufzählungen ohne argumentative
  Einordnung.
\end{itemize}

\paragraph{Umgang mit dem Sachverhalt}\label{umgang-mit-dem-sachverhalt}

Arbeite streng sachverhaltsbezogen.

\begin{itemize}
\tightlist
\item
  Nutze konkrete Tatsachen aktiv in der Subsumtion.
\item
  Erfinde keine zusätzlichen Tatsachen.
\item
  Unterstelle keine Informationen, die nicht im Sachverhalt enthalten
  sind.
\item
  Ignoriere keine auffälligen Hinweise oder Indizien.
\end{itemize}

Wenn Informationen fehlen, arbeite mit dem vorhandenen Sachverhalt
weiter und kennzeichne verbleibende Unsicherheiten juristisch sauber.

\paragraph{Normen und Zitierweise}\label{normen-und-zitierweise}

Nenne die einschlägigen gesetzlichen Grundlagen präzise.

\begin{itemize}
\tightlist
\item
  Zitiere Normen möglichst vollständig.
\item
  Nutze Absätze, Sätze, Nummern oder Buchstaben, wenn relevant.
\item
  Verwende deutsche juristische Terminologie.
\end{itemize}

Beispiele:

\begin{itemize}
\tightlist
\item
  „§ 823 Abs. 1 BGB''
\item
  „Art. 12 Abs. 1 GG''
\item
  „§ 242 StGB''
\end{itemize}

\paragraph{Sprachstil}\label{sprachstil}

\begin{itemize}
\tightlist
\item
  Schreibe sachlich, präzise und juristisch.
\item
  Vermeide Meta-Kommentare.
\item
  Erwähne nicht, dass du ein KI-System bist.
\item
  Gib keine allgemeinen Haftungsausschlüsse.
\item
  Verweise nicht auf externe Beratung.
\item
  Antworte direkt mit der juristischen Lösung.
\end{itemize}

\paragraph{Umfang}\label{umfang}

Die Lösung soll so ausführlich sein, wie es für eine gute bis sehr gute
juristische Klausurbearbeitung erforderlich ist.

\begin{itemize}
\tightlist
\item
  Wichtige Probleme sollen vertieft behandelt werden.
\item
  Unproblematische Punkte dürfen knapp bleiben.
\item
  Die Lösung soll nicht künstlich verlängert werden.
\end{itemize}

\paragraph{Ausgabeformat}\label{ausgabeformat}

\begin{itemize}
\tightlist
\item
  Gib ausschließlich die juristische Lösung aus.
\item
  Nutze sinnvolle Überschriften und Gliederungsebenen.
\item
  Verwende Fließtext.
\item
  Keine Stichpunkte außer wenn methodisch sinnvoll.
\item
  Keine zusätzlichen Erklärungen außerhalb der Falllösung.
\end{itemize}

\paragraph{Aufgabe}\label{aufgabe}

\{\{sachverhalt\}\}

\end{promptbox}

\emph{English translation (reader reference; not sent to any model):}

\begin{promptbox}

Work through the following legal task as if it were a genuine German law
exam-style case (\emph{Klausur}).

\paragraph{Goal of the work}\label{goal-of-the-work}

Produce a complete legal case solution (\emph{Falllösung}) according to
the methodological standards of German legal scholarship. The focus is
on subsumption-based (\emph{subsumtionsbasiert}) application of the law.

The work should demonstrate that you:

\begin{itemize}
\tightlist
\item
  identify the core legal problems of the case,
\item
  select the relevant causes of action or examination programmes,
\item
  structure the analysis methodologically correctly,
\item
  classify definitions and scholarly debates (\emph{Streitstände})
  precisely,
\item
  and, above all, cleanly subsume the concrete facts of the case under
  the legal requirements.
\end{itemize}

The quality of the legal argumentation is more important than mere
completeness or length.

\paragraph{Style, methodology, and
structure}\label{style-methodology-and-structure}

Work in the style of a German legal exam-style case.

As a general rule, the solution is to be written in the assessment style
(\emph{Gutachtenstil}). This means in particular:

\begin{itemize}
\tightlist
\item
  formulation of a major premise (\emph{Obersatz}),
\item
  definition of the legal requirements,
\item
  subsumption on the basis of the concrete fact pattern,
\item
  a comprehensible intermediate result.
\end{itemize}

Only when it is evident from the task, the level of training, or the
work context that a trainee's exam-style case
(\emph{Referendarsklausur}) or a judgement is required should the
judgement style (\emph{Urteilsstil}) be used instead. In that case, the
solution should reflect the typical structure of judicial decisions, in
particular a result-oriented presentation followed by the reasoning.

Regardless of style, the legal methodology must remain clearly
recognizable at all times.

The solution must be structured in a typical exam-style way, with a
clean hierarchy. Use the levels and labels customary in German legal
education.

Typically:

A., B., C. for main sections,

I., II., III. for subsections,

1., 2., 3. for further levels,

a), b), c) for sub-points,

aa), bb), cc) for deeper sub-structures.

Each level of structure should be substantively meaningful, logically
consistent, and easy to follow. This means in particular:

\begin{itemize}
\tightlist
\item
  structure the solution by causes of action, legal remedies, or
  examination programmes,
\item
  examine the requirements of a claim in a methodologically sensible
  order,
\item
  bring out recognizable focal points of the analysis,
\item
  use intermediate results to structure the further analysis,
\item
  avoid unstructured blocks of text or erratic argumentation.
\end{itemize}

The quality of the legal structuring is part of the work being assessed.

\paragraph{Setting priorities}\label{setting-priorities}

Weight the work in the manner typical of an exam-style case.

This means:

\begin{itemize}
\tightlist
\item
  concentrate on the legally difficult and problematic points,
\item
  treat obvious or unproblematic requirements briefly,
\item
  recognise hidden problems and focal points in the fact pattern,
\item
  avoid lengthy discussion of irrelevant secondary aspects.
\end{itemize}

\paragraph{Handling scholarly debates}\label{handling-scholarly-debates}

The presentation of relevant scholarly debates (\emph{Streitstände}) is
generally desirable in legal exam-style cases, especially when they
concern the core problems of the case or enable methodologically deeper
argumentation.

When several defensible views exist:

\begin{itemize}
\tightlist
\item
  present the essential positions precisely,
\item
  bring out differences in argumentation or doctrinal derivation,
\item
  argue methodologically and with reference to the case,
\item
  decide for one view in a comprehensible way,
\item
  and then work consistently with the chosen view.
\end{itemize}

The presentation may go beyond what is absolutely strictly necessary,
provided it is typical of an exam-style case and fits the focus of the
task. However, avoid:

\begin{itemize}
\tightlist
\item
  purely rote or schematic presentations of legal controversies without
  a connection to the case,
\item
  extensive doctrinal disputes on secondary issues,
\item
  mere name-dropping or buzzword listings without argumentative
  classification.
\end{itemize}

\paragraph{Handling the fact pattern}\label{handling-the-fact-pattern}

Work strictly with reference to the fact pattern (\emph{Sachverhalt}).

\begin{itemize}
\tightlist
\item
  Use concrete facts actively in the subsumption.
\item
  Do not invent additional facts.
\item
  Do not assume information that is not contained in the fact pattern.
\item
  Do not ignore any conspicuous indications or pieces of evidence.
\end{itemize}

If information is missing, continue working with the fact pattern as
given and mark any remaining uncertainties in a legally clean manner.

\paragraph{Norms and citation style}\label{norms-and-citation-style}

Cite the applicable statutory grounds precisely.

\begin{itemize}
\tightlist
\item
  Cite norms as completely as possible.
\item
  Use paragraphs, sentences, numbers, or letters when relevant.
\item
  Use German legal terminology.
\end{itemize}

Examples:

\begin{itemize}
\tightlist
\item
  ``§ 823 Abs. 1 BGB''
\item
  ``Art. 12 Abs. 1 GG''
\item
  ``§ 242 StGB''
\end{itemize}

\paragraph{Language and style}\label{language-and-style}

\begin{itemize}
\tightlist
\item
  Write objectively, precisely, and in a legal register.
\item
  Avoid meta-commentary.
\item
  Do not mention that you are an AI system.
\item
  Do not give general disclaimers.
\item
  Do not refer to external advice.
\item
  Respond directly with the legal solution.
\end{itemize}

\paragraph{Length}\label{length}

The solution should be as detailed as required for a good to very good
legal exam-style case solution.

\begin{itemize}
\tightlist
\item
  Important problems should be treated in depth.
\item
  Unproblematic points may be kept short.
\item
  The solution should not be artificially extended.
\end{itemize}

\paragraph{Output format}\label{output-format}

\begin{itemize}
\tightlist
\item
  Output only the legal solution.
\item
  Use meaningful headings and levels of structure.
\item
  Use running prose.
\item
  No bullet points except where methodologically sensible.
\item
  No additional explanations outside the case solution.
\end{itemize}

\paragraph{Task}\label{task}

\{\{sachverhalt\}\}

\end{promptbox}

\subsection{Doctrinal Principles
(Grundprinzipien)}\label{doctrinal-principles-grundprinzipien-1}

\begin{promptbox}

Bearbeite die folgende juristische Frage nach deutschem Recht.

\paragraph{Ziel}\label{ziel}

Gib eine kurze, präzise und methodisch saubere juristische Antwort auf
die konkrete Fragestellung.

Die Antwort soll:

\begin{itemize}
\tightlist
\item
  die rechtlich relevante Kernfrage erkennen,
\item
  die einschlägigen Normen oder Prinzipien benennen,
\item
  den konkreten Sachverhalt knapp subsumieren,
\item
  und zu einer klaren Ja-/Nein-Entscheidung gelangen.
\end{itemize}

\paragraph{Stil}\label{stil}

Die Antwort soll deutlich kürzer und fokussierter sein als eine
vollständige Klausurlösung.

\begin{itemize}
\tightlist
\item
  Keine umfangreiche Gliederung.
\item
  Keine langen Streitstandsdebatten.
\item
  Keine künstliche Verlängerung.
\item
  Keine allgemeinen Lehrbuchausführungen.
\end{itemize}

Wenn mehrere vertretbare Ansichten bestehen:

\begin{itemize}
\tightlist
\item
  stelle die wesentlichen Unterschiede knapp dar,
\item
  entscheide dich nachvollziehbar,
\item
  und arbeite konsistent mit der gewählten Ansicht weiter.
\end{itemize}

\paragraph{Sachverhaltsbezug}\label{sachverhaltsbezug}

Arbeite streng sachverhaltsbezogen.

\begin{itemize}
\tightlist
\item
  Nutze nur Informationen aus Fall und Fragestellung.
\item
  Erfinde keine zusätzlichen Tatsachen.
\item
  Vermeide abstrakte Standardformulierungen ohne Fallbezug.
\end{itemize}

\paragraph{Ausgabeformat}\label{ausgabeformat-1}

Gib ausschließlich gültiges JSON im folgenden Format aus:

\begin{verbatim}
{
  "answer": "Ja" oder "Nein",
  "reasoning": "kurze juristische Begründung"
}
\end{verbatim}

\paragraph{Fall}\label{fall}

\{\{fall\}\}

\paragraph{Fragestellung}\label{fragestellung}

\{\{task\}\}

\end{promptbox}

\emph{English translation (reader reference; not sent to any model):}

\begin{promptbox}

Work through the following legal question under German law.

\paragraph{Goal}\label{goal}

Give a short, precise, and methodologically clean legal answer to the
concrete question.

The answer should:

\begin{itemize}
\tightlist
\item
  identify the legally relevant core question,
\item
  name the applicable norms or principles,
\item
  briefly subsume the concrete fact pattern,
\item
  and arrive at a clear Ja/Nein decision.
\end{itemize}

\paragraph{Style}\label{style}

The answer should be considerably shorter and more focused than a full
exam-style case solution.

\begin{itemize}
\tightlist
\item
  No extensive structuring.
\item
  No long discussions of scholarly debates.
\item
  No artificial padding.
\item
  No general textbook expositions.
\end{itemize}

When several defensible views exist:

\begin{itemize}
\tightlist
\item
  briefly present the essential differences,
\item
  decide for one view in a comprehensible way,
\item
  and work consistently with the chosen view.
\end{itemize}

\paragraph{Reference to the fact
pattern}\label{reference-to-the-fact-pattern}

Work strictly with reference to the fact pattern.

\begin{itemize}
\tightlist
\item
  Use only information from the case and the question.
\item
  Do not invent additional facts.
\item
  Avoid abstract standard formulations without a connection to the case.
\end{itemize}

\paragraph{Output format}\label{output-format-1}

Output only valid JSON in the following format:

\begin{verbatim}
{
  "answer": "Ja" or "Nein",
  "reasoning": "short legal justification"
}
\end{verbatim}

\paragraph{Case}\label{case}

\{\{fall\}\}

\paragraph{Question}\label{question}

\{\{task\}\}

\end{promptbox}

\section{Evaluation prompts for the LLM
judge}\label{evaluation-prompts-for-the-llm-judge}

\subsection{Exam-style cases
(Klausuren)}\label{exam-style-cases-klausuren-2}

\begin{promptbox}

Du bist ein strenger, aber fairer Korrektor (``LLM Judge'') für
juristische Klausuren deutscher Studierender bis zum 1. Staatsexamen.

Du bewertest ausschließlich auf Basis von: 1. Angabe/Sachverhalt, 2.
Studierendenlösung, 3. Musterlösung.

Nutze keine externen Quellen und erfinde keine Tatsachen.

Grundsätze:

\begin{itemize}
\tightlist
\item
  Musterlösung ist Referenz. Vergib aber vergleichbar Punkte für
  vertretbare Alternativlösungen, wenn sie mit der Angabe vereinbar und
  methodisch sauber begründet sind.
\item
  Folgefehlerprinzip: Einen Fehler nicht doppelt bestrafen. Wenn ein
  früher Fehler spätere Teile beeinflusst, bewerte die spätere
  Argumentation unter der Prämisse des studentischen
  Zwischenergebnisses, soweit methodisch sauber.
\item
  Keine Punkte für bloße Schlagworte ohne Definition/Subsumtion.
\item
  Beziehe dich in der Begründung konkret darauf, was in der
  Studierendenlösung steht und wie es zur Musterlösung passt.
\end{itemize}

Bewertungsschema (insgesamt 100 Punkte):

\begin{enumerate}
\def\labelenumi{\arabic{enumi}.}
\tightlist
\item
  Ergebnisrichtigkeit (0-20): \textbf{Vollpunkte (≈18-20):}
  Kernergebnisse stimmen mit Musterlösung überein oder sind vertretbar
  gleichwertig (z.B. anderes, aber methodisch korrekt begründetes
  Ergebnis). \textbf{Mittel (≈10-14):} Haupttendenz stimmt, aber ein
  wesentliches Ergebnis ist falsch oder Rechtsfolgen sind merklich
  verfehlt. \textbf{Niedrig (≈0-8):} Zentrale Ergebnisse
  verfehlt/inkonsistent; Lösung läuft am Fall vorbei.
\item
  Vollständigkeit \& Problemidentifikation (0-10): \textbf{Vollpunkte
  (≈9-10):} Alle wesentlichen Probleme aus Angabe/Musterlösung erkannt;
  keine großen „blinden Flecken``. \textbf{Mittel (≈5-7):} Ein
  wesentliches Problem fehlt oder ein Schwerpunkt wird übersehen; sonst
  ordentlich. \textbf{Niedrig (≈0-4):} Mehrere zentrale Probleme fehlen
  oder es werden viele irrelevante Nebenthemen geprüft („Themenklausur``
  statt Falllösung).
\item
  Rechtsgrundlagen \& Prüfungssystematik (0-10): \textbf{Vollpunkte
  (≈9-10):} Richtige Anspruchsgrundlagen/Prüfungsprogramme, logische
  Reihenfolge, saubere Prüfungsabschnitte. \textbf{Mittel (≈5-7):}
  Kleine Systematikfehler (Reihenfolge, einzelne falsche/unnötige
  Grundlagen), aber Gesamtstruktur trägt. \textbf{Niedrig (≈0-4):}
  Falsche Grundstruktur (z.B. falscher Rechtsbehelf/Anspruch, falscher
  Prüfungsrahmen), sodass die Prüfung methodisch entgleist.
\item
  Rechtskenntnis (Definitionen, Normen, Streitstände) (0-15):
  \textbf{Vollpunkte (≈13-15):} Definitionen korrekt, Normen passend,
  Streitstände nur dort, wo nötig; Streitentscheid begründet.
  \textbf{Mittel (≈8-11):} Einzelne Definitionen/Normen ungenau;
  Streitstände schematisch oder lückenhaft, aber nicht fallentscheidend
  falsch. \textbf{Niedrig (≈0-7):} Häufig falsche Definitionen/Normen
  oder erfundene Anforderungen; Streitstände wirr/fehlerhaft.
\item
  Subsumtion \& Argumentationsqualität (Fallbezug) (0-15):
  \textbf{Vollpunkte (≈13-15):} Konsequenter Tatsachenbezug, saubere
  Subsumtion, nachvollziehbare Argumente, Abwägungen strukturiert.
  \textbf{Mittel (≈8-11):} Teilweise nur abstrakt/„leerformelhaft``,
  aber wesentliche Punkte werden noch fallbezogen gelöst.
  \textbf{Niedrig (≈0-7):} Kaum Subsumtion, überwiegend Definitionen
  ohne Anwendung; Ergebnisse nicht begründet oder widersprüchlich.
\item
  Schwerpunktsetzung \& Problemtiefe (0-10): \textbf{Vollpunkte
  (≈9-10):} Langer Sachverhaltsblock/Indizien → angemessen vertieft;
  einfache Punkte kurz; Schwerpunkt stimmt mit Musterlösung und
  „Signalen`` der Angabe überein. \textbf{Mittel (≈5-7):} Entweder
  Überlänge bei Nebensachen oder Untergewichtung eines Schwerpunkts,
  aber noch erkennbares Klausurbewusstsein. \textbf{Niedrig (≈0-4):}
  Massive Fehlgewichtung: Hauptproblem kaum behandelt, Nebensachen
  dominieren.
\item
  Methodischer Stil: Obersatz → Definition → Subsumtion → Ergebnis
  (0-10): \textbf{Vollpunkte (≈9-10):} Bei allen wichtigen
  Prüfungspunkten klar erkennbar; Zwischenergebnisse werden gezogen;
  auch im Urteilsstil sinngemäß eingehalten. \textbf{Mittel (≈5-7):}
  Grundschema vorhanden, aber häufiger vermischt/ausgelassen;
  Definitionen oder Ergebnisse fehlen gelegentlich. \textbf{Niedrig
  (≈0-4):} Kein prüfungsorientierter Stil; nur Nacherzählung,
  Stichworte, oder reine Ergebnisbehauptungen.
\item
  Gliederung, Gliederungsebenen \& Leseführung (0-5): \textbf{5:}
  Stringente Gliederung, Ebenen sauber, Prüfpunkte auffindbar.
  \textbf{3-4:} Kleinere Inkonsistenzen (Ebenensprünge, Überschriften
  nicht passgenau). \textbf{0-2:} Unstrukturierter Textblock; Gliederung
  irreführend.
\item
  Sprache \& juristische Terminologie (0-3): \textbf{3:} Präzise,
  juristisch sauber, gut verständlich. \textbf{2:} Teilweise
  umständlich/ungenau, aber verständlich. \textbf{0-1:} Häufig
  missverständlich, falsche Begriffe, viele sprachliche Brüche.
\item
  Formalia: Normzitierweise, Sachverhaltsarbeit, Sorgfalt (0-2):
  \textbf{2:} Normen überwiegend korrekt zitiert; Sachverhalt korrekt
  verarbeitet; keine groben Schnitzer. \textbf{1:} Einzelne
  Zitier-/Sorgfaltsfehler, nicht gravierend. \textbf{0:} Häufige grobe
  Fehler (Rollen vertauscht, zentrale Fakten falsch wiedergegeben,
  Normchaos).
\end{enumerate}

Umrechnung Rohpunkte → Notenpunkte (0-18; \textless4 = nicht bestanden):
0-12→0, 13-25→1, 26-38→2, 39-49→3, 50-53→4, 54-56→5, 57-59→6, 60-63→7,
64-66→8, 67-69→9, 70-73→10, 74-76→11, 77-79→12, 80-83→13, 84-86→14,
87-89→15, 90-93→16, 94-96→17, 97-100→18

Ausgabeformat:

\begin{itemize}
\tightlist
\item
  Gib eine JSON-Ausgabe ohne zusätzlichen Text aus.
\item
  Für jede Dimension: score (0 bis max, ggf. 0.5 Schritte), max, und
  eine Begründung als zusammenhängender Absatz (3-7 Sätze).
\item
  Danach: total\_score (0-100), grade\_points (0-18), passed
  (true/false), kurze Gesamtwürdigung (2-5 Sätze) und 3 konkrete
  Verbesserungshinweise.
\end{itemize}

ANGABE / SACHVERHALT: \{\{\{ANGABE\_TEXT\}\}\}

MUSTERLÖSUNG: \{\{\{MUSTERLOESUNG\_TEXT\}\}\}

STUDIERENDENLÖSUNG: \{\{\{LOESUNG\_TEXT\}\}\}

Aufgabe: Bewerte die Lösung nach dem vorgegebenen 100-Punkte-Schema.
Vergib zu jeder Dimension Punkte und schreibe eine Begründung als
Absatz. Berechne die Summe (0-100) und wandle in Notenpunkte (0-18) um.
Markiere passed=true nur, wenn grade\_points \textgreater= 4.

Gib ausschließlich JSON aus.

\end{promptbox}

\emph{English translation (reader reference; not sent to any model):}

\begin{promptbox}

You are a strict but fair grader (``LLM Judge'') for legal exam-style
cases written by German students up to the first state examination
(\emph{Erstes Staatsexamen}).

You evaluate exclusively on the basis of: 1. the task/fact pattern, 2.
the student solution (\emph{Studierendenlösung}), 3. the reference
solution (\emph{Musterlösung}).

Do not use any external sources and do not invent any facts.

Principles:

\begin{itemize}
\tightlist
\item
  The reference solution is the benchmark. However, award comparable
  points for defensible alternative solutions, provided they are
  compatible with the task and methodologically soundly reasoned.
\item
  Follow-on-error principle (\emph{Folgefehlerprinzip}): do not punish
  one error twice. If an early error affects later parts, evaluate the
  later argumentation on the assumption of the student's intermediate
  result, to the extent this is methodologically clean.
\item
  No points for mere buzzwords without definition / subsumption.
\item
  In your justification, refer concretely to what the student solution
  says and how it fits the reference solution.
\end{itemize}

Grading scheme (100 points in total):

\begin{enumerate}
\def\labelenumi{\arabic{enumi}.}
\tightlist
\item
  Correctness of result (0-20): \textbf{Full points (≈18-20):} core
  results match the reference solution or are defensibly equivalent
  (e.g.~a different but methodologically correctly reasoned result).
  \textbf{Middle (≈10-14):} the main tendency is right, but one
  essential result is wrong or the legal consequences are noticeably
  off. \textbf{Low (≈0-8):} central results are missed / inconsistent;
  the solution misses the point of the case.
\item
  Completeness and problem identification (0-10): \textbf{Full points
  (≈9-10):} all essential problems from the task / reference solution
  are recognised; no major ``blind spots''. \textbf{Middle (≈5-7):} one
  essential problem is missing or a focal point is overlooked; otherwise
  sound. \textbf{Low (≈0-4):} several central problems are missing, or
  many irrelevant side topics are examined (``topic essay'' instead of
  case solution).
\item
  Legal bases and examination systematics (0-10): \textbf{Full points
  (≈9-10):} correct causes of action / examination programmes, logical
  sequence, clean examination sections. \textbf{Middle (≈5-7):} small
  systematic errors (sequence, individual incorrect / unnecessary
  bases), but the overall structure holds. \textbf{Low (≈0-4):} wrong
  basic structure (e.g.~wrong legal remedy / claim, wrong examination
  framework), so that the analysis methodologically derails.
\item
  Legal knowledge (definitions, norms, scholarly debates) (0-15):
  \textbf{Full points (≈13-15):} definitions are correct, norms are apt,
  scholarly debates appear only where needed; the decision in the debate
  is reasoned. \textbf{Middle (≈8-11):} individual definitions / norms
  are imprecise; scholarly debates are schematic or patchy, but not
  wrong in a case-decisive way. \textbf{Low (≈0-7):} frequently wrong
  definitions / norms or invented requirements; scholarly debates are
  muddled / faulty.
\item
  Subsumption and quality of argumentation (connection to the case)
  (0-15): \textbf{Full points (≈13-15):} consistent reference to facts,
  clean subsumption, comprehensible arguments, structured balancing.
  \textbf{Middle (≈8-11):} in part only abstract /
  ``empty-formula-like'', but essential points are still solved with
  reference to the case. \textbf{Low (≈0-7):} hardly any subsumption,
  predominantly definitions without application; results are unreasoned
  or contradictory.
\item
  Setting priorities and depth of analysis (0-10): \textbf{Full points
  (≈9-10):} long fact-pattern blocks / indications are appropriately
  treated in depth; easy points are kept short; the focus matches the
  reference solution and the ``signals'' of the task. \textbf{Middle
  (≈5-7):} either overlength on minor matters or under-weighting of a
  focal point, but still recognizable exam awareness. \textbf{Low
  (≈0-4):} massive misallocation: the main problem is barely treated,
  secondary matters dominate.
\item
  Methodological style: major premise → definition → subsumption →
  result (0-10): \textbf{Full points (≈9-10):} clearly recognizable at
  all important examination points; intermediate results are drawn; also
  observed in spirit when written in judgement style. \textbf{Middle
  (≈5-7):} the basic scheme is present but is frequently blended or
  omitted; definitions or results are occasionally missing. \textbf{Low
  (≈0-4):} no examination-oriented style; only narration, bullet points,
  or bare assertions of results.
\item
  Structure, structural levels, and reader guidance (0-5): \textbf{5:}
  stringent structure, clean levels, examination points are findable.
  \textbf{3-4:} small inconsistencies (level jumps, headings not
  precisely fitting). \textbf{0-2:} unstructured block of text;
  structure is misleading.
\item
  Language and legal terminology (0-3): \textbf{3:} precise, legally
  clean, easily understandable. \textbf{2:} partly clumsy / imprecise,
  but understandable. \textbf{0-1:} frequently misleading, wrong terms,
  many linguistic breaks.
\item
  Formalia: norm citation style, work with the fact pattern, care (0-2):
  \textbf{2:} norms cited predominantly correctly; fact pattern
  processed correctly; no gross blunders. \textbf{1:} isolated citation
  / care errors, not serious. \textbf{0:} frequent gross errors (roles
  swapped, central facts misreported, norm chaos).
\end{enumerate}

Conversion of raw points → grade points (0-18; \textless4 = fail):
0-12→0, 13-25→1, 26-38→2, 39-49→3, 50-53→4, 54-56→5, 57-59→6, 60-63→7,
64-66→8, 67-69→9, 70-73→10, 74-76→11, 77-79→12, 80-83→13, 84-86→14,
87-89→15, 90-93→16, 94-96→17, 97-100→18

Output format:

\begin{itemize}
\tightlist
\item
  Output a JSON response without any additional text.
\item
  For each dimension: score (0 to max, in 0.5 steps if needed), max, and
  a justification as a coherent paragraph (3-7 sentences).
\item
  Then: total\_score (0-100), grade\_points (0-18), passed (true/false),
  a short overall assessment (2-5 sentences), and 3 concrete suggestions
  for improvement.
\end{itemize}

TASK / FACT PATTERN: \{\{\{ANGABE\_TEXT\}\}\}

REFERENCE SOLUTION: \{\{\{MUSTERLOESUNG\_TEXT\}\}\}

STUDENT SOLUTION: \{\{\{LOESUNG\_TEXT\}\}\}

Task: Evaluate the solution according to the prescribed 100-point
scheme. Award points for each dimension and write a justification as a
paragraph. Calculate the sum (0-100) and convert it into grade points
(0-18). Set passed=true only if grade\_points \textgreater= 4.

Output only JSON.

\end{promptbox}

\subsection{Doctrinal Principles
(Grundprinzipien)}\label{doctrinal-principles-grundprinzipien-2}

\begin{promptbox}

Du bist ein strenger, aber fairer juristischer Korrektor für kurze
juristische Begründungsfragen nach deutschem Recht.

Du bewertest ausschließlich auf Basis von:

\begin{enumerate}
\def\labelenumi{\arabic{enumi}.}
\tightlist
\item
  Fall
\item
  Fragestellung
\item
  Referenzlösung
\item
  Zu bewertende Antwort
\end{enumerate}

Nutze keine externen Quellen und erfinde keine Tatsachen.

\paragraph{Bewertungsgrundsätze}\label{bewertungsgrundsuxe4tze}

\begin{itemize}
\tightlist
\item
  Die Referenzlösung ist Orientierung, nicht die einzig mögliche
  vertretbare Lösung.
\item
  Vergib volle oder nahezu volle Punkte auch für methodisch vertretbare
  Alternativbegründungen.
\item
  Entscheidend ist die juristische Richtigkeit und Qualität der
  Begründung, nicht die sprachliche Ähnlichkeit zur Referenz.
\item
  Keine Punkte für bloße Schlagworte ohne nachvollziehbare Begründung.
\item
  Eine falsche Ja/Nein-Entscheidung begrenzt die Maximalpunktzahl
  erheblich, selbst wenn Teile der Begründung juristisch plausibel sind.
\end{itemize}

\paragraph{Bewertungsschema (100
Punkte)}\label{bewertungsschema-100-punkte}

\begin{enumerate}
\def\labelenumi{\arabic{enumi}.}
\tightlist
\item
  Ergebnisrichtigkeit (0-40)
\end{enumerate}

\begin{itemize}
\tightlist
\item
  Ist die Ja/Nein-Entscheidung juristisch zutreffend oder vertretbar?
\end{itemize}

\begin{enumerate}
\def\labelenumi{\arabic{enumi}.}
\tightlist
\item
  Rechtskenntnis und Normbezug (0-25)
\end{enumerate}

\begin{itemize}
\tightlist
\item
  Werden einschlägige Normen, Definitionen oder Prinzipien korrekt
  erkannt und verwendet?
\end{itemize}

\begin{enumerate}
\def\labelenumi{\arabic{enumi}.}
\tightlist
\item
  Subsumtion und Fallbezug (0-25)
\end{enumerate}

\begin{itemize}
\tightlist
\item
  Wird der konkrete Sachverhalt nachvollziehbar unter die rechtlichen
  Voraussetzungen subsumiert?
\end{itemize}

\begin{enumerate}
\def\labelenumi{\arabic{enumi}.}
\tightlist
\item
  Klarheit und Präzision (0-10)
\end{enumerate}

\begin{itemize}
\tightlist
\item
  Ist die Antwort präzise, verständlich und methodisch sauber
  formuliert?
\end{itemize}

\paragraph{Ausgabeformat}\label{ausgabeformat-2}

Gib ausschließlich gültiges JSON aus:

\begin{verbatim}
{
  "scores": {
    "result_correctness": {
      "score": ...,
      "max": 40,
      "reason": "..."
    },
  "legal_knowledge": {
    "score": ...,
    "max": 25,
    "reason": "..."
  },
"subsumption": {
  "score": ...,
  "max": 25,
  "reason": "..."
},
"clarity": {
  "score": ...,
  "max": 10,
  "reason": "..."
}
},
"total_score": ...,
"overall_assessment": "..."
}
\end{verbatim}

\paragraph{Fall}\label{fall-1}

\{\{fall\}\}

\paragraph{Fragestellung}\label{fragestellung-1}

\{\{task\}\}

\paragraph{Referenzlösung}\label{referenzluxf6sung}

Antwort: \{\{binary\_solution\}\}

Begründung: \{\{reasoning\}\}

\paragraph{Zu bewertende Antwort}\label{zu-bewertende-antwort}

\{\{answer\}\}

\end{promptbox}

\emph{English translation (reader reference; not sent to any model):}

\begin{promptbox}

You are a strict but fair legal grader for short legal reasoning
questions under German law.

You evaluate exclusively on the basis of:

\begin{enumerate}
\def\labelenumi{\arabic{enumi}.}
\tightlist
\item
  the case,
\item
  the question,
\item
  the reference solution,
\item
  the answer to be evaluated.
\end{enumerate}

Do not use any external sources and do not invent any facts.

\paragraph{Grading principles}\label{grading-principles}

\begin{itemize}
\tightlist
\item
  The reference solution is a guide, not the only defensible solution.
\item
  Award full or near-full points also for methodologically defensible
  alternative justifications.
\item
  What matters is the legal correctness and quality of the reasoning,
  not linguistic similarity to the reference.
\item
  No points for mere buzzwords without a comprehensible justification.
\item
  A wrong Ja/Nein decision substantially limits the maximum score, even
  if parts of the reasoning are legally plausible.
\end{itemize}

\paragraph{Grading scheme (100 points)}\label{grading-scheme-100-points}

\begin{enumerate}
\def\labelenumi{\arabic{enumi}.}
\tightlist
\item
  Correctness of result (0-40)
\end{enumerate}

\begin{itemize}
\tightlist
\item
  Is the Ja/Nein decision legally correct or defensible?
\end{itemize}

\begin{enumerate}
\def\labelenumi{\arabic{enumi}.}
\tightlist
\item
  Legal knowledge and reference to norms (0-25)
\end{enumerate}

\begin{itemize}
\tightlist
\item
  Are the applicable norms, definitions, or principles correctly
  identified and used?
\end{itemize}

\begin{enumerate}
\def\labelenumi{\arabic{enumi}.}
\tightlist
\item
  Subsumption and reference to the case (0-25)
\end{enumerate}

\begin{itemize}
\tightlist
\item
  Is the concrete fact pattern subsumed under the legal requirements in
  a comprehensible way?
\end{itemize}

\begin{enumerate}
\def\labelenumi{\arabic{enumi}.}
\tightlist
\item
  Clarity and precision (0-10)
\end{enumerate}

\begin{itemize}
\tightlist
\item
  Is the answer precisely, understandably, and methodologically cleanly
  formulated?
\end{itemize}

\paragraph{Output format}\label{output-format-2}

Output only valid JSON:

\begin{verbatim}
{
  "scores": {
    "result_correctness": {
      "score": ...,
      "max": 40,
      "reason": "..."
    },
  "legal_knowledge": {
    "score": ...,
    "max": 25,
    "reason": "..."
  },
"subsumption": {
  "score": ...,
  "max": 25,
  "reason": "..."
},
"clarity": {
  "score": ...,
  "max": 10,
  "reason": "..."
}
},
"total_score": ...,
"overall_assessment": "..."
}
\end{verbatim}

\paragraph{Case}\label{case-1}

\{\{fall\}\}

\paragraph{Question}\label{question-1}

\{\{task\}\}

\paragraph{Reference solution}\label{reference-solution}

Answer: \{\{binary\_solution\}\}

Reasoning: \{\{reasoning\}\}

\paragraph{Answer to be evaluated}\label{answer-to-be-evaluated}

\{\{answer\}\}

\end{promptbox}

\section{LLM-judge configuration}\label{llm-judge-configuration}

The primary judge GPT-5.4-mini is the one used for every reported
leaderboard score and for the RQ5 within-judge stochasticity analysis
(k=3 repeats per generation). The five secondary judges Opus-4.7,
Gemini-3.1-Pro, DeepSeek-V4-Pro, Qwen3.5-397B-A17B, and Sonnet-4.6
evaluate every Benchathon cell once for the RQ5 between-judge bias
analysis. The full model ids are shown in the \emph{Model} column below;
see Table~\ref{tbl-system-overview}
(Appendix~\ref{evaluated-systems-catalogue}) for the canonical
alias-to-id mapping.

\begin{table*}[t]
\centering\footnotesize
\caption{LLM-judge configuration. Temperature and max output tokens are the values captured in the per-call metadata of the Benchathon evaluation runs (primary 2026-05-31, cross-validation 2026-05-21 and -31); the retry budget is the worker-side default of 3, with the observed retry count across each run reported in parentheses. No provider-side snapshot id is returned for these aliases; the model id plus run date fixes the configuration for reproducibility.}
\label{tbl-judge-config}
\setlength{\tabcolsep}{4pt}
\begin{tabular*}{\textwidth}{@{\extracolsep{\fill}}lccccc@{}}
\toprule
Role & Model & Run date & $T$ & Max tok. & Retry budget \\
\midrule
Primary judge & \texttt{gpt-5.4-mini} & 2026-05-31 & $1.0$ & $8000$ & $3$ (obs.\ $0$) \\
Cross-validation (Anthropic) & \texttt{claude-opus-4-7} & 2026-05-21 & $1.0$ & $8000$ & $3$ (obs.\ $0$) \\
Cross-validation (Anthropic) & \texttt{claude-sonnet-4-6} & 2026-05-31 & $1.0$ & $8000$ & $3$ (obs.\ $0$) \\
Cross-validation (Google) & \texttt{gemini-3.1-pro-preview} & 2026-05-21 & $1.0$ & $8000$ & $3$ (obs.\ $0$) \\
Cross-validation (DeepSeek) & \texttt{deepseek-ai/DeepSeek-V4-Pro} & 2026-05-31 & $1.0$ & $8000$ & $3$ (obs.\ $0$) \\
Cross-validation (Alibaba) & \texttt{Qwen/Qwen3.5-397B-A17B} & 2026-05-31 & $1.0$ & $8000$ & $3$ (obs.\ $0$) \\
\bottomrule
\end{tabular*}
\end{table*}

The full judge prompts are reproduced in Appendix
\ref{evaluation-prompts-for-the-llm-judge}: the exam-case rubric and the
Doctrinal Principles rubric. All six judges receive the same prompt and
the same input ordering (task, reference solution, solution to be
evaluated); only the underlying model changes.

\section{Judge calibration - supplementary
detail}\label{sec-judge-calibration-supplementary-detail}

This appendix collects the supplementary numerical detail behind the RQ5
per-judge calibration analysis: the judge-swapped system leaderboard and
its rank agreement in Tables~\ref{tbl-judge-swap-agreement} and
\ref{tbl-judge-swap}, the six-judge calibration table in
Table~\ref{tbl-judge-calibration}, the within-judge per-pass breakdown
of the primary GPT-5.4-mini judge in
Table~\ref{tbl-judge-calibration-detail}, the per-annotator alt-test
p-values, a robustness check on the intersection of solutions scored by
every judge, and the bootstrap uncertainty on the alt-test winning rate
\(\omega\).

\paragraph{Judge-swapped leaderboard.} On the cross-judge Benchathon
subset (every leaderboard system covered; per-system n in
Table~\ref{tbl-judge-swap}), each of the five cross-validation judges
re-scores the same generations as the primary judge. Per-system n is 15,
except Llama-4 (n=14), where one generation lacks a parseable score from
one judge. Table~\ref{tbl-judge-swap-agreement} reports the system-level
rank agreement of each judge's per-system means against the primary
judge's on the identical subset, together with each judge's mean pass
rate at the fixed 50-point threshold, raw and after removing that
judge's measured offset against the blind human pool
(Table~\ref{tbl-judge-calibration}). Strictness shifts pass rates at the
fixed threshold; the offset correction largely restores them, and the
ordering itself is stable under every judge.

\begin{table*}[t]
\centering\footnotesize
\caption{System-level rank agreement of each cross-validation judge against the primary judge on the shared Benchathon subset (Spearman $\rho$ / Kendall $\tau$ over 12 per-system means), and each judge's mean pass rate at the fixed $\geq 50/100$ threshold, raw and after removing the judge's measured offset vs.\ the blind human pool.}
\label{tbl-judge-swap-agreement}
\setlength{\tabcolsep}{4pt}
\begin{tabular*}{\textwidth}{@{\extracolsep{\fill}}lcccc@{}}
\toprule
Judge & $\rho$ & $\tau$ & Pass (raw) & Pass (offset-corr.) \\
\midrule
GPT-5.4-mini (primary) & — & — & 62\% & 62\% \\
Opus-4.7 & 0.97 & 0.85 & 58\% & 55\% \\
Sonnet-4.6 & 0.95 & 0.85 & 71\% & 58\% \\
Gemini-3.1-Pro & 0.92 & 0.79 & 74\% & 46\% \\
DeepSeek-V4-Pro & 0.94 & 0.84 & 59\% & 53\% \\
Qwen3.5-397B & 0.92 & 0.79 & 88\% & 58\% \\
\bottomrule
\end{tabular*}
\end{table*}

\begin{table*}[t]
\centering\footnotesize
\caption{Judge-swapped Benchathon leaderboard: per-system mean raw score under each judge on the identical generation subset, systems ordered by the primary judge ($\dagger$). Strictness moves the level; the ordering is preserved (rank agreement in Table~\ref{tbl-judge-swap-agreement}).}
\label{tbl-judge-swap}
\setlength{\tabcolsep}{3pt}
\begin{tabular*}{\textwidth}{@{\extracolsep{\fill}}lccccccc@{}}
\toprule
System & n & GPT-5.4-mini$^\dagger$ & Opus-4.7 & Sonnet-4.6 & Gemini-3.1-Pro & DeepSeek-V4-Pro & Qwen3.5-397B \\
\midrule
Opus-4.7 & 15 & 69.3 & 71.7 & 74.7 & 82.8 & 76.5 & 81.5 \\
Gemini-3.1-Pro & 15 & 68.3 & 68.4 & 69.6 & 81.7 & 74.6 & 76.3 \\
GPT-5.4 & 15 & 66.6 & 71.5 & 72.1 & 83.7 & 73.3 & 82.2 \\
Gemini-3.1-Flash-Lite & 15 & 63.5 & 63.1 & 67.1 & 82.0 & 68.1 & 77.3 \\
Sonnet-4.6 & 15 & 58.9 & 66.0 & 70.3 & 75.9 & 74.6 & 77.4 \\
GPT-5.4-mini & 15 & 58.6 & 51.1 & 57.3 & 60.3 & 50.0 & 65.5 \\
DeepSeek-V4-Pro & 15 & 56.1 & 56.9 & 62.5 & 70.1 & 60.5 & 71.4 \\
Qwen3.5-122B & 15 & 49.4 & 42.8 & 48.5 & 54.6 & 49.6 & 59.6 \\
DeepSeek-V4-Flash & 15 & 49.3 & 47.8 & 55.3 & 62.2 & 52.9 & 65.1 \\
Qwen3.6-35B & 15 & 42.4 & 31.5 & 44.2 & 40.2 & 40.7 & 54.2 \\
Qwen3-235B & 15 & 41.9 & 34.4 & 43.9 & 46.1 & 39.6 & 54.1 \\
Llama-4 & 14 & 37.7 & 29.0 & 37.0 & 37.8 & 31.6 & 51.8 \\
\bottomrule
\end{tabular*}
\end{table*}

\paragraph{Per-judge calibration.} Table~\ref{tbl-judge-calibration}
reports the per-judge calibration statistics. It gives two quantities,
both averaged over the validation solutions. The direct judge-vs-pool
offset \(\bar{\Delta}_{\text{dir}}=\overline{\text{judge}-\text{full}}\)
is the mean difference between the judge's score and the full blind-pool
mean; it is the number to quote when prose says ``the judge sits \(X\)
raw points above the pool'\,'. The Calderon pool-substitution shift
\(\bar{\Delta}_{\text{sub}}=\overline{\text{sub}-\text{full}}\) compares
the pool mean after substituting the judge for one blind reviewer
against the full-pool mean; replacing one of \(k=3\) raters scales the
direct offset down by \(1/3\), so
\(\bar{\Delta}_{\text{sub}}\approx\bar{\Delta}_{\text{dir}}/3\). Both
come with a paired \(t\)-test. Shapiro-Wilk does not reject normality in
all but two cells; in those two the near-zero offset is unchanged under
a Wilcoxon signed-rank cross-check, so the parametric test applies
throughout.

\begin{table*}[t]
\centering\footnotesize
\caption{Per-judge calibration on the two blind-graded subsets of the Benchathon validation set, ordered by increasing leniency on human-written solutions. Direct offset $\bar{\Delta}_{\text{dir}}$ is the per-solution paired difference between the judge and the mean of three blind reviewers. $\bar{\Delta}_{\text{sub}}$ is the same quantity after the Calderon \S1 pool substitution and is attenuated by $1/3$ at $k=3$ raters. SW $p$ tests normality of the paired sub-full differences. Six judges: GPT-5.4-mini is the primary judge that anchors RQ1-RQ2; Opus-4.7, Sonnet-4.6, DeepSeek-V4-Pro, Gemini-3.1-Pro and Qwen3.5-397B-A17B are Benchathon-only cross-validation judges.}
\label{tbl-judge-calibration}
\setlength{\tabcolsep}{4pt}
\resizebox{\textwidth}{!}{%
\begin{tabular}{lcccccc}
\toprule
Configuration & $n$ & $\bar{\Delta}_{\text{dir}}$ (sd) & $t$ ($p$) & $\bar{\Delta}_{\text{sub}}$ (sd) & SW $p$ & $t_{\text{sub}}$ ($p$) \\
\midrule
\textit{Human-written:}~Baseline GPT-5.4-mini (primary) & $n=30$ & $-0.01$ (sd $13.43$) & $-0.00$ ($0.9964$) & $0.20$ (sd $4.99$) & $0.02$ & $0.22$ ($0.8279$) \\
\quad Opus-4.7 & $n=28$ & $-2.06$ (sd $13.93$) & $-0.78$ ($0.4409$) & $-0.52$ (sd $4.71$) & $0.78$ & $-0.59$ ($0.5611$) \\
\quad Sonnet-4.6 & $n=30$ & $0.62$ (sd $12.96$) & $0.26$ ($0.7945$) & $0.41$ (sd $4.46$) & $0.94$ & $0.50$ ($0.6178$) \\
\quad DeepSeek-V4-Pro & $n=30$ & $2.11$ (sd $14.89$) & $0.77$ ($0.4448$) & $0.91$ (sd $4.69$) & $0.99$ & $1.06$ ($0.2987$) \\
\quad Gemini-3.1-Pro & $n=30$ & $6.82$ (sd $15.25$) & $2.45$ ($0.0206$) & $2.48$ (sd $4.79$) & $0.96$ & $2.83$ ($0.0084$) \\
\quad Qwen3.5-397B-A17B & $n=30$ & $8.72$ (sd $15.00$) & $3.19$ ($0.0034$) & $3.11$ (sd $4.91$) & $0.31$ & $3.47$ ($0.0017$) \\
\textit{LLM-generated:}~Baseline GPT-5.4-mini (primary) & $n=15$ & $-0.47$ (sd $16.24$) & $-0.11$ ($0.9130$) & $0.83$ (sd $7.65$) & $0.33$ & $0.42$ ($0.6795$) \\
\quad Opus-4.7 & $n=15$ & $1.37$ (sd $10.95$) & $0.48$ ($0.6362$) & $1.44$ (sd $5.62$) & $0.52$ & $0.99$ ($0.3368$) \\
\quad Sonnet-4.6 & $n=15$ & $6.87$ (sd $13.25$) & $2.01$ ($0.0645$) & $3.28$ (sd $6.62$) & $0.85$ & $1.92$ ($0.0757$) \\
\quad DeepSeek-V4-Pro & $n=15$ & $5.83$ (sd $16.98$) & $1.33$ ($0.2045$) & $2.93$ (sd $6.63$) & $0.48$ & $1.71$ ($0.1088$) \\
\quad Gemini-3.1-Pro & $n=15$ & $18.17$ (sd $8.40$) & $8.37$ ($0.0000$) & $7.04$ (sd $5.03$) & $0.89$ & $5.42$ ($0.0001$) \\
\quad Qwen3.5-397B-A17B & $n=15$ & $16.67$ (sd $13.47$) & $4.79$ ($0.0003$) & $6.54$ (sd $6.03$) & $0.32$ & $4.20$ ($0.0009$) \\
\bottomrule
\end{tabular}
}
\end{table*}

\paragraph{Per-pass detail.} Table~\ref{tbl-judge-calibration-detail}
reports the same per-judge calibration statistics as
Table~\ref{tbl-judge-calibration}, but with the three Config-A
GPT-5.4-mini intra-judge passes broken out individually alongside the
baseline single-pass. This gives a \(k=4\) within-judge stochasticity
check on the primary judge. The four per-pass offsets lie close together
(stdev \(1.5\) raw points on human-written solutions, \(1.7\) on
LLM-generated ones). That is small compared to the differences between
judges (the lenient judges sit +6 to +18 raw points above the pool on
LLM-generated solutions) and to the \$\approx\$24-raw-point
within-solution max-min range across the blind reviewers. The same
re-runs give the within-judge stochasticity figure referenced in
§\ref{llm-as-a-judge-evaluation}: across the \(k=3\) Config-A re-runs on
the Benchathon LLM generations, the within-cell standard deviation is
3.45 raw points, while the blind reviewers' within-solution grades span
a 23.9-raw-point max-min range. The two are different dispersion
measures, but as expected, a single LLM judge is far more
self-consistent across re-runs than the human pool is internally.

\begin{table*}[t]
\centering\footnotesize
\caption{Within-judge stochasticity detail for the primary GPT-5.4-mini judge. The Baseline row is the single-pass run that anchors RQ1-RQ2. The Config A block adds three intra-judge re-runs ($3\times$) so the four offsets in each block give a $k=4$ within-judge stochasticity check. The per-pass offset $\bar{\Delta}_{\text{dir}}$ stdev across the four passes is $1.53$ raw points on human-written solutions and $1.65$ on LLM-generated ones, with max $|$pass\,$-$\,mean$|$ of $2.23$ (human) and $2.11$ (LLM). Same columns as Table~\ref{tbl-judge-calibration}.}
\label{tbl-judge-calibration-detail}
\setlength{\tabcolsep}{4pt}
\resizebox{\textwidth}{!}{%
\begin{tabular}{lcccccc}
\toprule
Configuration & $n$ & $\bar{\Delta}_{\text{dir}}$ (sd) & $t$ ($p$) & $\bar{\Delta}_{\text{sub}}$ (sd) & SW $p$ & $t_{\text{sub}}$ ($p$) \\
\midrule
\textit{Human-written:}~Baseline GPT-5.4-mini ($1\times$, primary) & $n=30$ & $-0.01$ (sd $13.43$) & $-0.00$ ($0.9964$) & $0.20$ (sd $4.99$) & $0.02$ & $0.22$ ($0.8279$) \\
\quad Config A GPT-5.4-mini ($3\times$ mean) & $n=30$ & $-1.68$ (sd $14.43$) & $-0.64$ ($0.5278$) & $-0.36$ (sd $5.41$) & $0.10$ & $-0.36$ ($0.7202$) \\
\qquad pass 1 & $n=30$ & $-0.88$ (sd $18.22$) & $-0.26$ ($0.7937$) & $-0.09$ (sd $6.62$) & $0.02$ & $-0.07$ ($0.9419$) \\
\qquad pass 2 & $n=30$ & $-0.68$ (sd $13.82$) & $-0.27$ ($0.7901$) & $-0.02$ (sd $5.31$) & $0.10$ & $-0.02$ ($0.9819$) \\
\qquad pass 3 & $n=30$ & $-3.49$ (sd $13.57$) & $-1.41$ ($0.1692$) & $-0.96$ (sd $5.00$) & $0.57$ & $-1.05$ ($0.3010$) \\
\textit{LLM-generated:}~Baseline GPT-5.4-mini ($1\times$, primary) & $n=15$ & $-0.47$ (sd $16.24$) & $-0.11$ ($0.9130$) & $0.83$ (sd $7.65$) & $0.33$ & $0.42$ ($0.6795$) \\
\quad Config A GPT-5.4-mini ($3\times$ mean) & $n=15$ & $2.34$ (sd $14.64$) & $0.62$ ($0.5450$) & $1.77$ (sd $7.05$) & $0.38$ & $0.97$ ($0.3475$) \\
\qquad pass 1 & $n=15$ & $2.97$ (sd $17.50$) & $0.66$ ($0.5222$) & $1.98$ (sd $7.85$) & $0.45$ & $0.98$ ($0.3455$) \\
\qquad pass 2 & $n=15$ & $1.13$ (sd $14.89$) & $0.29$ ($0.7724$) & $1.37$ (sd $6.77$) & $0.12$ & $0.78$ ($0.4472$) \\
\qquad pass 3 & $n=15$ & $2.93$ (sd $14.04$) & $0.81$ ($0.4319$) & $1.97$ (sd $7.15$) & $0.89$ & $1.07$ ($0.3046$) \\
\bottomrule
\end{tabular}
}
\end{table*}

\paragraph{Per-annotator alt-test detail.}
Tables~\ref{tbl-calderon-per-annotator-human} and
\ref{tbl-calderon-per-annotator-llm} report the full Calderon blind-pool
alt-test by judge for the human-written and LLM-generated solutions
respectively. Each block's header states the winning rate \(\omega\)
that block accumulates. The rows below show the five blind reviewers'
per-annotator \(n_j\), \(\rho^f_j\), \(\rho^h_j\), \(\bar{d}_j\),
one-sided \(p\)-value (Wilcoxon signed-rank throughout, since every
\(n_j<30\)), and whether the Benjamini-Yekutieli FDR procedure at
\(q=0.05\) across \(m=5\) rejected the null. Only annotators flagged
\(\star\) contribute to \(\omega\).

\begin{table*}[p]
\centering\footnotesize
\caption{Human-written solutions: Calderon \S3 alt-test detail by judge. Each block lists the five blind reviewers in the roster (graders 01, 02, 04, 05, 07) for one judge, with that block's headline $\omega$ stated in its italic header. $\rho^f_j$ = mean over $i \in I_j$ of $\mathbf{1}\{S(f,x_i,j) \geq S(h_j,x_i,j)\}$ (probability the LLM judge aligns with the remaining $H_i\setminus\{j\}$ at least as closely as $h_j$); $\rho^h_j$ symmetrically. $\bar{d}_j = \rho^h_j - \rho^f_j$ is the per-annotator advantage probability gap. $p$-value is one-sided Wilcoxon signed-rank against $\bar{d}_j \geq \varepsilon = 0.15$. $\star$ marks rejection under Benjamini-Yekutieli FDR at $q=0.05$ across $m=5$.}
\label{tbl-calderon-per-annotator-human}
\setlength{\tabcolsep}{4pt}
\begin{tabular*}{\textwidth}{@{\extracolsep{\fill}}lccccc@{}}
\toprule
 & $n_j$ & $\rho^f / \rho^h$ & $\bar{d}$ & $p$ & rej. \\
\midrule
\multicolumn{6}{l}{\textit{Baseline GPT-5.4-mini (primary) ($\omega=0.60$ passes)}} \\
\quad grader\_01 & $14$ & $0.57$ / $0.43$ & $-0.143$ & $0.0247$ &  \\
\quad grader\_02 & $19$ & $0.68$ / $0.32$ & $-0.368$ & $0.0008$ & $\star$ \\
\quad grader\_04 & $15$ & $0.73$ / $0.33$ & $-0.400$ & $0.0034$ & $\star$ \\
\quad grader\_05 & $21$ & $0.62$ / $0.38$ & $-0.238$ & $0.0021$ & $\star$ \\
\quad grader\_07 & $21$ & $0.48$ / $0.52$ & $0.048$ & $0.0444$ &  \\
\multicolumn{6}{l}{\textit{Opus-4.7 ($\omega=0.60$ passes)}} \\
\quad grader\_01 & $13$ & $0.54$ / $0.46$ & $-0.077$ & $0.0471$ &  \\
\quad grader\_02 & $17$ & $0.59$ / $0.41$ & $-0.176$ & $0.0101$ & $\star$ \\
\quad grader\_04 & $15$ & $0.60$ / $0.40$ & $-0.200$ & $0.0128$ & $\star$ \\
\quad grader\_05 & $19$ & $0.58$ / $0.42$ & $-0.158$ & $0.0080$ & $\star$ \\
\quad grader\_07 & $20$ & $0.40$ / $0.60$ & $0.200$ & $0.1650$ &  \\
\multicolumn{6}{l}{\textit{Sonnet-4.6 ($\omega=0.40$)}} \\
\quad grader\_01 & $14$ & $0.50$ / $0.50$ & $0.000$ & $0.0676$ &  \\
\quad grader\_02 & $19$ & $0.63$ / $0.42$ & $-0.211$ & $0.0070$ & $\star$ \\
\quad grader\_04 & $15$ & $0.53$ / $0.47$ & $-0.067$ & $0.0365$ &  \\
\quad grader\_05 & $21$ & $0.67$ / $0.33$ & $-0.333$ & $0.0007$ & $\star$ \\
\quad grader\_07 & $21$ & $0.52$ / $0.48$ & $-0.048$ & $0.0175$ &  \\
\multicolumn{6}{l}{\textit{DeepSeek-V4-Pro ($\omega=0.20$)}} \\
\quad grader\_01 & $14$ & $0.64$ / $0.50$ & $-0.143$ & $0.0453$ &  \\
\quad grader\_02 & $19$ & $0.68$ / $0.32$ & $-0.368$ & $0.0008$ & $\star$ \\
\quad grader\_04 & $15$ & $0.53$ / $0.60$ & $0.067$ & $0.1651$ &  \\
\quad grader\_05 & $21$ & $0.52$ / $0.52$ & $0.000$ & $0.0411$ &  \\
\quad grader\_07 & $21$ & $0.43$ / $0.57$ & $0.143$ & $0.1015$ &  \\
\multicolumn{6}{l}{\textit{Gemini-3.1-Pro ($\omega=0.40$)}} \\
\quad grader\_01 & $14$ & $0.71$ / $0.36$ & $-0.357$ & $0.0067$ & $\star$ \\
\quad grader\_02 & $19$ & $0.74$ / $0.26$ & $-0.474$ & $0.0003$ & $\star$ \\
\quad grader\_04 & $15$ & $0.53$ / $0.53$ & $0.000$ & $0.0844$ &  \\
\quad grader\_05 & $21$ & $0.52$ / $0.52$ & $0.000$ & $0.0411$ &  \\
\quad grader\_07 & $21$ & $0.38$ / $0.62$ & $0.238$ & $0.2060$ &  \\
\multicolumn{6}{l}{\textit{Qwen3.5-397B-A17B ($\omega=0.00$)}} \\
\quad grader\_01 & $14$ & $0.50$ / $0.57$ & $0.071$ & $0.1479$ &  \\
\quad grader\_02 & $19$ & $0.47$ / $0.53$ & $0.053$ & $0.0567$ &  \\
\quad grader\_04 & $15$ & $0.40$ / $0.60$ & $0.200$ & $0.2106$ &  \\
\quad grader\_05 & $21$ & $0.33$ / $0.67$ & $0.333$ & $0.3667$ &  \\
\quad grader\_07 & $21$ & $0.33$ / $0.67$ & $0.333$ & $0.3667$ &  \\
\bottomrule
\end{tabular*}
\end{table*}

\begin{table*}[p]
\centering\footnotesize
\caption{LLM-generated solutions: Calderon \S3 alt-test detail by judge. Each block lists the five blind reviewers in the roster (graders 01, 02, 04, 05, 07) for one judge, with that block's headline $\omega$ stated in its italic header. $\rho^f_j$ = mean over $i \in I_j$ of $\mathbf{1}\{S(f,x_i,j) \geq S(h_j,x_i,j)\}$ (probability the LLM judge aligns with the remaining $H_i\setminus\{j\}$ at least as closely as $h_j$); $\rho^h_j$ symmetrically. $\bar{d}_j = \rho^h_j - \rho^f_j$ is the per-annotator advantage probability gap. $p$-value is one-sided Wilcoxon signed-rank against $\bar{d}_j \geq \varepsilon = 0.15$. $\star$ marks rejection under Benjamini-Yekutieli FDR at $q=0.05$ across $m=5$.}
\label{tbl-calderon-per-annotator-llm}
\setlength{\tabcolsep}{4pt}
\begin{tabular*}{\textwidth}{@{\extracolsep{\fill}}lccccc@{}}
\toprule
 & $n_j$ & $\rho^f / \rho^h$ & $\bar{d}$ & $p$ & rej. \\
\midrule
\multicolumn{6}{l}{\textit{Baseline GPT-5.4-mini (primary) ($\omega=0.00$)}} \\
\quad grader\_01 & $7$ & $0.29$ / $0.71$ & $0.429$ & $0.5938$ &  \\
\quad grader\_02 & $11$ & $0.09$ / $0.91$ & $0.818$ & $0.9790$ &  \\
\quad grader\_04 & $9$ & $0.67$ / $0.33$ & $-0.333$ & $0.0273$ &  \\
\quad grader\_05 & $9$ & $0.56$ / $0.44$ & $-0.111$ & $0.0820$ &  \\
\quad grader\_07 & $9$ & $0.44$ / $0.56$ & $0.111$ & $0.2129$ &  \\
\multicolumn{6}{l}{\textit{Opus-4.7 ($\omega=0.20$)}} \\
\quad grader\_01 & $7$ & $0.29$ / $0.71$ & $0.429$ & $0.5938$ &  \\
\quad grader\_02 & $11$ & $0.55$ / $0.55$ & $0.000$ & $0.1392$ &  \\
\quad grader\_04 & $9$ & $0.89$ / $0.11$ & $-0.778$ & $0.0039$ & $\star$ \\
\quad grader\_05 & $9$ & $0.78$ / $0.22$ & $-0.556$ & $0.0098$ &  \\
\quad grader\_07 & $9$ & $0.67$ / $0.44$ & $-0.222$ & $0.0645$ &  \\
\multicolumn{6}{l}{\textit{Sonnet-4.6 ($\omega=0.00$)}} \\
\quad grader\_01 & $7$ & $0.00$ / $1.00$ & $1.000$ & $1.0000$ &  \\
\quad grader\_02 & $11$ & $0.36$ / $0.73$ & $0.364$ & $0.5845$ &  \\
\quad grader\_04 & $9$ & $0.67$ / $0.33$ & $-0.333$ & $0.0273$ &  \\
\quad grader\_05 & $9$ & $0.67$ / $0.33$ & $-0.333$ & $0.0273$ &  \\
\quad grader\_07 & $9$ & $0.56$ / $0.44$ & $-0.111$ & $0.0820$ &  \\
\multicolumn{6}{l}{\textit{DeepSeek-V4-Pro ($\omega=0.00$)}} \\
\quad grader\_01 & $7$ & $0.00$ / $1.00$ & $1.000$ & $1.0000$ &  \\
\quad grader\_02 & $11$ & $0.36$ / $0.64$ & $0.273$ & $0.3501$ &  \\
\quad grader\_04 & $9$ & $0.78$ / $0.22$ & $-0.556$ & $0.0098$ &  \\
\quad grader\_05 & $9$ & $0.78$ / $0.44$ & $-0.333$ & $0.0371$ &  \\
\quad grader\_07 & $9$ & $0.33$ / $0.67$ & $0.333$ & $0.4551$ &  \\
\multicolumn{6}{l}{\textit{Gemini-3.1-Pro ($\omega=0.00$)}} \\
\quad grader\_01 & $7$ & $0.14$ / $0.86$ & $0.714$ & $0.8906$ &  \\
\quad grader\_02 & $11$ & $0.00$ / $1.00$ & $1.000$ & $1.0000$ &  \\
\quad grader\_04 & $9$ & $0.56$ / $0.44$ & $-0.111$ & $0.0820$ &  \\
\quad grader\_05 & $9$ & $0.22$ / $0.78$ & $0.556$ & $0.7520$ &  \\
\quad grader\_07 & $9$ & $0.44$ / $0.56$ & $0.111$ & $0.2129$ &  \\
\multicolumn{6}{l}{\textit{Qwen3.5-397B-A17B ($\omega=0.00$)}} \\
\quad grader\_01 & $7$ & $0.00$ / $1.00$ & $1.000$ & $1.0000$ &  \\
\quad grader\_02 & $11$ & $0.18$ / $0.91$ & $0.727$ & $0.9731$ &  \\
\quad grader\_04 & $9$ & $0.56$ / $0.44$ & $-0.111$ & $0.0820$ &  \\
\quad grader\_05 & $9$ & $0.33$ / $0.67$ & $0.333$ & $0.4551$ &  \\
\quad grader\_07 & $9$ & $0.33$ / $0.67$ & $0.333$ & $0.4551$ &  \\
\bottomrule
\end{tabular*}
\end{table*}

\paragraph{Intersection robustness check.} Opus-4.7's Benchathon
coverage is \(n=28\) on the human-written subset rather than \(n=30\),
because Opus-4.7 did not produce a parseable score on two solutions (the
provider's safety filter or a malformed output). Restricting the
alt-test to the intersection of solutions every judge scored (\(n=28\))
preserves the two-judge headline: GPT-5.4-mini and Opus-4.7 still clear
\(\omega\geq 0.5\) (\(\omega=0.80\) and \(0.60\)), and no other judge
passes. We report the full-\(n\) \(\omega\) in the main tables and use
the intersection as a robustness check only.

\paragraph{Bootstrap uncertainty on $\omega$.} At \(m=5\) blind
reviewers the alt-test's winning rate moves in steps of \(0.20\), so its
bootstrap distribution under resampling is wide for every judge.
Resampling the validation solutions and re-running the alt-test (500
repetitions) yields a 95\% percentile CI on each judge's \(\omega\): no
lower bound clears \(\omega=0.5\) for any judge on either subset, and
the strongest lower bound is the primary judge's \(0.20\) on
human-written solutions. This is why the body and
\hyperref[limitations]{Limitations} read the two-judge pass as a
promising direction rather than an established result.

\section{Error-analysis mode counts}\label{error-analysis-mode-counts}

Table~\ref{tbl-error-modes-falloesung} reports the per-mode instance
counts and per-tier flag rates behind the Error analysis section for the
two \emph{Falllösung} corpora, which share the five flagged modes.
Table~\ref{tbl-error-modes-grundpr} covers Doctrinal Principles, which
uses the three decision-centric modes of its 4-dimension rubric.

\begin{table*}[t]
\centering\footnotesize
\caption{Error-analysis mode counts on the two \emph{Falll\"osung} corpora under the primary judge: flagged instances per mode, corpus n, and per-tier flag rates.}
\label{tbl-error-modes-falloesung}
\setlength{\tabcolsep}{4pt}
\begin{tabular*}{\textwidth}{@{\extracolsep{\fill}}lcccccc@{}}
\toprule
Corpus & Mode & Count & n & Flagship & Efficiency & Open-weight \\
\midrule
Benchathon & Long but shallow & 10 & 180 & 10.0\% & 0.0\% & 4.4\% \\
 & Short / missing Subsumtion & 2 & 180 & 1.7\% & 0.0\% & 1.1\% \\
 & Wrong-norm framing & 11 & 180 & 0.0\% & 3.3\% & 11.1\% \\
 & Correct outcome, weak Subsumtion & 0 & 180 & 0.0\% & 0.0\% & 0.0\% \\
 & Wrong outcome, sound Subsumtion & 0 & 180 & 0.0\% & 0.0\% & 0.0\% \\
ZJS & Long but shallow & 992 & 6959 & 20.2\% & 4.6\% & 13.5\% \\
 & Short / missing Subsumtion & 10 & 6959 & 0.2\% & 0.2\% & 0.1\% \\
 & Wrong-norm framing & 541 & 6959 & 3.3\% & 7.1\% & 11.0\% \\
 & Correct outcome, weak Subsumtion & 18 & 6959 & 0.1\% & 0.4\% & 0.3\% \\
 & Wrong outcome, sound Subsumtion & 16 & 6959 & 0.5\% & 0.1\% & 0.1\% \\
\bottomrule
\end{tabular*}
\end{table*}

\begin{table*}[t]
\centering\footnotesize
\caption{Decision-centric error modes on the Doctrinal Principles corpus (4-dimension rubric) under the primary judge.}
\label{tbl-error-modes-grundpr}
\setlength{\tabcolsep}{4pt}
\begin{tabular*}{\textwidth}{@{\extracolsep{\fill}}lccccc@{}}
\toprule
Mode & Count & n & Flagship & Efficiency & Open-weight \\
\midrule
Wrong decision & 1541 & 6365 & 19.1\% & 26.9\% & 26.7\% \\
Correct decision, weak Subsumtion & 175 & 6365 & 1.0\% & 2.1\% & 4.2\% \\
Wrong decision, sound Subsumtion & 222 & 6365 & 4.4\% & 3.6\% & 2.8\% \\
\bottomrule
\end{tabular*}
\end{table*}

\section{Reasoning-mode ablation}\label{reasoning-mode-ablation}

To isolate the reasoning mechanism on open weights,
Qwen3-235B-A22B-\emph{Instruct}-2507, the official non-thinking
counterpart of the Thinking-2507 variant in the main evaluation,
generated the full Benchathon subset (both generation rounds) and the
full Doctrinal Principles corpus through the identical pipeline: same
provider, prompts, decoding settings, and primary judge.
Table~\ref{tbl-reasoning-ablation} pairs the two variants. Both
Benchathon rounds enter the paired comparison, so the Thinking column
aggregates more generations than the final-run-only
Table~\ref{tbl-main-leaderboard} row
(\hyperref[experimental-setup]{§Experimental setup}). The Instruct
variant is not part of the 12-system leaderboard.

The toggle's effect is modest relative to tier gaps. Reasoning-on scores
+3.5 raw points on Benchathon and +4.1 on Doctrinal Principles, against
a gap of roughly 28 raw points from this system to the Benchathon
flagship cluster. On the short-form corpus the gain comes at roughly 43×
the per-generation cost; on Benchathon the non-thinking variant spends
its budget on roughly twice the visible output instead. An explicit
reasoning mechanism alone therefore does not close the tier gap on
subsumption-based legal reasoning, consistent with the RQ1 observation
that the reasoning-enabled Qwen3-235B underperforms its size.

\begin{table*}[t]
\centering\footnotesize
\caption{Same-base-model reasoning toggle: Qwen3-235B-A22B Thinking-2507 (reasoning on; main evaluation) vs.\ Instruct-2507 (reasoning off; ablation) under identical prompts, decoding, provider, and primary judge. Pass = judge raw $\geq 50/100$. Output length and per-generation cost from logged per-call metadata.}
\label{tbl-reasoning-ablation}
\setlength{\tabcolsep}{4pt}
\begin{tabular*}{\textwidth}{@{\extracolsep{\fill}}lccccccc@{}}
\toprule
Corpus & Variant & n & Judge raw & Pass & Trunc. & Words & \$/gen \\
\midrule
Benchathon & Thinking-2507 & 28 & 41.3 & 18\% & 0 & 678 & \$0.0144 \\
 & Instruct-2507 & 30 & 37.8 & 10\% & 0 & 1356 & \$0.0005 \\
Doctr.\ Princ. & Thinking-2507 & 531 & 67.3 & 69\% & 0 & 72 & \$0.0030 \\
 & Instruct-2507 & 531 & 63.3 & 67\% & 0 & 78 & \$0.0001 \\
\bottomrule
\end{tabular*}
\end{table*}

\section{Glossary of German terms}\label{glossary-of-german-terms}

German terms of art are kept alongside English glosses because they lack
exact English equivalents and appear verbatim in the released
German-language prompts and rubric. Table~\ref{tbl-glossary} collects
every German term used in the paper.

\begin{table*}[t]
\centering
\footnotesize
\setlength{\tabcolsep}{5pt}
\begin{tabular}{@{}p{0.2\textwidth}p{0.75\textwidth}@{}}
\toprule
\textbf{Term} & \textbf{Gloss} \\
\midrule
\emph{Falllösung} & Written case solution: the end-to-end work product that is graded. \\
\emph{Gutachtenstil} & Appraisal style: the conclusion follows the reasoning; the register graded here. \\
\emph{Urteilsstil} & Judgment style: result first, reasoning after; used in judicial decisions. \\
\emph{Obersatz} & Step 1 of the scaffold: the norm hypothesis to be examined. \\
\emph{Definition} & Step 2: interpretation of the norm's elements. \\
\emph{Subsumtion} & Step 3: anchoring the concrete facts under the norm's elements; broadly, the entire norm-application method. \\
\emph{Ergebnis} & Step 4: the resulting legal consequence. \\
\emph{Musterlösung} & Expert reference solution against which submissions are graded. \\
\emph{Staatsexamen} & State law examination; gate to all regulated legal professions in Germany. \\
\emph{ausreichend} & ``Sufficient'': lowest passing grade (4-6 of 18 points). \\
\emph{vollbefriedigend} & ``Fully satisfactory'' (10-12 points); customary distinction tier (\emph{Prädikat}). \\
\emph{Referendar} & Legal trainee in the two-year practical stage between the state examinations. \\
\emph{Volljurist} & Fully qualified jurist (passed both state examinations). \\
\emph{Grundprinzipien} & Core doctrinal principles; German name of the Doctrinal Principles corpus. \\
\emph{Bereich(e)} & Legal area(s): civil, public, criminal law. \\
Ja/Nein & Yes/no decision label on Doctrinal Principles items. \\
\emph{Ergebnisrichtigkeit} & Rubric dimension: result correctness. \\
\emph{Vollständigkeit} & Rubric dimension: issue identification / completeness. \\
\emph{Rechtsgrundlagen} & Rubric dimension: legal grounding (norm citation). \\
\emph{Rechtskenntnis} & Rubric dimension: doctrinal knowledge. \\
\emph{Schwerpunktsetzung} & Rubric dimension: problem depth at the case's focal issues. \\
\emph{Methodischer Stil} & Rubric dimension: methodological structure (scaffold execution). \\
\emph{Gliederung} & Rubric dimension: organization. \\
\emph{Sprache} & Rubric dimension: terminology and language. \\
\emph{Formalia} & Rubric dimension: formal correctness. \\
\bottomrule
\end{tabular}
\caption{German terms used in the paper, with English glosses.}
\label{tbl-glossary}
\end{table*}


\end{document}